%% file: main.tex
\documentclass[10pt]{ic_eee_thesis}
\usepackage[utf8]{inputenc}
\usepackage[round]{natbib}

\degree{B.Sc.}

\title{Solving the Job Shop Scheduling Problem with Graph Neural Networks}
\subtitle{A Customizable Reinforcement Learning Environment} 
\author{Pablo Ariño Fernández}

\supervisor{Prof. Carlos Quesada González} 
\submityear{2025}

\course{Data Science and Artificial Intelligence}

\setboolean{list_of_figures}{true} 
\setboolean{list_of_tables}{false} 
\setboolean{acknowledgement}{true} 
\setboolean{acronyms}{true} 

\setboolean{edge_labels}{true} 

\setboolean{double_spacing}{true} 

\setboolean{final_thesis}{true} 
\secondmarker{} 

\usepackage{amsmath}
\usepackage{amsfonts} 
\usepackage{algorithm}
\usepackage{algpseudocode}
\usepackage{minitoc}
\usepackage{booktabs}
\usepackage{makecell}
\usepackage{changepage}
\usepackage{setspace} 
\hypersetup{
    colorlinks=true,
    linkcolor=black,
    filecolor=magenta,      
    urlcolor=cyan,
    citecolor=black
}
\usepackage{subcaption}
\usepackage{float}
\usepackage{amssymb}
\usepackage{url}
\usepackage[dvipsnames]{xcolor}
\usepackage{tcolorbox}
\newtcolorbox[auto counter, number within=section]{codebox}[2][]{colback=gray!10, colframe=black, fonttitle=\bfseries,
title=Box~\thetcbcounter: #2, #1}

\usepackage{array}
\usepackage{svg}
\usepackage{seqsplit}
\usepackage{longtable}
\usepackage{listings}
\usepackage{enumitem}

\DeclareFixedFont{\ttb}{T1}{txtt}{bx}{n}{12} 
\DeclareFixedFont{\ttm}{T1}{txtt}{m}{n}{12}  

\usepackage{color}
\definecolor{deepblue}{rgb}{0,0,0.5}
\definecolor{deepred}{rgb}{0.6,0,0}
\definecolor{deepgreen}{rgb}{0,0.5,0}

\newcommand\pythonstyle{\lstset{
language=Python,
basicstyle=\ttfamily,
morekeywords={self, as},              
keywordstyle=\color{deepblue},
emph={Operation, JobShopInstance, __init__},          
emphstyle=\color{deepred},    
stringstyle=\color{deepgreen},
showstringspaces=false,
backgroundcolor=\color{white},
numberstyle=\tiny\color{blue},
breaklines=true
}}

\lstnewenvironment{python}[1][]
{
\pythonstyle
\lstset{#1}
}
{}

\newcommand\pythoninline[1]{{\pythonstyle\lstinline!#1!}}

\newtheorem{example}{Example}

\begin{document}

\preamble 


\input{chapters/01-Introduction}

\part{Background}
\input{chapters/02-JSSP}
\input{chapters/03-GNNs}
\input{chapters/04-RLvsIL}
\input{chapters/05-RelatedWork}

\part{Contribution and Results}
\input{chapters/06-JobShopLib}
\input{chapters/07-TheRLEnv}
\input{chapters/08-ExperimentsAndResults}
\input{chapters/09-Ethics}

\conclusions{Conclusions and Future Directions} 

\appendix 
\input{chapters/AppendixA}

\cleardoublepage 
\RemoveLabels 
\thesisspacing 
\bibliographystyle{plainnat}
\bibliography{references}

\end{document}

%% file: chapters/01-Introduction.tex
\doublespacing 

\chapter{Introduction}
\label{ch1}

\begin{spacing}{1} 
\minitoc 
\end{spacing} 
\thesisspacing 

This project addresses the \ac{JSSP}, one of the most classical and essential problems in scheduling. Scheduling is a subset of combinatorial optimization, a branch of applied mathematics and computer science that involves finding the best solution from a finite set of possibilities. This set of feasible solutions is usually too large to perform an exhaustive search.

The \ac{JSSP} has special relevance in the manufacturing sector \citep{Gupta2006jssp_manufacturing, Wang2021drl_jobshop, Zhang2023jssp_manufacturing_review}, where the order in which tasks or operations are performed on each machine can significantly impact productivity. It also has important applications in hospitals \citep{PHAM2008jssp_surgical, Sarfaraj2021jssp_hospitals} and supply chain management \citep{liao2019jssp_supply, lei12023jssp_supply}. Let's look at an example to illustrate this problem:

\begin{example}
A custom furniture workshop with three workers receives daily orders to manufacture various pieces. Today, they have been requested to build a chair, a table, and a cabinet. Each object represents a job. To complete them, they must undergo a series of phases, each requiring a different machine and worker. In particular, a cutting machine is first used to obtain the wooden pieces needed to assemble each object. Then, the chair and the table must be sanded by a sanding machine before being assembled at the assembly station. The cabinet requires sanding after assembly. The goal is to complete all three jobs in the shortest possible time, considering that each machine can only process one operation at a time. We can visualize the times and order requirements in the following table:
\end{example}
\label{ejemplo1}

\begin{table}[!ht]
\begin{center}
\begin{tabular}{c|c|c|c}
\textbf{Job} & \textbf{Operation} & \textbf{Machine} & \textbf{Duration (hours)} \\
\hline & 1. Cut wood & Cutting machine & 2 \\
\textbf{Table}  & 2. Sand pieces & Sanding machine & 2 \\
& 3. Assemble table & Assembly station & 2 \\
\hline & 1. Cut wood & Cutting machine & 1 \\
\textbf{Chair} & 2. Sand pieces & Sanding machine & 1 \\
& 3. Assemble chair & Assembly station & 1 \\
\hline  & 1. Cut wood & Cutting machine & 2 \\
\textbf{Cabinet} & 2. Assemble cabinet & Assembly station & 3 \\
& 3. Sand cabinet & Sanding machine & 3
\end{tabular}
\caption{Example of a \ac{JSSP}. Operations must be completed in the order presented in the table (from top to bottom).}
\end{center}
\end{table}
\label{tab:ejemplo_introduccion}

\begin{figure}[H]
\centering
\includegraphics[width=0.8\textwidth]{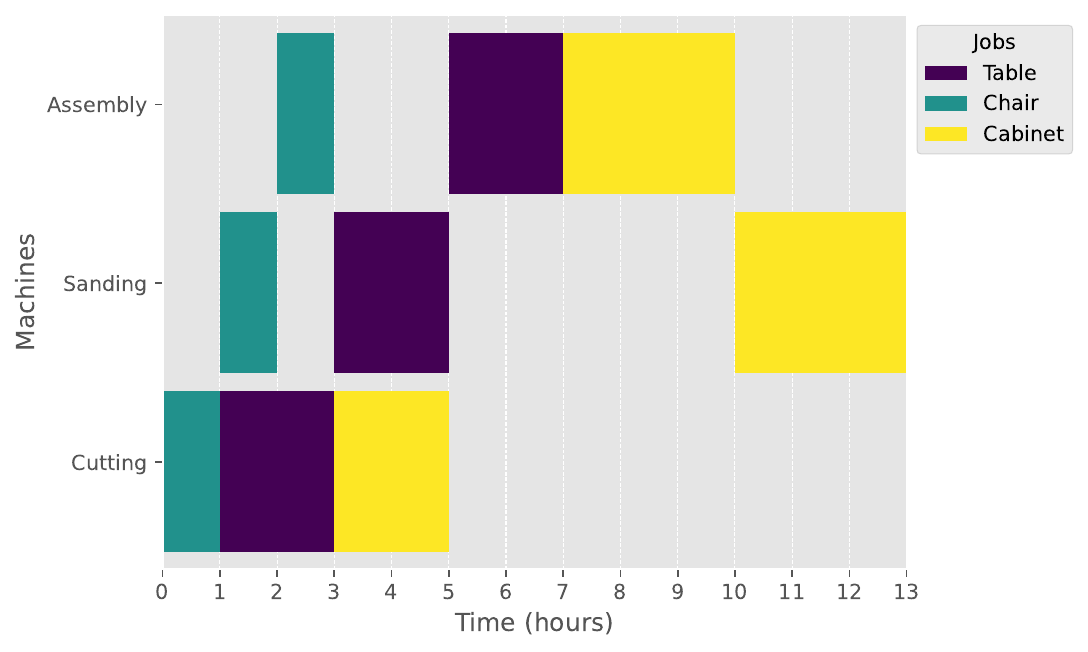}
\caption[Gantt chart representing a possible solution to Example \ref{ejemplo1} (Table \ref{tab:ejemplo_introduccion}) prioritizing operations with shorter duration]{Gantt chart representing a possible solution to Example \ref{ejemplo1} (Table \ref{tab:ejemplo_introduccion}) prioritizing operations with shorter duration. On the vertical axis are the three machines involved: cutting, sanding, and assembly, while the horizontal axis shows the total processing time measured in hours. Each colored block represents an operation of a specific job on a particular machine, with its corresponding duration. For example, the cutting machine begins by processing the chair for one hour (green color), followed by the table for two hours (purple color), and finally, the cabinet for another two hours (yellow color).}
\label{fig:gantt_chart_introduction_example_spt}
\end{figure}

\begin{figure}[H]
\centering
\includegraphics[width=0.8\textwidth]{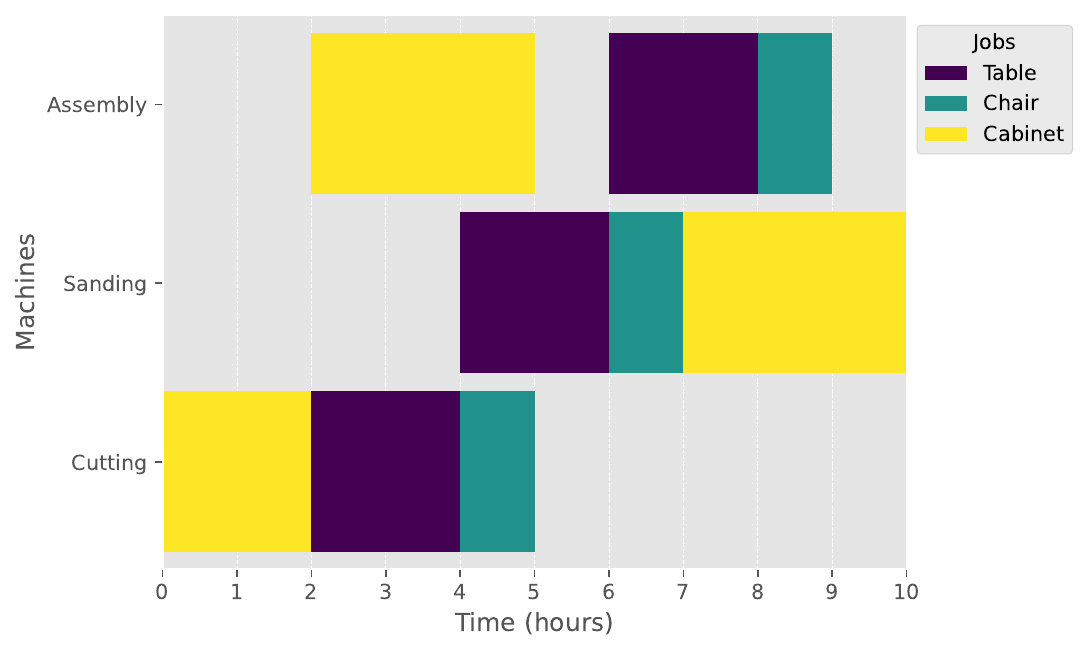}
\caption{Gantt chart representing an optimal solution to Example \ref{ejemplo1} (Table \ref{tab:ejemplo_introduccion}).}
\label{fig:gantt_chart_introduction_example_optimal}
\end{figure}

Not carefully considering the scheduling of each task could lead to significant delays. For example, if we decide to schedule operations according to Figure \ref{fig:gantt_chart_introduction_example_spt}, we would need a total of thirteen hours of work, while with the optimal solution, the same result can be achieved in just ten hours (see Figure \ref{fig:gantt_chart_introduction_example_optimal}).

As we can see, even in this simple example, we achieve a productivity improvement of over 20\%. This difference can mean millions of dollars in economic savings for many manufacturing companies, especially in large-scale systems.

Despite the apparent simplicity of the JSSP, it belongs to the category of NP-hard problems \citep{garey1976complexity}. This means that it is extremely unlikely that an efficient algorithm exists to solve the problem. Finding an optimal solution in scenarios with hundreds or thousands of jobs and machines is virtually impossible.

Given this complexity, heuristics are frequently used to find solutions that, while not optimal, are quick to obtain and meet the problem requirements at a level considered acceptable in practice \citep{boussaid2013optimization_metaheuristics}. Heuristic-based approaches are valuable when speed and practicality take priority over absolute precision. On the other hand, “approximation algorithms" give guarantees on how close the found solution is to the optimum \citep{vazirani2010approximation_algorithms}.

Another problem is the great uncertainty in real situations, such as large factories. Sometimes, unexpected jobs arrive, some machines may temporarily break down, or the duration of each task may not be deterministic, among other difficulties. These considerations quickly render any schedule obsolete, forcing operations to be rescheduled in real-time. We refer to this problem variant as a dynamic \ac{JSSP}.

\ac{PDRs} are frequently used to counter this uncertainty \citep{haupt1989survey_pdr}. Instead of building the complete schedule directly, they create it sequentially, determining which operation should be scheduled next. They can also be used as a starting point for methods that iteratively adjust a given schedule through local search, such as simulated annealing or tabu search. These rules are usually based on simple heuristics. For example, the first schedule (Figure \ref{fig:gantt_chart_introduction_example_spt}) was created using the shortest processing time rule, prioritizing operations with the shortest duration. Additionally, custom dispatching rules are often used. To effectively design these heuristics in real environments, the knowledge and intuition of experts who understand the most common cases and patterns in the factory are often required \citep{kathawala1993expert}. In Section \ref{ch2:pdr}, more details about \ac{PDRs} are given.

This complexity represents a great opportunity to automate the heuristics of dispatching rules through \ac{ML} techniques \citep{bengio2021ml}. With \ac{ML}, the dispatcher learns directly from experience, improving its decisions over time by interacting with the environment \citep{zhang2020l2d, park2021schedule_net, ho2023residual, yuan2023jssp_rl} or by imitating optimal solutions obtained through more computationally expensive algorithms \citep{lee2022imitation_jssp, lee2024il_jssp}. The first family of methods leverages \ac{RL}, a framework in which an agent interacts with an environment to maximize a reward. The second approach employs \ac{IL}, a method for learning directly from optimal schedules through supervised learning. More details about these two learning methods are given in Chapter \ref{ch4}.

Moreover, despite its relevance and being a widely studied problem \citep{xiong2022jssp_survey}, the \ac{JSSP} has received less attention regarding the application of \ac{ML} methods \citep{yuan2023jssp_rl} than other combinatorial optimization problems such as vehicle routing or the traveling salesperson problem.

One advantage of \ac{ML} is that it allows exploiting particular patterns of each use case. This property is especially relevant because, in most real scenarios, one is not interested in solving the problem in general terms but rather a subset of problems with common characteristics \citep{bengio2021ml}. Take the case of the custom furniture manufacturing workshop seen earlier. There may be different orders daily, but the jobs encountered present similar patterns. For example, all furniture will always require a cutting machine initially. A machine learning-based algorithm could be capable of specializing in solving \ac{JSSP} instances (particular definitions of the \ac{JSSP} problem) related to this workshop efficiently.

However, applying this learning-based approach also presents several challenges. One of the most important is adequately representing the relationships between tasks and resources and the constraints that define the order in which operations must be completed. Indices of machines and jobs can be permuted arbitrarily without changing the underlying problem instance. Therefore, representations should be invariant to these permutations.

An effective way to address this complexity is through graphs. They represent elements (nodes) and the connections between them (edges). For example, a disjunctive graph represents operations as nodes, while dependencies between them or shared resources, such as machines, are modeled as edges \citep{blazewicz2000disjunctive_graph}. Graphs are a common way of representing combinatorial optimization problems. The traveling salesperson problem, for example, is about finding the shortest path that traverses all nodes.

\ac{GNNs} \citep{scarselli2009gnns, gilmer2017message_passing, cappart2021co_gnns} are a powerful tool for solving tasks on graphs. Specifically, they can learn vector representations of graph nodes by iteratively aggregating the features of neighboring nodes. An initial node feature could be its processing time, for example. Once computed, the vectors can be used to make predictions. For example, we could predict the dispatching priority of each operation. More details about \ac{GNNs} are given in Chapter \ref{ch3}.

\section{Contribution}

Despite the growing interest in learning \ac{PDRs} with \ac{GNNs} \citep{smit2024gnns_jssp_survey}, there is still a lack of tools for efficient experimentation. Most of the available implementations force users to experiment with a limited set of options.

As we will explore in Chapter \ref{ch5}, previous works differ in several design choices related to the following open questions:
\begin{itemize}
    \item \textbf{What is the optimal graph representation of a partial \ac{JSSP} solution?} The cliques\footnote{A subset of nodes in a graph where every node is connected to every other one. In this case, cliques are formed between operations that share resources.} present in traditional disjunctive graphs can be expensive, computationally speaking, and lead to problems such as oversmoothing\footnote{A phenomenon in \ac{GNNs} where node embeddings become overly similar, regardless of their original differences.}. Other alternatives such as Resource-Task graphs, create artificial machine nodes to remove such cliques. However, more experimentation is still needed to determine what the optimal graph configuration for representing the \ac{JSSP} is.
    \item \textbf{What is the optimal initial set of node features (e.g., each operation)?} While some features can be very straightforward, such as including the operation's processing time as part of this set, other features may require more creativity to be devised. The experiments run in this project suggest that these features can considerably impact performance. In particular, by only using \textit{JobShopLib}'s defined features (i.e., not graph connectivity), we managed to outperform several works in the literature.
    \item \textbf{What operations should be considered available for dispatch?} Sometimes, restricting the dispatcher from selecting probably suboptimal actions can lead to an increase in performance.
    \item \textbf{If using \ac{RL}, what is the best reward function?} The naive reward function gives a zero reward for every step, except for the last one, and yields the makespan in the last step. However, this reward has an important downside---it is sparse. A sparse reward is one that occurs infrequently during an agent's learning process. This sparsity makes it difficult for agents to correlate actions with eventual rewards (the credit assignment problem).
\end{itemize}
The main contribution of this work is the development of \textit{\href{https://github.com/Pabloo22/job_shop_lib}{JobShopLib}}---a versatile Python library---and two highly flexible \ac{RL} environments built upon it. This project also demonstrates these tools' capabilities through experiments requiring minimal setup.

One of the trained models outperformed various GNN-based dispatchers while considering only operation features (no graph). Moreover, our GNN-based model has achieved very close to state-of-the-art results in large-scale problems.

\subsection{\textit{JobShopLib}: A Foundation for JSSP Research}
\textit{JobShopLib} is an open-source library that supports exploring all of these open questions. Key contributions of the library include:
\begin{sloppypar}
\begin{itemize}[itemsep=0pt, topsep=0pt]
    \item Basic \textbf{data structures} to create and manipulate \ac{JSSP} instances and its solutions. These classes make creating, representing, and working with job shop scheduling problems easy.
    \item Access to a \textbf{dataset containing various benchmarks} without downloading external files. \textit{JobShopLib} also introduces a mechanism for storing instances and solutions that supports arbitrary metadata.
    \item A \textbf{random instance generation} module for creating random instances with customizable sizes and properties.
    \item Support for \textbf{multiple solvers}, including \textbf{dispatching rules} and an exact solver based on \textbf{constraint programming} capable of obtaining optimal schedules for small problems.
    \item Support for \textbf{arbitrary graph representations} of the \ac{JSSP}, including specific built-in functions for building disjunctive and Resource-Task graphs. 
    \item A \textbf{visualization} module for plotting Gantt charts and graphs. It also supports creating videos or GIFs from sequences of dispatching decisions or from a dispatching rule solver
    \item Two \textbf{\ac{RL} environments}. The first one (\texttt{SingleJobShopGraphEnv}) allows users to solve a single \ac{JSSP} instance. The \texttt{MultiJobShopGraphEnv} builds upon this environment but supports training \ac{ML} models over a distribution of instances. In particular, after solving one instance, a new random one is generated. Since this second environment can be considered a wrapper of the first, we refer in this project to \texttt{SingleJobShopGraphEnv} as \textit{JobShopLib}'s environment. These two environments utilize almost all the features described and, as mentioned, they are the main contribution of this project.
    
    An important consideration is that \textit{JobShopLib}'s environment is agnostic to the learning method employed. In particular, it follows the standard Gymnasium interface \citep{towers2024gymnasium}. \ac{RL} algorithms are typically designed to interact with this interface. Moreover, despite not using reinforcement learning in our experiments, we prove that the \ac{RL} environment generates training data valuable for imitation learning.

\end{itemize}
\end{sloppypar}
\subsection{Training of a GNN-based Dispatcher}
We demonstrated the practical capabilities of \textit{JobShopLib}'s environment by showcasing its use in training a GNN-based dispatcher. This training uses imitation learning; the GNN is trained to predict the operations that lead to optimal schedules that a constraint programming solver previously found. In other words, the GNN dispatcher learns to imitate the scheduling decisions of an optimal solver. These optimal schedules can only be computed for small instances. The hope is that the model will generalize the patterns found in small instances to bigger ones, where applying a constraint programming solver would be computationally expensive. 

On top of this GNN-based dispatcher, we trained a simpler baseline model to assess \textit{JobShopLib}'s built-in features' expressiveness. This model does not use the message passing mechanism of \ac{GNNs}. Removing this component is equivalent to disconnecting the graph and considering nodes in isolation. We show that these features are enough to outperform several state-of-the-art GNN-based approaches.

\noindent The code of these experiments is available in \href{https://github.com/Pabloo22/gnn_scheduler}{this GitHub repository}. 

\section{Project's Structure}
This project is divided into two parts:
\begin{itemize}[itemsep=0pt, topsep=0pt]
    \item Part I discusses the theoretical background needed to understand this project's contribution. It comprises chapters \ref{ch2}, \ref{ch3}, \ref{ch4}, and \ref{ch5}:

\begin{itemize}[itemsep=0pt, topsep=0pt]
    \item \textbf{Chapter \ref{ch2} }introduces formally the \ac{JSSP} and some core concepts that will be revised later, such as the different types of schedules and solving methods. Moreover, we show how to use basic features of \textit{JobShopLib}, such as creating and solving \ac{JSSP} instances, by showing boxes with brief code snippets and explanations. 

    \item \textbf{Chapter \ref{ch3}} explains how \ac{GNNs} work, including the main architectures employed by \ac{GNNs}-based dispatchers. In particular, the message passing framework is discussed and how it can be adapted to handle heterogeneous graphs (needed for the \ac{JSSP}).

    \item \textbf{Chapter \ref{ch4}} introduces the two main learning methods (\ac{RL} and \ac{IL}) for training deep-learning-based dispatchers. It also introduces the concept of Markov Decision Processes, the theoretical framework used to model \ac{RL} environments. 

    \item Once the theoretical background is set, in \textbf{Chapter \ref{ch5}}, we analyze the literature related to solving the \ac{JSSP} with a GNN-based dispatcher. Here, we identify the core design choices that researchers face. \textit{JobShopLib} abstracts these features to allow users to customize and experiment with a wide range of options.
\end{itemize}

\item Part II explains the core contributions of this project. It consists of chapters \ref{ch6}, \ref{ch7} and \ref{ch8}:

\begin{itemize}[itemsep=0pt, topsep=0pt]
    \item \textbf{Chapter \ref{ch6}} introduces \textit{JobShopLib}'s components necessary to solve the \ac{JSSP} with \ac{PDRs} following a first-principles approach. These components include some of the basic data structures and objects defined, as well as some of the design patterns employed to maintain extensibility. 

    \item \textbf{Chapter \ref{ch7}} discusses the \ac{RL} environment's design and their components. It builds on the concepts introduced in the previous chapters and mentions the built-in classes and functions that replicate some of the specific design choices analyzed in Chapter \ref{ch5}.

    \item \textbf{Chapter \ref{ch8} }contains experiments that prove the environment's usefulness. First, we evaluate the quality of \textit{JobShopLib}'s built-in node features. To accomplish this, we show how a fully equipped GNN can be trained using our library and compare the results with the state of the art.

    \item Finally, \textbf{Chapter \ref{ch9}} reflects on the ethical, environmental, and social aspects affecting our project.
\end{itemize}
\end{itemize}

Finally, we conclude the project by evaluating the project's successes and failures and suggesting future work to take the project further.

%% file: chapters/02-JSSP.tex
\doublespacing 

\chapter{The Job Shop Scheduling Problem}
\label{ch2}

\begin{spacing}{1} 
\minitoc 
\end{spacing} 
\thesisspacing 

The job shop scheduling problem was formally formulated in the 1950s \citep{johnson1954jssp_definition}. Since then, it has been a relevant research topic due to its complexity and broad applicability in manufacturing, including semiconductors, automotive, and electronics production. Although there are numerous variations, in this project, we will address the classical version:

\begin{itemize}[itemsep=0pt, topsep=0pt]
    \item We have a job shop environment with a set of jobs $\mathcal{J}$ and machines $\mathcal{M}$. The size of a problem is denoted as $|\mathcal{J}| \times |\mathcal{M}|$.
    \item Each job $J_i \in \mathcal{J}$ consists of a sequence of operations $O_{i1} \rightarrow O_{i2} \rightarrow \dots \rightarrow O_{in_i}$. These operations must be processed in a given order. The set of all operations in a problem is denoted as $\mathcal{O}$.
    \item Each operation is assigned to a machine $M_j \in \mathcal{M}$, with a processing time $p_{ij} \in \mathbb{N}$. The said machine can only process one operation at a time.
    \item A schedule is defined by assigning to each operation $O_{ij}$ a start time $S_{ij} \in \mathbb{N}$ that satisfies the mentioned constraints.
    \item The main objective is to find a schedule that minimizes the maximum completion time of all jobs, also known as the makespan. This objective is formally defined as the maximum completion time among all tasks $C_{max} = max_{i,j}\{C_{ij} = S_{ij} + p_{ij}\}$.
\end{itemize}
In Section \ref{sec:disjuntive_graph}, we show how the problem can be visualized with a graph, which can be useful for understanding the notation.

In this work, we will focus on problems of this type: processing times are constant, and there are no setup times, due dates, or release dates, among other factors.

\begin{codebox}[label=box:code1]{Representing an Instance in \textit{JobShopLib}}
    To define the problem in \textit{JobShopLib}, the \texttt{JobShopInstance} class is used. This class comprises a list of lists of \texttt{Operation} objects, which act as containers for the properties of each operation. Let's see how the instance from Example \ref{ejemplo1} was defined:
    
    \begin{python}
from job_shop_lib import JobShopInstance, Operation

# Each machine is represented with an ID (starting from 0)
CUTTING_MACHINE_ID = 0
SANDING_MACHINE_ID = 1
ASSEMBLY_STATION_ID = 2
table = [
    Operation(CUTTING_MACHINE_ID, duration=2),
    Operation(SANDING_MACHINE_ID, duration=2),
    Operation(ASSEMBLY_STATION_ID, duration=2),
]
chair = [
    Operation(CUTTING_MACHINE_ID, duration=1),
    Operation(SANDING_MACHINE_ID, duration=1),
    Operation(ASSEMBLY_STATION_ID, duration=1),
]
cabinet = [
    Operation(CUTTING_MACHINE_ID, duration=2),
    Operation(ASSEMBLY_STATION_ID, duration=3),
    Operation(SANDING_MACHINE_ID, duration=3),
]
jobs = [table, chair, cabinet]
instance = JobShopInstance(jobs)
    \end{python}
    
    Through this class, one can access statistics and properties of the problem. Some examples are the number of jobs, machines, or the total duration per job or machine. For more information, see \href{https://job-shop-lib.readthedocs.io/en/latest/api/job_shop_lib.html#job_shop_lib.JobShopInstance}{\texttt{JobShopInstance}'s API reference}.
\end{codebox}

\section{Representing Solutions}
\label{sec:representing_solutions}
On the other hand, a schedule or solution to the problem $\textbf{Y}$ is represented as a matrix of size $|\mathcal{M}| \times |\mathcal{J}|$ in which each row $i$ represents the sequence of operations that machine $M_i \in \mathcal{M}$ will process. For example, the optimal solution to Example \ref{ejemplo1} shown in Figure \ref{fig:gantt_chart_introduction_example_optimal} is a schedule that can be represented by the matrix:

$$ \textbf{Y} =
\begin{pmatrix}
3 & 1 & 2 \\ 
1 & 2 & 3 \\ 
3 & 1 & 2 \\
\end{pmatrix}.
$$

In this matrix, the first row corresponds to the cutting machine $M_1$, the second to the sanding machine $M_2$, and the third to the assembly station $M_3$. The numbers within the matrix represent the jobs that each machine will process in order. The jobs are indexed as $J_1$ for the table, $J_2$ for the chair, and $J_3$ for the cabinet. For example, the first number in the first row indicates that the cutting machine ($M_1$) will begin with the cabinet ($J_3$), then continue with the table ($J_1$), and finally with the chair ($J_2$). This same pattern applies to the other machines, according to the index shown in each row. Note that the indices chosen to represent each machine or job are arbitrary, and any permutation of these would represent the same problem.

\subsection{Types of Schedules}

To obtain a complete schedule with specific timings from a sequence matrix like $\textbf{Y}$, we typically generate the corresponding semi-active schedule. A \textbf{semi-active schedule} is constructed by starting each operation as early as possible, ensuring that no precedence constraints within a job are violated, the required machine is available, and crucially, the operation order specified by the matrix $\textbf{Y}$ for that machine is maintained. For example, if the matrix stipulates that operation $O_{22}$ must be executed before $O_{33}$ on their shared machine, $O_{33}$ will begin only once both its preceding operation within the same job (e.g., $O_{32}$) \textit{and} its machine predecessor ($O_{22}$) have finished processing. Following this ``start as early as possible'' rule for a given sequence naturally eliminates unnecessary idle time for that specific sequence and guarantees finding the schedule with the smallest possible makespan among all schedules respecting the matrix's sequence constraints. Considering semi-active schedules is generally sufficient, as the set of all semi-active schedules is known to contain at least one optimal solution for regular objectives like makespan.

\begin{figure}
\centering
\includegraphics[width=0.5\textwidth]{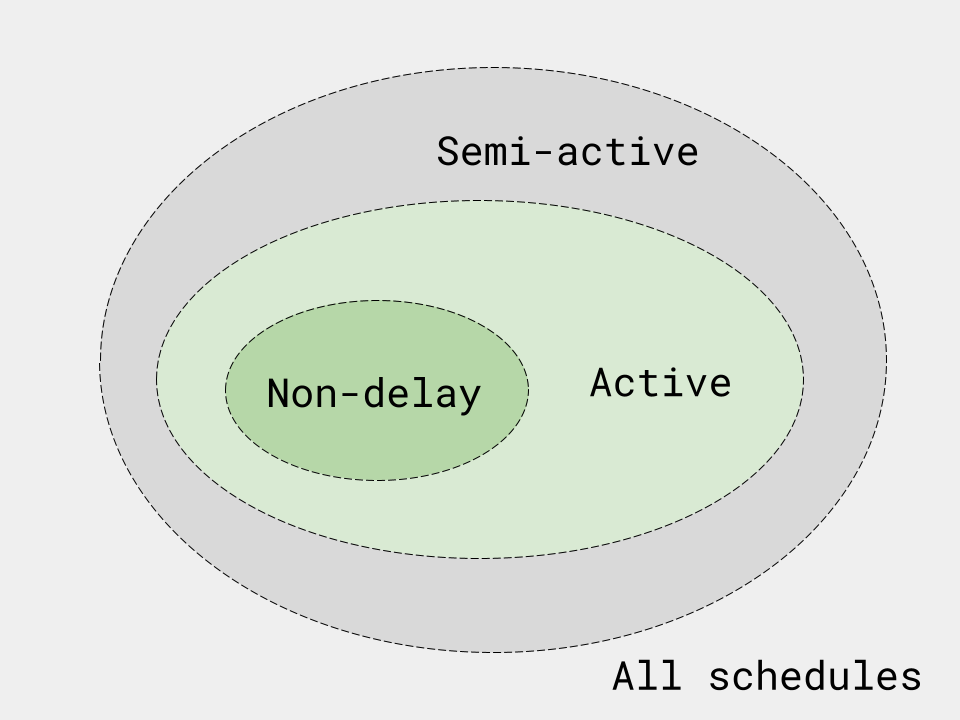}
\caption{Groups of schedules.}
\label{fig:types_of_schedules}
\end{figure}

While semi-active schedules are efficient for a given sequence, it may be possible to improve some of them further by changing the sequence itself (e.g., allowing an operation to ``jump ahead'' of another on the same machine if it doesn't delay anything else overall). Schedules where no such sequence-altering improvements (termed global left-shifts) are possible are called \textbf{active schedules}. This set is a subset of the semi-active schedules and is theoretically important because it also forms a dominant set. This means that we only need to search within active schedules to ensure we find an optimal solution.

Finally, an even smaller subset consists of \textbf{non-delay schedules}. These are active schedules with the additional strict condition that no machine is ever kept idle if there is an operation ready and waiting for processing on that machine. Non-delay schedules are often computationally easier to generate and analyze, and frequently provide high-quality solutions. However, unlike the active set, the non-delay set is not guaranteed to contain an optimal schedule, as sometimes strategically inserting idle time can lead to a better overall makespan (later on, a detailed example will be considered, see Figure \ref{fig:combined}).

\begin{codebox}{Representing Schedules in \textit{JobShopLib}}
    In \textit{JobShopLib}, solutions are represented using the \texttt{Schedule} class. This class comprises a list of lists of \texttt{ScheduledOperation} objects, which relate each operation to a machine and a start time.
    
    To define a schedule through the solution matrix $\textbf{Y}$ mentioned above, we can use the static method \texttt{from\_job\_sequences} of the \texttt{Schedule} class:
    
    \begin{python}
from job_shop_lib import Schedule

# The id of each job is the index in the jobs list
TABLE_ID = 0
CHAIR_ID = 1
CABINET_ID = 2

cutting_machine_order = [CABINET_ID, TABLE_ID, CHAIR_ID]
sanding_machine_order = [TABLE_ID, CHAIR_ID, CABINET_ID]
assembly_station_order = [CABINET_ID, TABLE_ID, CHAIR_ID]
y = [
    cutting_machine_order,
    sanding_machine_order,
    assembly_station_order,
]

schedule = Schedule.from_job_sequences(instance, y)
    \end{python}
    

\end{codebox}


\section{The Disjunctive Graph}
\label{sec:disjuntive_graph}

To better visualize this problem, a disjunctive graph representation $G = (V, C \cup D)$ is commonly used \citep{blazewicz2000disjunctive_graph}; this name is simply due to it being composed of two types of edges ($C$ and $D$). $V$ denotes the set of nodes, each representing an operation $O_{ij}$. An initial node $S$ and a final node $T$ are also added. They serve as starting and ending points for all operations and can be considered operations with zero duration. Directed edges of two types connect nodes. Conjunctive edges $C$ represent precedence relationships: if an operation $O_{ij}$ must be completed before the next operation $O_{i(j+1)}$, there is a directed arc from $O_{ij}$ to $O_{i(j+1)}$. On the other hand, disjunctive edges $D$ are added between operations that require the same machine $M_j$. These edges are not initially oriented, as they represent a resource conflict: the machine can only process one operation at a time, so deciding which operation will be processed first will be necessary.

\begin{figure}[H]
\centering
\includegraphics[width=0.99\textwidth]{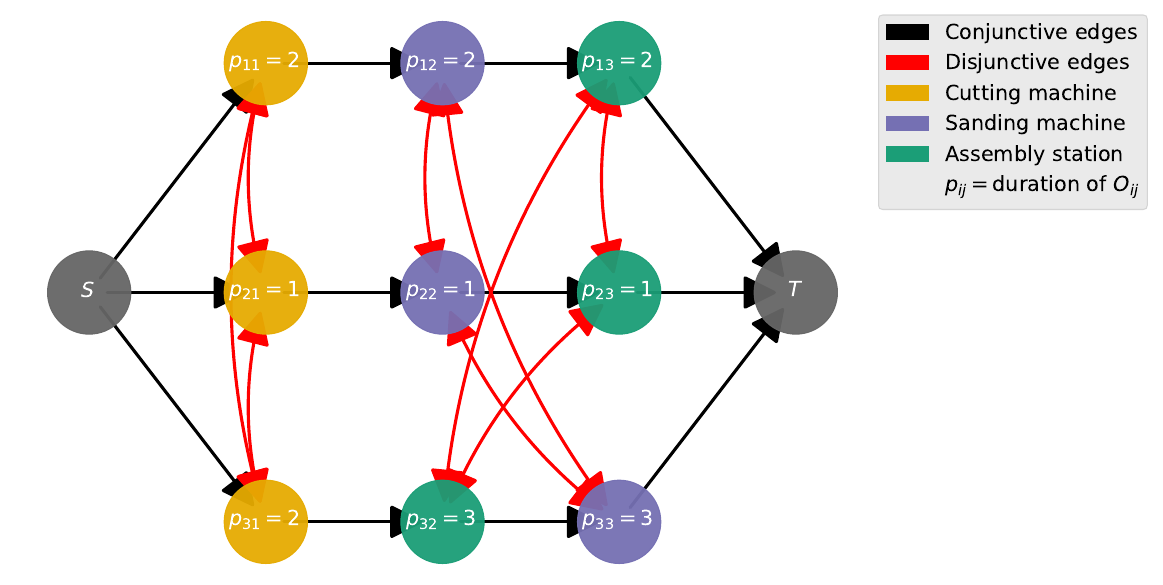}
\caption[Disjunctive graph of Example \ref{ejemplo1}]{Disjunctive graph of Example \ref{ejemplo1}. The label \textit{m} indicates the identifier of the machine required to process the operation, while \textit{d} indicates the duration of the operation ($p_{ij}$) in hours. Regarding colors, disjunctive edges are shown in red, while conjunctive edges are colored black. Gray nodes represent the graph's start ($S$) and end ($T$) nodes. Finally, operations that share resources are colored the same. Each row $i$ represents the operations of job $J_i$.}
\label{fig:disjunctive_graph}
\end{figure}

\begin{figure}[H]
\centering
\includegraphics[width=0.99\textwidth]{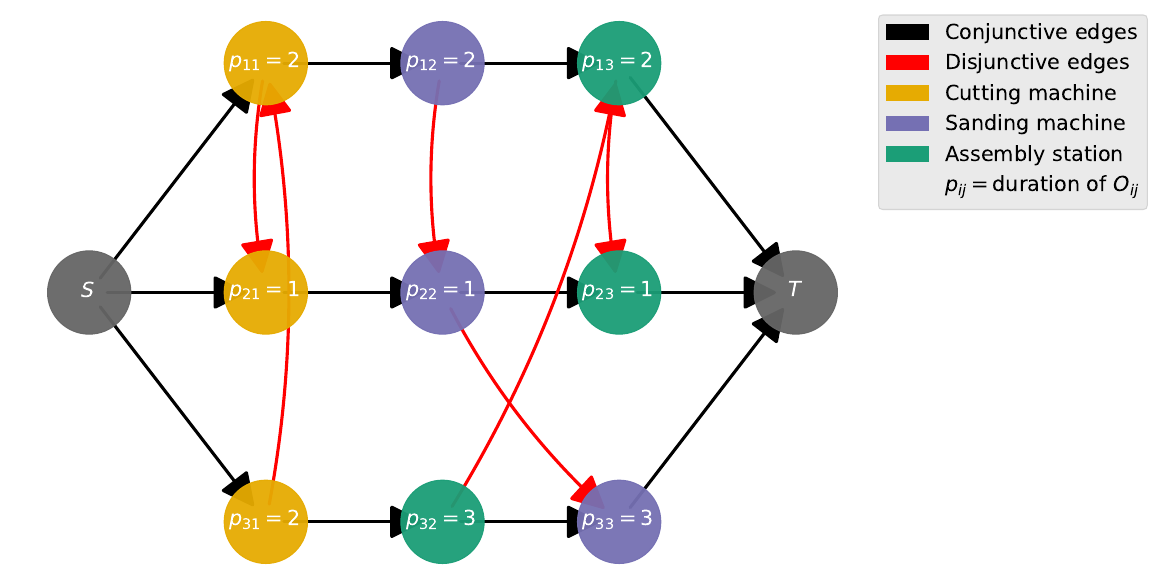}
\caption{Disjunctive graph of Example \ref{ejemplo1} representing the optimal solution presented in Figure \ref{fig:gantt_chart_introduction_example_optimal}.}
\label{fig:disjunctive_graph_solved}
\end{figure}

Disjunctive graphs can also represent solutions through the appropriate orientation of disjunctive arcs (see Figure \ref{fig:disjunctive_graph_solved}). Specifically, the direction indicates the order in which each machine will process each task. For example, the first node of the third row $O_{31}$ is connected to the first node of the first row $O_{11}$. This node, in turn, points to the node representing the first task of the second row $O_{21}$. Therefore, the cutting machine (represented by ID 0) will first process the operation $O_{31}$, then $O_{11}$, and finally $O_{21}$.

\section{Solving the Problem}

Various methods exist to solve the JSSP. They can be classified as ``exact'' \citep{korte2012combinatorial_optimization}, heuristic, or metaheuristic \citep{boussaid2013optimization_metaheuristics}. Exact algorithms, such as \ac{MILP} or branch-and-bound, seek to find the optimal solution but are limited to smaller problem instances due to their computational demands. Heuristic methods provide faster, albeit suboptimal, solutions and are often used in real-time applications where fast decision-making is crucial. This is the category of priority dispatching rules. Finally, metaheuristic approaches, such as genetic algorithms, simulated annealing, and tabu search, balance solution quality and computational efficiency, making them suitable for larger and more complex problem instances. Their name comes from operating at a more general level (``meta'') than traditional heuristics, meaning they can be applied to different combinatorial optimization problems.

In this section, we describe \ac{CP}, an exact method utilized in this work to generate a dataset of optimal instances with which we trained an \ac{ML} model. Secondly, we go into detail about \ac{PDRs}, a method used to compare the performance of this model.

\subsection{Constraint Programming}
\label{subsec:cp}
\ac{CP} \citep{zhou1996cp, rossi2006cp_chapter1} is a paradigm for solving combinatorial search problems belonging to the family of exact algorithms described above. This method allows us to find optimal solutions to small problems quickly. To solve the problem, the user must define a series of variables and constraints for a general optimizer to solve it. In our case, the \ac{JSSP} is encoded in the following way:\\
\textbf{Variables:}
\begin{itemize}[itemsep=0pt, topsep=0pt]
\item $S_{ij}$: Start time of operation $O_{ij}$.
\item $C_{ij}$: Completion time of operation $O_{ij}$: $C_{ij} = S_{ij} + p_{ij}$.
\end{itemize}
\textbf{Objective:} Minimize the makespan ($\max_{i,j}{C_{ij}}$)\\
\textbf{Constraints:}
\begin{itemize}[itemsep=0pt, topsep=0pt]
\item Job sequencing constraints: the completion time $C_{ij}$ must be less than or equal to the start time of the next operation in the job $S_{i(j+1)}$: $C_{ij} \leq S_{i(j+1)}$.
\item Machine non-overlapping constraints:
for each machine $M_{k} \in \mathcal{M}$, none of the operations $O_{ij}$ and $O_{pq}$ assigned to $M_{k}$ can overlap in time:
$S_{ij} \geq C_{pq}$ or $S_{pq} \geq C_{ij}$.
\end{itemize}
Once defined, we use Google's CP-SAT optimizer from the OR-Tools library \citep{laurent2024cpsatlp_ortools}. It searches for possible solutions that satisfy (SAT) the constraints defined. It applies a broad range of techniques, each with strengths and weaknesses. Although each technique runs concurrently, once any of these processes finds a better solution, the result is communicated between them so they can take advantage of this information \citep{krupke2024cpsat}. In contrast, other exact algorithms, such as those mentioned above, specialize in a single strategy that does not always turn out to be the most efficient.

\begin{codebox}[label=box:code3]{Using the CP-SAT optimizer from \textit{OR-Tools} in \textit{JobShopLib}}
    
    \textit{JobShopLib} provides the \texttt{ORToolsSolver} class, which encodes the problem in the previously presented form transparently to the user:
    
    \begin{python}
from job_shop_lib.constraint_programming import ORToolsSolver

cp_sat_optimizer = ORToolsSolver()
optimal_schedule = cp_sat_optimizer(instance)
    \end{python}
\end{codebox}

\subsection{Priority Dispatching Rules}
\label{ch2:pdr}
As mentioned in the Introduction, \ac{PDRs} are one of the most used methods in practice due to their simplicity and ability to be applied to dynamic environments. In a dynamic environment, new jobs may appear, or certain machines may temporarily break down, for example. This uncertainty is something that other techniques, such as \ac{CP}, struggle with. When using \ac{CP}, redefining the constraints and variables and solving the problem again is necessary every time a change occurs in the factory. This limitation, added to the time needed to find an acceptable solution in large-scale problems, significantly hinders their application in real environments.

Dispatching rules do not present these limitations because they create the schedule sequentially. Specifically, each time a machine becomes idle\footnote{The classical definition of \ac{PDRs} assumes that only operations that can start immediately are eligible to be dispatched. The use of alternative definitions is analyzed in Section \ref{sec:action_def} and Subsection \ref{subsec:ready_op_filters}.}, a heuristic that tends to be simple and easy to interpret computes the priority of each operation that can be assigned to that machine \citep{HOLTHAUS1997dispatchingrules}. Table \ref{tab:dispatch_rules} shows classic \ac{PDRs} for the previously defined \ac{JSSP} scenario.

\begin{table}[htbp]
   \centering
   \begin{tabular}{>{\centering\arraybackslash}m{6cm}|>{\centering\arraybackslash}m{8cm}}
       \textbf{Rule} & \textbf{Description} \\
       \hline
       \ac{SPT} & Selects the operation with the shortest processing time $p_{ij}$. \\

       \ac{MWKR} & Prioritizes the job with the greatest total remaining processing time. \\

       \ac{MOR} & Selects the job with the most pending operations.  \\
       \ac{FCFS} & Selects the job with the lowest ID.  \\

       Random & Selects a random operation from those available. \\
   \end{tabular}
   \caption{Some of the standard \ac{PDRs} in \ac{JSSP}}
   \label{tab:dispatch_rules}
\end{table}

An important observation is that by defining the time an operation is selected as ``when a machine becomes idle,'' we implicitly discard the possibility of keeping the machine waiting for a potentially more productive operation that is not yet available. Therefore, this definition also acts as a heuristic that limits the space of available actions to operations that can begin immediately.

\begin{codebox}[label=box:code4]{Using PDRs in \textit{JobShopLib}}
    
    \textit{JobShopLib} provides the \texttt{DispatchingRuleSolver} class, which is capable of applying any assignment rule that a callable (e.g., a function) can represent. Furthermore, the library itself includes the previously mentioned rules, meaning you don't need to program them. These can be invoked using a text string. For example, solving the previous instance using the \textit{most work remaining} rule can be achieved as follows:
    \begin{python}
from job_shop_lib.dispatching import DispatchingRuleSolver

solver_mwkr = DispatchingRuleSolver("most_work_remaining")
schedule_mwkr = solver_mwkr(instance)
    \end{python}
\end{codebox}

%% file: chapters/03-GNNs.tex
\doublespacing 

\chapter{Graph Neural Networks}
\label{ch3}

\begin{spacing}{1} 
\minitoc 
\end{spacing} 
\thesisspacing 

As we mentioned in the introductory chapter, combinatorial optimization presents unique challenges for traditional machine learning. One is that input data often presents irregular structures with a dynamic number of elements and relationships between them. In the \ac{JSSP}, each instance can have a different number of jobs and machines, and the precedence relationships between operations can vary significantly. Due to these characteristics, the \ac{JSSP} is naturally expressed as a graph. For this reason, graph neural networks are an efficient tool for processing the problem \citep{bengio2021ml}.

One of the most important properties of these models is their permutation invariance. In other words, the order in which the graph nodes are presented does not affect the model's final result. In the context of the \ac{JSSP}, this is crucial since the order in which machines or jobs are enumerated should not affect the solution found. For example, given two instances of the same problem where only the numbering order of the machines changes, a traditional neural network might produce different results. At the same time, a GNN guarantees the same output. Possessing this property is essential because it means that, in practice, the network will need less data during its training than a non-permutation-invariant counterpart.

Another fundamental advantage is their ability to process graphs of variable size. This property is essential for the \ac{JSSP}, as it allows us to train a model with small instances and apply it to instances with a larger number of machines and jobs. This generalization capability has been demonstrated empirically in several previous works \citep{zhang2020l2d, park2021schedule_net}.

Furthermore, these models also facilitate the incorporation of domain-specific knowledge. For example, in the \ac{JSSP}, we can encode relevant information in the nodes, such as their processing time, the number of remaining operations in their job, or even the maximum deadline by which we want to complete the operation. This flexibility allows us to represent various problem variants through these algorithms. Although not considered in this work, encoding information in the edges is also possible.

\section{Recap on Neural Networks}
\label{sec:nn_recap}

Before diving into \ac{GNNs}, it is convenient to briefly revisit the fundamentals of traditional \ac{NNs}, also known as \ac{ANNs}. Inspired by biological neurons, these models are the fundamental building block of modern \ac{DL} architectures \citep{lecun2015deep}.

The core component of an NN is the artificial neuron. It receives a set of input signals, computes a weighted sum of these inputs, adds a bias term, and then applies a nonlinear activation function to produce an output. Mathematically, the output $z$ of a single neuron processing an input vector $\mathbf{x} \in \mathbb{R}^n$ can be expressed as:
$$ z = \sigma \left( \sum_{i=1}^{n} w_i x_i + b \right) = \sigma(\mathbf{w}^\top \mathbf{x} + b).$$
Here, $\mathbf{w} \in \mathbb{R}^n$ represents the vector of weights, $b \in \mathbb{R}$ is the bias term, and $\sigma(\cdot)$ denotes the activation function. Common choices for $\sigma$ include the sigmoid function, the \ac{tanh}, or the \ac{ReLU} \citep{nair2010relu}. The introduction of nonlinearity via the activation function is crucial, as it allows the network to model complex, nonlinear relationships within the data.

Neurons are typically organized into layers to form a network. The most prevalent architecture is the \ac{MLP}, which comprises an input layer, one or more hidden layers, and an output layer. In a standard \ac{MLP}, neurons in one layer are fully connected to the neurons in the subsequent layer. The input layer receives the raw feature vector, the hidden layers process this information through successive transformations, learning increasingly abstract representations, and the output layer generates the final prediction, such as a classification score or a regression value. This computation can be formulated as:
$$ \mathbf{h}^{(l+1)} = \sigma^{(l)}(\mathbf{W}^{(l)} \mathbf{h}^{(l)} + \mathbf{b}^{(l)}) $$
where $\mathbf{W}^{(l)}$ is the weight matrix connecting layer $l$ to layer $l+1$, $\mathbf{b}^{(l)}$ is the bias vector for layer $l$, $\mathbf{h}^{(0)} = \mathbf{x}$ represents the initial input features, and $\sigma^{(l)}$ is the activation function applied element-wise to the neurons in layer $l$.

Training an NN involves optimizing its parameters (weights $\mathbf{W}$ and biases $\mathbf{b}$) to minimize a loss function $L$, which measures the discrepancy between the network's predictions and the ground truth. This optimization is achieved using the backpropagation algorithm \citep{rumelhart1986backprop}, which provides an efficient way to compute the gradient of the loss function with respect to the network parameters by applying the chain rule recursively. These gradients are then used by an optimization algorithm, such as  \ac{SGD} or its variants like Adam \citep{kingma2014adam}, to iteratively update the parameters in the direction that minimizes the loss.

\ac{NNs} function as powerful universal function approximators, capable of learning complex mappings between inputs and outputs \citep{Hornik1989universal_approximator, cybenko1989approximation}. However, conventional architectures like MLPs operate on fixed-size input vectors. This property makes them inherently ill-suited for processing data with irregular structures, variable sizes, or explicit relational dependencies, such as the graphs representing \ac{JSSP} instances. For this reason, \ac{GNNs} were designed to operate on and leverage the information present in graph-structured data.

\section{The Message Passing Algorithm}
\label{sec:message_passing}
The message passing algorithm is a fundamental idea behind graph neural networks \citep{gilmer2017message_passing}. At a high level, the concept of ``message passing'' refers to the transmission of information between connected nodes.

Formally, a graph, in the context of \ac{GNNs}, can be defined as $G = (V, E, \mathbf{X})$, where $V$ represents the set of nodes and $E$ the set of edges. The rows of the matrix $\mathbf{X}$ represent the feature vectors $\mathbf{x}_i$ for each node $i \in V$. The algorithm operates in two phases:
\begin{enumerate}[itemsep=0pt, topsep=0pt]
    \item \textbf{Aggregation:} Each node collects information from its neighbors.
    \item \textbf{Update:} Each node updates its representation using the aggregated information.
\end{enumerate}
In particular, for each node $i$, the updated feature vectors for layer $l+1$ can be expressed as:
$$\mathbf{h}^{(l+1)}_i = \phi \Big(\mathbf{h}^{(l)}_i, \bigoplus_{j \in \mathcal{N}_i} \psi^{(l)}(\mathbf{h}^{(l)}_i, \mathbf{h}^{(l)}_j) \Big)$$
where $\mathcal{N}_i$ denotes the set of neighboring nodes of $i$, $\psi^{(l)}$ is a ``message function'' that returns a message $\mathbf{m_j}$ for each neighbor $j$,  $\bigoplus$ is a permutation invariant aggregation operator, $\phi^{(l)}$ is an update function, and $\mathbf{h}_i^{(0)} = \mathbf{x_i}$. The most common options for the aggregation operator $\bigoplus$ are:

$$\bigoplus_{j \in \mathcal{N}_i} \mathbf{m}_j = \sum_{j \in \mathcal{N}_i} \mathbf{m}_j \quad \text{(sum)}$$
$$\bigoplus_{j \in \mathcal{N}_i} \mathbf{m}_j = \max_{j \in \mathcal{N}_i} \mathbf{m}_j \quad \text{(maximum)}$$
$$\bigoplus_{j \in \mathcal{N}_i} \mathbf{m}_j = \frac{1}{|\mathcal{N}_i|} \sum_{j \in \mathcal{N}_i} \mathbf{m}_j \quad \text{(mean)}$$
On the other hand, the update function $\phi^{(l)}$ is typically implemented as a neural network that combines the current node representation with the aggregated messages $\mathbf{m}$ from its neighbors. A common implementation is:
\begin{equation}
    \phi(\mathbf{x}_i, \mathbf{m}) = \text{MLP}([\mathbf{x}_i,\mathbf{m}])
\end{equation}
where $[,]$ denotes vector concatenation.

By stacking multiple layers of this algorithm, each node can receive information from increasingly distant nodes in the graph. Specifically, with $L$ layers, each node can receive information from nodes up to $L$ steps away. In the context of \ac{JSSP}, this property allows dispatching decisions to consider interactions between distant operations in the graph.

This framework can support various architectures, each with its advantages and disadvantages. In the following sections, we will examine some of the most relevant ones for \ac{JSSP}.

\subsection{Graph Convolutional Networks}

\ac{GCNs} \citep{kipf2017semisupervised_gcn} represent one of the first and most influential architectures based on the message-passing algorithm. Their design is inspired by traditional convolutional neural networks (CNNs), extending the convolution operation, widely used in image processing, to the domain of graphs.

\noindent In a GCN, the update of a node's representation can be expressed more specifically as:
$$\mathbf{h}_j^{(l+1)} = \sigma\Bigg(\sum_{j \in \mathcal{N}_i \cup \{i\}} \frac{1}{\sqrt{d_i d_j}} \mathbf{W}^{(l)}\mathbf{h}_j^{(l)}\Bigg)$$
where $d_i$ and $d_j$ are the degrees of nodes $i$ and $j$ respectively, $\mathbf{W}^{(l)}$ is a learnable weight matrix, and $\sigma$ is a nonlinear activation function (for example, ReLU).

In a traditional CNN, a convolution applies a filter that averages neighboring pixels with fixed weights. Similarly, a GCN averages the features of neighboring nodes, but normalizes according to each node's degree ($\sqrt{d_i d_j}^{-1}$). By taking into account the degree of the neighboring node $u$, the importance of highly connected nodes is reduced. Otherwise, the vectors of nodes with very high degrees could significantly contribute to homogenizing the vectors of the remaining nodes, especially after applying several layers of message passing. This phenomenon, known as ``over-smoothing'' is a well-known problem in \ac{GNNs} \citep{Rusch2023over_smoothing}. In the case of \ac{JSSP}, for example, it would mean that the nodes representing each operation would have very similar vectors, which would make it impossible for the model to distinguish which operation to dispatch. However, in practice, the optimal normalization coefficient may vary depending on the problem.

This formulation can be interpreted as a simplified version of the message passing algorithm where the aggregation function is a weighted average by node degree and the update function is a linear transformation followed by a non-linearity. However, \ac{GCNs} have certain limitations: all neighboring nodes contribute in a fixed manner to the update. This limitation motivated the development of more sophisticated architectures such as graph attention networks.

\subsection{Graph Attention Networks}
\label{subsec:gat}
\ac{GATs} \citep{veličković2018gat} extend the capability of \ac{GCNs} through the incorporation of attention mechanisms. Similar to how \ac{GCNs} extend the concept of convolution in images to graphs, \ac{GATs} generalize the concept of self-attention used by transformers \citep{Vaswani2017transformer}. In fact, the transformer architecture can be considered a GAT specialized in fully connected graphs, where each token (subword in a sentence) is considered a node.

The central idea is to use an attention mechanism---a learned weighted average across a set of nodes (tokens in the case of transformers). In particular, for each pair of connected nodes $i$ and $j$, a learned attention coefficient $\alpha_{ij}$ determines the importance of node $j$'s information for updating node $i$'s representation. This way, it solves the limitation of \ac{GCNs}, which use fixed coefficients determined solely by the node's degree. Formally, the update of a node in a GAT layer is expressed as:
$$ \mathbf{h}_i^{(l+1)} = \sigma\Bigg(\sum_{j \in \mathcal{N}_j \cup \{i\}} \alpha_{ij}^{(l)} \mathbf{W}^{(l)}\mathbf{h}_j^{(l)}\Bigg).$$
The attention coefficients are calculated using an attention mechanism $a : \mathbb{R}^d \times \mathbb{R}^d \rightarrow \mathbb{R}$ that calculates $e_{ij}^{(l)}$, the unnormalized importance of node $i$ for node $j$ based on their features:
$$e_{ij}^{(l)} = a(\mathbf{h}_i^{(l)}, \mathbf{h}_j^{(l)}).$$

Once these coefficients have been calculated for all neighbors $j \in \mathcal{N}_i$ of $i$, the coefficients are normalized using the softmax function:
$$ \alpha_{ij}^{(l)} = \frac{\exp(e_{ij}^{(l)})}{\sum_{k \in \mathcal{N}_i} \exp(e_{ik}^{(l)})}.$$

Note that this model is agnostic to the attention mechanism used. For example, in the original paper, a single-layer neural network was used:
$$ e_{ij} = \text{LeakyReLU}(\mathbf{a}^\top [\mathbf{W}\mathbf{h}_j, \mathbf{W}\mathbf{h}_i]) $$
\noindent where $[,]$ denotes the concatenation operation, $\mathbf{a}$ is a vector of learnable parameters, $\mathbf{W}$ is the linear transformation matrix, and LeakyReLU is the activation function:
$$ \text{LeakyReLU}(x) = \begin{cases} 
x & \text{if } x > 0 \\
\alpha x & \text{if } x \leq 0
\end{cases} $$
where $\alpha$ is a small positive value (0.2 in this case) that allows a non-zero gradient for negative inputs. An important observation is that, in this approach, a linear transformation $\mathbf{W}$ is applied to the feature vectors, followed by a concatenation and another linear transformation $\mathbf{a}^\top$. Being consecutive linear operations, they can mathematically collapse into a single linear transformation.
To understand why, let's examine the equation $\mathbf{a}^\top [\mathbf{W}\mathbf{h}_j, \mathbf{W}\mathbf{h}_i]$ more closely. Despite involving a concatenation operation, this can actually be decomposed into a simple linear function of the input node features.
Specifically, if we partition the attention vector $\mathbf{a}$ into two parts, $\mathbf{a} = [\mathbf{a}_1, \mathbf{a}_2]$, corresponding to the dimensions used for $\mathbf{W}\mathbf{h}_j$ and $\mathbf{W}\mathbf{h}_i$ respectively, then:
$$\mathbf{a}^\top [\mathbf{W}\mathbf{h}_j, \mathbf{W}\mathbf{h}_i] = [\mathbf{a}_1^\top \mathbf{W}\mathbf{h}_j, \mathbf{a}_2^\top \mathbf{W}\mathbf{h}_i].$$
This can be further simplified by defining $\mathbf{W}_1 = \mathbf{a}_1^\top\mathbf{W}$ and $\mathbf{W}_2 = \mathbf{a}_2^\top\mathbf{W}$ and $\mathbf{W}_2$, yielding:
$$\mathbf{a}^\top [\mathbf{W}\mathbf{h}_j, \mathbf{W}\mathbf{h}_i] = [\mathbf{W}_1\mathbf{h}_j, \mathbf{W}_2\mathbf{h}_i]$$

Thus, without the LeakyReLU, the sequence of linear transformations followed by concatenation and another linear transformation would mathematically collapse into a single weighted sum of the original node features. 
This limits the model's ability to capture complex relationships between nodes. It is for this reason that it has recently been proposed to modify this attention mechanism as follows \citep{brody2022gatv2}:
$$ e_{ij} = \mathbf{a}^\top\text{LeakyReLU}([\mathbf{W}\mathbf{h}_j, \mathbf{W}\mathbf{h}_i]).$$

By introducing the LeakyReLU non-linearity between linear transformations, the model can learn more complex functions that better capture interactions between nodes. This seemingly simple change allows the model to approximate any desired attention function thanks to the universal approximation theorem \citep{Hornik1989universal_approximator}. For this reason, this latter variant is the most used in practice. However, in general, any attention mechanism can be used.

On the other hand, multiple attention ``heads'' are employed in parallel to improve the stability and expressive power of the model (in the same way as transformers). To do this, the aforementioned process is replicated independently $K$ times using different parameters. This way, each head can learn to capture different types of relationships between nodes. The final aggregation is obtained by concatenating or averaging the outputs of all heads:
$$ \mathbf{h}_i^{(l+1)} = \sigma\Bigg(\frac{1}{K}\sum_{k=1}^K \sum_{j \in \mathcal{N}_i \cup \{i\}} \alpha_{ij}^{(l,k)} \mathbf{W}^{(l,k)}\mathbf{h}_j^{(l)}\Bigg)  \quad \text{(average)} $$
$$\mathbf{h}^{(l+1)}_i = \mathop{\big\|}_{k=1}^{K} \sigma\left(\sum_{j \in \mathcal{N}_i\cup \{i\}} \alpha_{ij}^{(l,k)} {\bf W}^{(l,k)} \mathbf{h}_j\right)  \quad \text{(concatenation)}
$$
here, we denote the concatenation operation with $\|$.

In summary, a GAT layer with multi-head attention operates as follows: each neighboring node $j \in \mathcal{N}_i$ sends a vector of attention coefficients $\boldsymbol{\alpha}_{ij} = [\alpha_{ij}^{(1)}, \ldots, \alpha_{ij}^{(K)}]$, where each element corresponds to a different attention head. These coefficients are used to compute $K$ independent linear combinations of the neighbors' features $\mathbf{h}_j$. Subsequently, these representations are aggregated (typically through concatenation or averaging) to obtain the updated representation of node $i$, denoted as $\mathbf{h}_i^{(l+1)}$.

\subsection{Graph Isomorphism Networks}
\label{subsec:gin}
\ac{GINs} emerge as an architecture designed to maximize discriminative power in graph processing \citep{xu2019gin}. GIN bases its design on the Weisfeiler-Lehman (WL) isomorphism test, an algorithm for determining whether two graphs are non-isomorphic. It aims to develop a GNN whose representational power is equivalent to the WL test.

The objective is to ensure that each layer's aggregation and update functions can distinguish between different multisets of neighbor representations as well as the WL test. Mathematically, this requires the update function to be capable of learning an injective mapping for these multisets. A function $f: X \rightarrow Y$ is injective if for all $x_1, x_2 \in X$, $f(x_1) = f(x_2)$ implies $x_1 = x_2$. This property, applied within the layer update, is crucial for preserving structural information and distinguishing nodes with different neighborhood structures.

\noindent The update of a node's representation in a GIN layer is expressed as:
$$ \mathbf{h}_i^{(l+1)} = \text{MLP}^{(l)} \Bigg((1 + \epsilon^{(l)})\mathbf{h}_i^{(l)} + \sum_{j \in \mathcal{N}_i} \mathbf{h}_j^{(l)}\Bigg)$$
where $\epsilon^{(l)}$ is a learnable or fixed parameter that determines the relative importance of the target node versus its neighbors.

\noindent GIN differs from previous architectures in two fundamental aspects:
\begin{enumerate}[itemsep=0pt, topsep=0pt]
    \item \textbf{Sum aggregation:} Unlike the weighted averaging or attention mechanisms used in GCN and GAT, GIN employs summation as its neighborhood aggregation function. \citet{xu2019gin} proved that summation is the most expressive aggregation function for multisets. It is capable of preserving maximum information about the neighborhood feature distribution, mirroring the counting aspect of the WL test. Mean or max aggregators, in contrast, can map different multisets to the same representation, limiting their discriminative power. For example, the mean and max aggregators cannot distinguish these two different sets of neighbor features: $\text{mean}(\{1, 1, 1\}) = \text{max}(\{1, 1, 1\}) = 1$ and $\text{mean}(\{1, 1\}) = \text{max}(\{1, 1\}) = 1$.
    \item \textbf{MLP for update:} The aggregated representation (sum of neighbors plus the node's own representation) is processed by an \ac{MLP}. As universal function approximators \citep{Hornik1989universal_approximator}, \ac{MLP}s can learn arbitrarily complex functions. Importantly, they can approximate the injective functions required to ensure that different input multisets (representing different local structures) are mapped to different output embeddings, thus maximizing the layer's ability to distinguish nodes based on their neighborhood structure.
\end{enumerate}

\section{Relational Graph Neural Networks}
\label{sec:rgnns}
\ac{RGNNs} extend the GNN paradigm to handle graphs with different types of relationships between nodes. This generalization is crucial for modeling complex systems where interactions between entities can be of a diverse nature. For example, in the context of \ac{JSSP}, this means that the network can process information coming through disjunctive edges (connecting operations that share a machine) and conjunctive edges (connecting operations of the same job) differently. Similarly, this paradigm allows the designing of new graph representations by introducing nodes of different types (e.g., machine nodes).

The fundamental idea behind these architectures is that relationships are not simply binary links between nodes but possess their own semantics that must be considered during message aggregation. Thus, when a node updates its representation, it must take into account not only the characteristics of its neighbors but also the specific nature of each connection that links it to them. This approach allows the network to learn patterns specific to each type of relationship, improving its ability to model systems with heterogeneous interactions.

Message aggregation in these networks must adapt to process multiple types of relationships simultaneously. For this, typically, different transformations are employed for different types of edges, allowing the network to learn how each one should influence the update of node features. That is, a different weight matrix is applied for each type of edge.

For example, with \ac{GCNs}, we can modify the update equation to account for different types of relationships \citep{Schlichtkrull2018rgcn}. Let $\mathcal{R}$ be the set of relationship types and $\mathcal{N}^r_i$ the set of neighbors of node $i$ connected through a relationship of type $r$. The update equation can be expressed as:

$$\begin{aligned} h_i^{(l+1)}= \sigma \left( \sum _{r \in \mathcal {R}}\sum _{j \in \mathcal {N}^r_i} \frac{1}{c_{i,j,r}}W_r^{(l)} h_j^{(l)} + W_0^{(l)}h_i^{(l)} \right) \end{aligned}
$$
\noindent where $\mathbf{W}_r^{(l)}$ is a weight matrix specific to each relationship type $r$ and $c_{i,j,r}$ is a normalization constant that can be learned or fixed beforehand (e.g., $|\mathcal{N}^r_i|$). 

\subsection{Relational Message Passsing}
\label{subsec:relational_mp}
Similarly, relational variants of the other presented architectures can be defined. In general, we can extend the message passing framework
$$\mathbf{h}^{(l+1)}_i = \phi \Big(\mathbf{h}^{(l)}_i, \bigoplus_{j \in \mathcal{N}_i} \psi^{(l)}(\mathbf{h}^{(l)}_i, \mathbf{h}^{(l)}_j) \Big)$$
to handle heterogeneous graphs containing multiple node types ($\mathcal{T}_V$) and edge types ($r = (t_{src}, \text{rel}, t_{dst}) \in \mathcal{T}_E$).

\noindent The adaptation involves making the core components type-aware. For a node $i$ of type $t_{dst}$:
\begin{itemize}[itemsep=0pt, topsep=0pt]
    \item \textbf{Type-specific messages ($\psi_r^{(l)}$):} A distinct message function $\psi_r^{(l)}$, often with relation-specific parameters (like $\mathbf{W}_r^{(l)}$ from the relational GCN), is used for each incoming relation type $r$. It computes messages from neighbors $j \in \mathcal{N}_r(i)$.
    \item \textbf{Hierarchical aggregation ($\bigoplus$):} First, messages $\psi_r^{(l)}(\mathbf{h}_i^{(l)}, \mathbf{h}_j^{(l)})$ from neighbors $j$ connected via the same relation $r$ are aggregated. Then, these relation-specific results are aggregated across all incoming relations targeting node $i$.
    \item \textbf{Type-specific update ($\phi_{t_{dst}}^{(l)}$):} The update function $\phi^{(l)}$ can be specialized for the type $t_{dst}$ of the node being updated, combining its previous state $\mathbf{h}_i^{(l)}$ with the final aggregated message.
\end{itemize}
Thus, the update for node $i$ (of type $t_{dst}$) becomes:
$$\mathbf{h}_i^{(l+1)} = \phi_{t_{dst}}^{(l)} \left( \mathbf{h}_i^{(l)}, \bigoplus_{r \in \mathcal{R}} \left( \bigoplus_{j \in \mathcal{N}_i^r} \psi_r^{(l)}(\mathbf{h}_i^{(l)}, \mathbf{h}_j^{(l)}) \right) \right).$$
This structure allows diverse GNN layers to be applied selectively based on the relationship type within a single heterogeneous graph layer, offering significant modeling flexibility. If we replace the message function and aggregation $\bigoplus_{j \in \mathcal{N}_r(i)} \psi_r^{(l)}(\mathbf{h}_i^{(l)}, \mathbf{h}_j^{(l)})$ by one of the aforementioned layers, such as GIN or GAT, we would get the exact formula for these variants.

The relational GAT variant can be defined by applying the GAT update independently for each relation $r$ and aggregating the results using $\bigoplus$:
\begin{equation}
    \mathbf{h}_i^{(l+1)} = \bigoplus_{r \in \mathcal{R}} \underbrace{\left( \sigma_r\Bigg(\sum_{j \in \mathcal{N}_i^r \cup \{i\}} \alpha_{ij}^{(l,r)} \mathbf{W}_r^{(l)}\mathbf{h}_j^{(l)}\Bigg) \right)}_{\text{Output of GAT layer for relation } r}. 
\end{equation}
\label{eq:rgat}

This formula represents the output for node $i$ by first computing the standard GAT update independently for each incoming relation type $r \in \mathcal{R}$. $\alpha_{ij}^{(l,r)}$ is the corresponding relation-specific attention coefficient calculated based on the neighborhood $\mathcal{N}_i^r$. The activation $\sigma_r$ is also applied per relation. Finally, we could also implement an across-relation attention mechanism by defining $\bigoplus$ as another learned weighted average across the embeddings representing each relationship.

Similarly, the relational GIN variant applies the GIN update per relation and aggregates using $\bigoplus$:
$$ \mathbf{h}_i^{(l+1)} = \bigoplus_{r \in \mathcal{R}} \underbrace{\left( \text{MLP}_r^{(l)} \Bigg((1 + \epsilon_r^{(l)})\mathbf{h}_i^{(l)} + \sum_{j \in \mathcal{N}_i^r} \mathbf{h}_j^{(l)}\Bigg) \right)}_{\text{Output of GIN layer for relation } r}.$$

Here, the standard GIN update is computed for each relation $r$, using a relation-specific MLP ($\text{MLP}_r^{(l)}$) and parameter $\epsilon_r^{(l)}$. The neighborhood $\mathcal{N}_i^r$ is defined by the relation's structure $\mathbf{A}_r$. Crucially, $\mathbf{h}_i^{(l)}$ represents features of the destination node $i$, while $\mathbf{h}_j^{(l)}$ represents features of the source-type neighbors $j \in \mathcal{N}_i^r$. This standard GIN formulation assumes these features are directly combinable (e.g., same dimension); if not, modifications within the $\text{MLP}_r^{(l)}$ or prior feature transformation would be necessary. The outer $\bigoplus$ aggregates the outputs generated independently by the GIN layer for each relevant relation type $r$. To preserve its expressive power, a typical choice for $\bigoplus$ is the sum for the reasons mentioned in Subsection \ref{subsec:gin}.

%% file: chapters/04-RLvsIL.tex
\doublespacing 

\chapter{Reinforcement Learning vs. Imitation Learning}
\label{ch4}

\begin{spacing}{1} 
\minitoc 
\end{spacing} 
\thesisspacing 

Reinforcement learning and imitation learning represent two distinct paradigms for training agents to solve sequential decision-making problems such as job shop scheduling. Reinforcement learning aims to maximize a reward function (e.g., negative makespan). To achieve this objective, an agent (e.g., a GNN dispatcher) interacts with the environment, learning from its mistakes. On the other hand, imitation learning utilizes a dataset generated by an expert (e.g., a \ac{CP} solver). The agent does not interact with the environment but predicts what the expert would have done in that particular situation. In both cases, \ac{ML} models are trained employing small instances, hoping to generalize this learned knowledge to bigger ones.

This chapter also introduces Markov decision processes. They are the mathematical framework used to model sequential decision-making problems, including the \ac{JSSP}. This framework helps us explain the components present in both reinforcement and imitation learning. It is also the foundation for Chapter \ref{ch5}, where we analyze how previous work has defined each element. This chapter is essential to understand \textit{JobShopLib}'s contribution, because our library aims to support experimenting with changes to these components.

This section introduces the foundational concepts of these approaches, with particular focus on the methods typically employed to solve the \ac{JSSP}: policy gradient methods (\ac{RL}) and behavioral cloning (\ac{IL}). Understanding behavioral cloning is especially relevant because it is the learning method employed by the experiments described in Chapter \ref{ch8}. Policy gradient methods are described to understand previous works and justify the reward element of \textit{JobShopLib}'s environment. While they are not necessary for understanding this project's experiments, they are an example of specific popular learning methods that can benefit from \textit{JobShopLib}'s environment. 

\section{Markov Decision Processes}
\label{ch4:mdp}

To understand reinforcement learning, we first need to introduce the concept of an \ac{MDP} \citep{bellman1957markovian}, which provides a mathematical framework for modeling sequential decision-making problems. An \ac{MDP} is defined as a tuple $(\mathcal{S}, \mathcal{A}, P, R, \gamma)$ where:
\begin{itemize}[itemsep=0pt, topsep=0pt]
    \item $\mathcal{S}$ is a set of states representing the environment's configuration.
    \item $\mathcal{A}$ is a set of actions the agent can take.
    \item $P: \mathcal{S} \times \mathcal{A} \times \mathcal{S} \rightarrow [0, 1]$ is the transition probability function, where $P(s'|s,a)$ represents the probability of transitioning to state $s'$ given that the agent takes action $a$ in state $s$.
    \item $R: \mathcal{S} \times \mathcal{A} \times \mathcal{S} \rightarrow \mathbb{R}$ is the reward function, where $R(s,a,s')$ represents the immediate reward received after transitioning from state $s$ to state $s'$ by taking action $a$.
    \item $\gamma \in [0, 1]$ is the discount factor that determines the importance of future rewards.
\end{itemize}

In the context of the job shop scheduling problem, the state $s \in \mathcal{S}$ represents the current partial schedule, including unscheduled operations and which machines are currently busy. The action $a \in \mathcal{A}$ corresponds to selecting the next operation to schedule on an available machine. The transition function updates the schedule deterministically based on the chosen operation, and the reward function provides feedback based on metrics such as makespan reduction or any other custom objective.

The goal in an MDP is to find a policy $\pi: \mathcal{S} \rightarrow \mathcal{A}$ that maximizes the expected cumulative discounted reward:
$$J(\pi) = \mathbb{E}_{\pi}\left[\sum_{t=0}^{\infty} \gamma^t R(s_t, \pi(s_t), s_{t+1})\right]$$
\noindent where $\mathbb{E}_{\pi}$ denotes the expected value when following policy $\pi$. Note that $J(\pi)$ represents the optimization objective that quantifies the long-term performance of policy $\pi$ from the initial state distribution, accounting for both immediate rewards and future consequences of current decisions through the discounting mechanism.  Note also that this formula contains an implicit recurrence, as each state $s_{t+1}$ is determined by the current state $s_t$ and action $\pi(s_t)$ according to the transition probabilities $P$. This creates a chain of dependencies where each state influences all future states and rewards in the sequence.

A key characteristic of \ac{MDP} is the Markov property, which states that the future state depends only on the current state and action, not on the history of previous states and actions.

\subsection{Semi-Markov Decision Processes}

While \ac{MDP}s assume that actions take a uniform amount of time, the \ac{JSSP} is usually formulated in a way that involve actions with variable durations. A \ac{SMDP}\citep{ross1992applied} extends \ac{MDP}s to account for this temporal variability.

An \ac{SMDP} is defined similarly to an \ac{MDP} but includes an additional component for the duration of actions. Formally, an \ac{SMDP} is a tuple $(\mathcal{S}, \mathcal{A}, P, R, F, \gamma)$ where:
\begin{itemize}[itemsep=0pt, topsep=0pt]
    \item $\mathcal{S}$, $\mathcal{A}$, $P$, $R$, and $\gamma$ are defined as in an \ac{MDP}.
    \item $F: \mathcal{S} \times \mathcal{A} \times \mathcal{S} \times \mathbb{R}^+ \rightarrow [0, 1]$ is the sojourn time distribution, which measures the time spent in a state before transitioning to another state. In other words, $F(s, a, s', \tau)$ represents the probability that the state transition from $s$ to $s'$, under action $a$, takes time $\tau$.
\end{itemize}

The objective in an \ac{SMDP} is to find a policy that maximizes the expected cumulative discounted reward. The only difference in practice with an \ac{MDP} is that the discount factor needs to be adjusted to account for the variable time between actions:

$$J(\pi) = \mathbb{E}_{\pi}\left[\sum_{k=0}^{\infty} \gamma^{t_k} R(s_k, \pi(s_k), s_{k+1})\right]$$

\noindent where $t_k$ represents the time at which the $k$-th decision is made.

\section{Reinforcement Learning}

\begin{figure}
    \centering
    \includegraphics[width=0.75\linewidth]{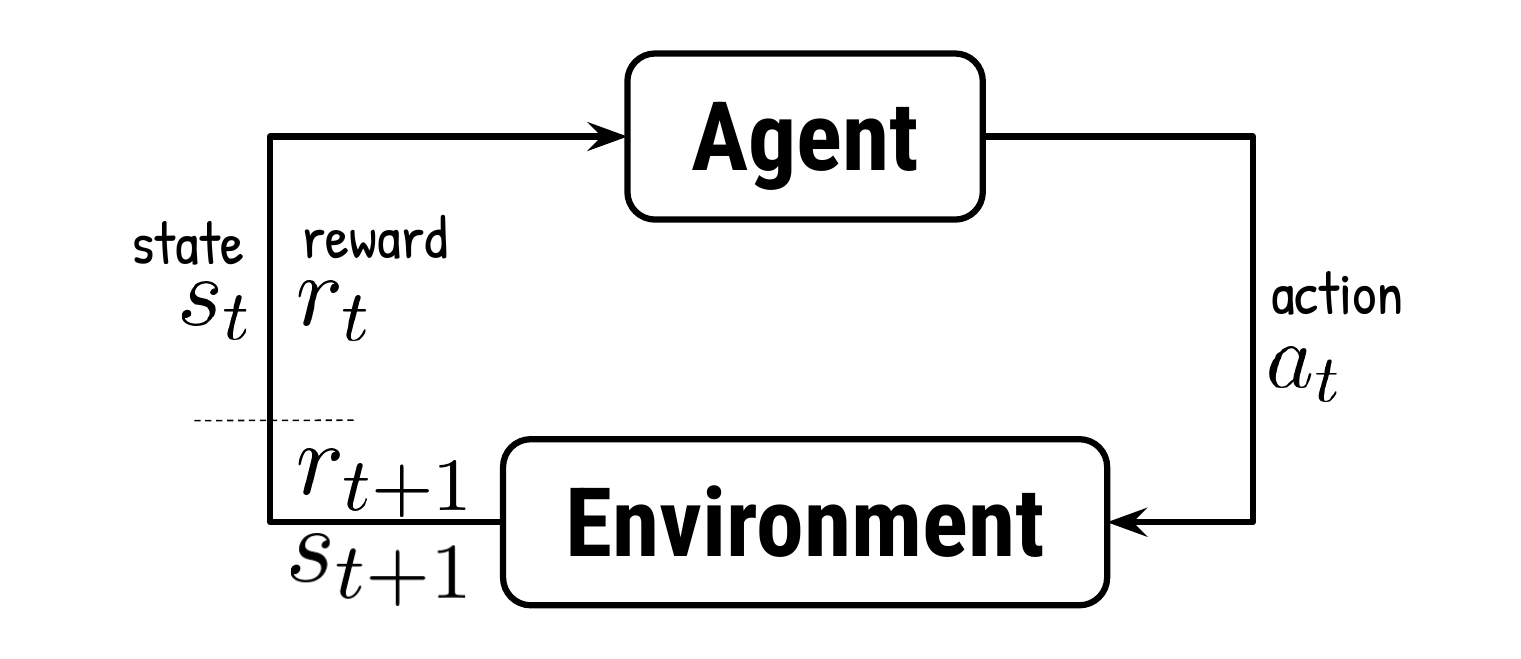}
    \caption{The reinforcement learning loop}
    \label{fig:rl-loop}
\end{figure}

Reinforcement learning \citep{Sutton1998} is a learning paradigm where an agent learns to make sequential decisions by interacting with an environment modeled as an \ac{MDP}. The fundamental difference between \ac{RL} and other machine learning approaches is that \ac{RL} involves learning from interaction rather than from labeled examples.

In \ac{RL}, the agent does not have explicit knowledge of the transition and reward functions but must learn them through experience. The learning process involves exploring the environment, trying different actions, and observing the resulting states and rewards. The goal is to learn a policy that maximizes the expected cumulative reward.

\subsection{Value Functions}
In reinforcement learning, agents learn through experience by interacting with their environment across many different states. While some \ac{RL} approaches like policy gradient methods can directly optimize policies without explicit state evaluations, many powerful algorithms rely on the concept of “value"---how good it is to be in a particular state or to take a specific action.

Value functions provide an intuitive way to evaluate decisions: a state is valuable if it leads to high immediate rewards plus access to other valuable future states. Mathematically, the state-value function for policy $\pi$ is denoted as:
$$V^{\pi}(s) = \mathbb{E}_{\pi} \left[ \sum_{t=0}^{\infty} \gamma^t R(s_t, \pi(s_t), s_{t+1}) \mid s_0 = s \right].$$

Similarly, we can define how good it is to take a specific action in a given state. The action-value function returns the expected return after taking action $a$ in state $s$ and then following policy $\pi$:
$$Q^{\pi}(s,a) = \mathbb{E}_{\pi} \left[ \sum_{t=0}^{\infty} \gamma^t R(s_t, \pi(s_t), s_{t+1}) \mid s_0 = s, a_0 = a \right].$$

For example, if we know the value associated with each state-action pair for our policy, a better policy can be derived by choosing the action with the highest associated value:
$$a(s) = \arg\max_a Q^{\pi}_(s,a).$$
\noindent This is what many \ac{RL} methods, such as Q-learning, do.

However, sometimes it is not necessary to calculate the absolute value of the expected return after taking action $a$, but rather to know the relative advantage of taking that action. The advantage function calculates this relative improvement, which measures how much better it is to take action $a$ compared to sampling an action following the probability distribution given by policy $\pi(\cdot|s)$. Mathematically, this can be expressed as:
$$A^{\pi}(s,a) = Q^{\pi}(s,a) - V^{\pi}(s).$$

\subsection{Policy Gradient Methods}

In practice, the state and action spaces are often too large, or even continuous, making it infeasible to maintain a table of values for each state or state-action pair. Instead, function approximation methods are used to represent the policy, value function, or action-value function. For example, methods in the $Q$-learning family attempt to approximate the function $Q^*(s, a)$ associated with the optimal policy using a parameterized function $Q_{\theta}(s,a)$, where $\theta$ are the parameters of the function (for example, the weights of a neural network).

In this section, we focus on methods that try to learn the optimal policy directly because they are the most used in practice for solving the \ac{JSSP} \citep{zhang2020l2d, Park2021l2s, park2021schedule_net, ho2023residual}. For this, the policy is usually parameterized using a differentiable function $\pi_{\theta}(a|s)$ with respect to its parameters. The goal is to find the values of $\theta$ that maximize the expected return $J(\pi_{\theta})$. Specifically, we would like to optimize the policy using gradient ascent:

$$\theta_{k+1} = \theta_k + \alpha \nabla_{\theta}J(\pi_{\theta})|_{\theta_k}.$$

The policy gradient theorem \citep{sutton1999policy_gradient} provides us with a way to compute this gradient (i.e., how to adjust policy parameters to increase expected rewards). At its core, the policy gradient theorem tells us we can improve a policy by making advantageous actions more likely and disadvantageous actions less likely, without needing to model how our policy changes affect the environment's state distribution.

$$\nabla_\theta J(\pi_\theta) = \mathbb{E}_{s \sim d^{\pi}, a \sim \pi_\theta}\left[\nabla_\theta \log \pi_\theta(a|s) \cdot Q^{\pi_\theta}(s,a)\right]$$

\noindent where $d^{\pi}$ is the state distribution induced by policy $\pi$ and $Q^{\pi_\theta}(s,a)$ is the expected return when taking action $a$ in state $s$ and following policy $\pi_\theta$ thereafter. This formula tells us to adjust policy parameters in proportion to how much an action change affects the policy (the gradient term) and how good the action is (the Q-value).

\noindent Some of the more commonly used policy gradient algorithms are:
\begin{itemize}[itemsep=0pt, topsep=0pt]
    \item \textbf{REINFORCE} \citep{williams1992reinforce} is the most basic policy gradient algorithm, which uses Monte Carlo estimates of the return to update the policy. After collecting a batch of $N$ episodes, the policy is updated with the following formula:
    
    $$\nabla_\theta J(\pi_\theta) \approx \frac{1}{N} \sum_{i=1}^{N} \sum_{t=1}^{T_i} \nabla_\theta \log \pi_\theta(a_t^i|s_t^i) \cdot G_t^i$$
    
    where $G_t^i = \sum_{k=t}^{T_i} \gamma^{k-t} r_k^i$ is the observed return starting from step $t$ in episode $i$, and $T_i$ is the length of episode $i$. While simple, REINFORCE suffers from high variance in gradient estimates.
    
    \item \textbf{\ac{A2C}} \citep{mnih2016asynchronous} reduces variance by using a critic network to estimate the advantage function $A(s,a)$, which measures how much better taking action $a$ is compared to the average action in state $s$:
    
    $$\nabla_\theta J(\pi_\theta) \approx \frac{1}{B} \sum_{i=1}^{B} \nabla_\theta \log \pi_\theta(a_i|s_i) \cdot A(s_i,a_i)$$
    
    where $B$ is the batch size, and $A(s_i,a_i)$ is typically estimated as $r_i + \gamma V(s'_i) - V(s_i)$ for a one-step advantage, or using Generalized Advantage Estimation \citep{Schulman2015GAE} for multi-step advantages.
    
    \item \textbf{\ac{PPO}} \citep{Schulman2017PPO} improves training stability by limiting the policy update size through a clipped objective function:
    $$L^{CLIP}(\theta) = \frac{1}{B} \sum_{i=1}^{B}\left[\min\left(r_i(\theta) A(s_i,a_i), \text{clip}(r_i(\theta), 1-\epsilon, 1+\epsilon) A(s_i,a_i)\right)\right]$$
    where $r_i(\theta) = \frac{\pi_\theta(a_i|s_i)}{\pi_{\theta_{old}}(a_i|s_i)}$ is the probability ratio between the new and old policies, $B$ is the batch size, and $\epsilon$ is a hyperparameter that limits the policy change magnitude, typically 0.2.
\end{itemize}

\section{Imitation Learning}
\label{sec:il}

Unlike reinforcement learning, which learns from environmental feedback, \ac{IL} \citep{hussein2017imitation} learns a policy by mimicking demonstrations provided by an expert. In the context of job shop scheduling, the expert is typically a \ac{CP} solver that produces optimal schedules \citep{lee2022imitation_jssp}.

\subsection{Behavioral Cloning}

In this work, we focus specifically on \ac{BC} \citep{pomerleau1999imitation_learning}, which treats imitation learning as a supervised learning problem. Given a dataset of expert demonstrations\footnote{In our case, the “expert" is a \ac{CP} solver capable of obtaining optimal schedules for small problems.} $\mathcal{D}_E = \{(s_i, a^*_i)\}_{i=1}^N$, where $a^*_i$ is the action chosen by the expert in state $s_i$, \ac{BC} learns a policy $\pi_\theta$ by minimizing the difference between the policy's predictions and the expert's actions:

$$\theta^* = \arg\min_\theta \mathbb{E}_{(s,a^*) \sim \mathcal{D}_E}\left[L(\pi_\theta(s), a^*)\right]$$
\noindent where $L$ is a loss function measuring the dissimilarity between predicted and expert actions. For example, in the experiments made in this project, the GNN will learn to predict what actions are optimal for a given state $s_k$. In other words, it will try to learn a mapping $f: a_k \rightarrow [0, 1]$ for $a_k \in \mathcal{A}(s_k)$, where $\mathcal{A}(s_k)$ is the set of available actions at state $s_k$. The loss function employed in this case is binary cross-entropy:
$$L = -\sum_{O_{ij} \in \mathcal{A}(s_k)} [y_{ij} \log(\hat{y}_{ij}) + (1-y_{ij}) \log(1-\hat{y}_{ij})]$$
where $y_{ij}$ is the true label (0 or 1) for operation $O_{ij}$, $\hat{y}_{ij}$ is the predicted probability that operation $O_{ij}$ is optimal (belongs to class 1). This formula handles both possible outcomes: $L = -\log(\hat{y})$  when $y = 1$, and $L = -\log(1-\hat{y})$ when $y = 0$.

One of the advantages of \ac{BC} over \ac{RL} is that it is more stable and simpler to train. While \ac{RL} requires exploring different approaches, \ac{BC} methods can directly learn from optimal actions. However, \ac{BC} has limitations. It can struggle with distributional shift---situations where the learned policy encounters states not covered in the training data. Additionally, expert demonstrations can be expensive to obtain for large problem instances where computing optimal solutions with exact solvers becomes intractable.

%% file: chapters/05-RelatedWork.tex
\doublespacing 

\chapter{Literature Review and Problem Formulation}
\label{ch5}

\begin{spacing}{1} 
\minitoc 
\end{spacing} 
\thesisspacing 

As we mentioned in the Introduction, one of the core objectives of this project is developing an open-source and flexible \ac{RL} environment for solving the \ac{JSSP} sequentially. However, before describing its design, we need to outline the design space for modeling the \ac{JSSP} as a sequential decision-making problem. For example, one of these choices is the definition of operations available to be dispatched at each step. Some works consider only operations that can start immediately \citep{zhang2020l2d, Park2021l2s}, while others \citep{park2021schedule_net, lee2022imitation_jssp, lee2024il_jssp} allow the GNN dispatcher to reserve operations for later.

Once this design space is outlined, we will be in a position to abstract these choices into our \ac{RL} environment. Following the example above, all the possible definitions of available actions at state $s$ ($\mathcal{A}(s)\subseteq\mathcal{A}$) can be considered as the application of a filter $f$ to the most flexible possible action space definition $\mathcal{A}'(s)$ ($\mathcal{A}(s) = f(\mathcal{A}'(s) ) \subseteq \mathcal{A}'(s)  \subseteq\mathcal{A}$) This will be described in more detail in Section \ref{sec:action_def} of this chapter, and in Subsection \ref{subsec:ready_op_filters} of the next one. 

This broad design space stems from the multiple ways we can define any of the \ac{MDP} components mentioned in the previous chapter. As we mentioned in Section \ref{ch4:mdp}, an \ac{MDP} is a tuple $(\mathcal{S}, \mathcal{A}, P, R, \gamma)$.

Some of these components are constant when solving the classical \ac{JSSP} described in Chapter \ref{ch2}. First, in the case of the classical \ac{JSSP}, the transition probability function $P$ is deterministic and is common for every formulation. Scheduling an available operation on a machine is always successful, and its processing time is fixed. Similarly, the probability distribution of the time spent in a particular state before transitioning to another state $F$ is also deterministic; the transition time $\tau$ between the $k$-th state and the next one is always the same given a state $s_k$ and action $a_k$. 

One customizable component is the discount factor $\gamma$. It is typically set to 1 in the studies that employ \ac{RL} to solve standard \ac{JSSP}s\footnote{It can be set to 1 safely because of the lack of uncertainty in the classical scenario and the episodic nature of the problem. No uncertainty means there is no need to prioritize short-term rewards over long-term ones. Additionally, since the \ac{SMDP} is episodic, there are no convergence issues by setting $\gamma$ to 1.} \citep{zhang2020l2d, Park2021l2s, ho2023residual}. The only exception is \citep{park2021schedule_net}, which sets it to 0.9. However, this is usually considered an additional hyperparameter of the \ac{RL} agent rather than something hardcoded into the environment. It also does not seem to have a considerable effect.

However, previous works differ significantly on the remaining components that define the \ac{JSSP}'s \ac{SMDP}. These components are the state representation, the definition of available actions for each state, and the reward function. The depiction of many of these components can be critical. For instance, the state transition proposed by \cite{Park2021l2s} impedes applications in real-time environments because dispatching an operation can modify the start times of already scheduled ones. 

In this chapter, we analyze the different choices of these elements made by previous papers. This analysis allows us to devise all the features that a customizable \ac{RL} environment for the \ac{JSSP} must have. Additionally, we review other \ac{RL} environments, showing how they fail to meet the identified requirements.

\section{State Representation}
\label{sec:graph_representations}
As mentioned, we focus on graph representations of the environment in this project. Therefore, a state is represented by a dynamic graph $G_k = (V_k, E_k, \mathbf{X}_k)$. This graph definition expands the one presented in Section \ref{sec:disjuntive_graph} by introducing a node feature matrix $\mathbf{X}_k$ and a step index $k$. We use $k$ instead of $t$ to differentiate it from the current time of the schedule, denoted by $t_k$.

In this section, we first look at the main definition variants of $V_k$ and $E_k$. In Subsections \ref{subsec:disjunctive_graphs_for_gnns} and \ref{subsec:resource_task_graphs}, we explore the two main graph variants: disjunctive and resource-task graphs, respectively. In Subsection \ref{subsec:residual_scheduling}, we expand on the different ways that these graphs can be updated after each step by introducing the concept of residual scheduling (introduced in \cite{lee2022imitation_jssp} and \cite{ho2023residual}). Finally, in Subsection \ref{subsec:feature_matrix} we mention some of the features typically embedded into the feature matrix $\mathbf{X}_k$ and the principles that all features should follow, as stated by \cite{ho2023residual} and \cite{lee2024il_jssp}.

\subsection{Disjunctive Graphs}
\label{subsec:disjunctive_graphs_for_gnns}
The first approach for encoding the problem into a graph uses the disjunctive graph without the dummy nodes (source and sink nodes) \citep{Park2021l2s}. After each step, the dynamic information of the schedule (e.g., if an operation has been scheduled) needs to be updated. This information is typically added by removing or adding disjunctive arcs and updating node features representing the state of the operation. The direction of the disjunctive arc represents scheduling decisions. For example, if operation $O_{11}$ is scheduled before $O_{22}$, then there is an arc $O_{11} \rightarrow O_{22}$. See Figure \ref{fig:disjunctive_graph_solved} for an example of a complete schedule represented by the disjunctive graph. In \cite{zhang2020l2d}, two graph updating strategies are defined:
\begin{itemize}[itemsep=0pt, topsep=0pt]
    \item \textbf{``Adding-arc'' strategy:} The graph starts with no disjunctive arcs. After each decision, the corresponding disjunctive arc is added. This is the approach used by \cite{zhang2020l2d}. It has the advantage of creating sparser graphs, which are more computationally efficient for GNN-based processing. However, an important limitation is the loss of valuable information.
    \item \textbf{``Removing-arc'' strategy:} The graph is initialized with disjunctive arcs pointing in both directions, and operation nodes that belong to the same machine form a clique. After an operation is dispatched, one of the two arcs that connected it with the previous operation is dropped. Similarly, the disjunctive arcs associated with the previous scheduled operation are removed. This strategy, followed by \cite{Park2021l2s} and \cite{lee2024il_jssp}, has given better results than the previous one.
\end{itemize}

\subsubsection{Adding Extra Edges}
\label{subsec:adding_extra_edges}
The main limitation of standard disjunctive graphs used by \cite{zhang2020l2d} is that operations of the same job are not well connected. The conjunctive edge always points to future operations (i.e., the one that has to be scheduled after the current one). In other words, the operations in which we are the most interested (the first operations of each job) do not receive messages directly from future operations of the same job. For this reason, artificial conjunctive edges are typically added in other works. For example, in \cite{Park2021l2s}, an additional edge that connects each operation with its consecutive predecessor is considered. Additionally, they consider each of these edges to be of a different type and are processed independently:
$$
\begin{aligned}
h_i^{(l)} &= \text{MLP}_n^{(l)} \left(
    \Biggl[\text{ReLU}\left(\text{MLP}_p^{(l)} \left(\sum_{j \in \mathcal{N}_v^p} h_j^{(l-1)}\right)\right)
    ;\,\text{ReLU}\left(\text{MLP}_s^{(l)} \left(\sum_{j \in \mathcal{N}^s_v)} h_j^{(l-1)}\right)\right)\right.;\,\\[1ex]
    &\qquad\qquad\quad\ \left.\text{ReLU}\left(\text{MLP}_d^{(l)} \left(\sum_{j \in \mathcal{N}^d_v} h_j^{(l-1)}\right)\right)
    ;\,\text{ReLU}\left(\sum_{i \in \mathcal{V}} h_j^{(l-1)}\right)
    ;\,h_i^{(l-1)}
    ;\,h_i^{(0)})\Biggl]
\right)
\end{aligned}
$$
\noindent where $\text{ReLU}(x) = \max(0, x)$, $\mathcal{N}^d_i$, $\mathcal{N}^p_i$, and $\mathcal{N}^s_i$ represent the set of disjunctive, precedent, and successor nodes, respectively; and $[;]$ is the concatenation operation. They also connect each node with all the nodes in the graph, including itself, by aggregating them with a sum. Since this general vector is the same for all nodes, it only needs to be computed once.

The study that best exploits adding extra edges is \cite{lee2024il_jssp}. In addition to considering the aforementioned neighborhoods, they connect each operation with \textit{all} its successors and predecessors to ``effectively deliver information between nodes.''

\subsection{Resource-Task Graphs}
\label{subsec:resource_task_graphs}
\cite{park2021schedule_net} introduced the second type of representation, which \cite{ho2023residual} also used. They added nodes representing the machines and removed the disjunctive edges. While the literature only presents this basic version with machine nodes, it can also be easily extended to include job and global nodes. In these extended variants, conjunctive edges can also be removed. Here, we present three possible variants of resource-task graphs:
\begin{itemize}[itemsep=0pt, topsep=0pt]
    \item \textbf{Basic Resource-Task Graph:} Contains operation nodes and machine nodes. Machines are connected to their respective operations, and machine nodes form a clique (fully connected graph). Operations of the same job are also connected through a structure similar to a machine clique. This is the representation presented in \cite{park2021schedule_net}.
    
    \item \textbf{Resource-Task Graph with Jobs:} Extends the basic version by adding job nodes. In this representation, job nodes are connected to their respective operations, and job nodes also form a clique among themselves. This variant is a novel contribution not found in the literature.
    
    \item \textbf{Complete Resource-Task Graph:} Further extends the representation by adding a global node that connects to all job and machine nodes, serving as an information aggregator. This variant is also a novel contribution.
\end{itemize}

\begin{figure}[!b]
    \makebox[\textwidth][c]{%
    \begin{minipage}{1.2\textwidth}
        \begin{subfigure}[b]{0.32\textwidth}
            \centering
            \includegraphics[height=7cm]{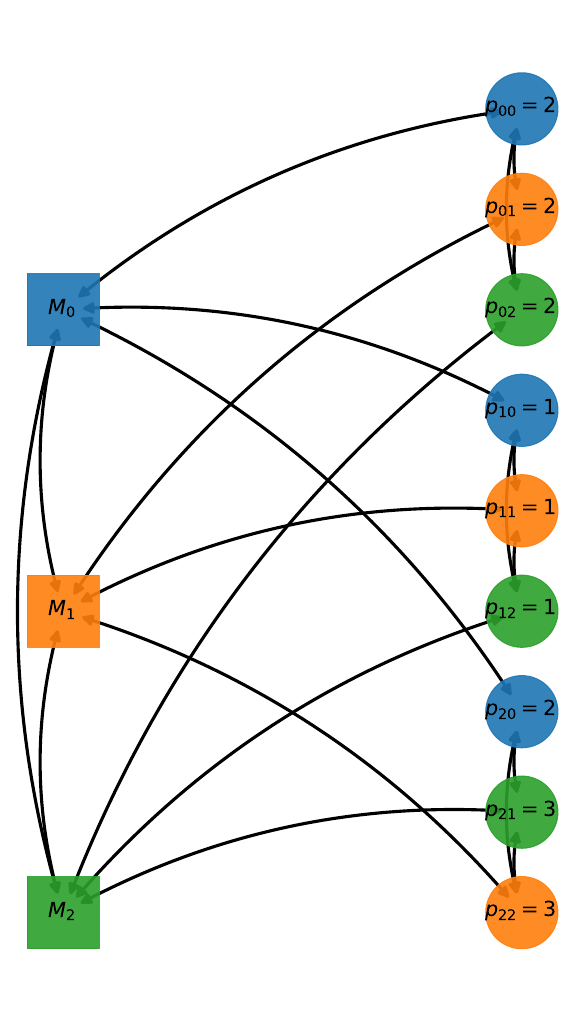}
            \caption{}
            \label{fig:resource_task_graph_example}
        \end{subfigure}
        \hfill
        \begin{subfigure}[b]{0.32\textwidth}
            \centering
            \includegraphics[height=7cm]{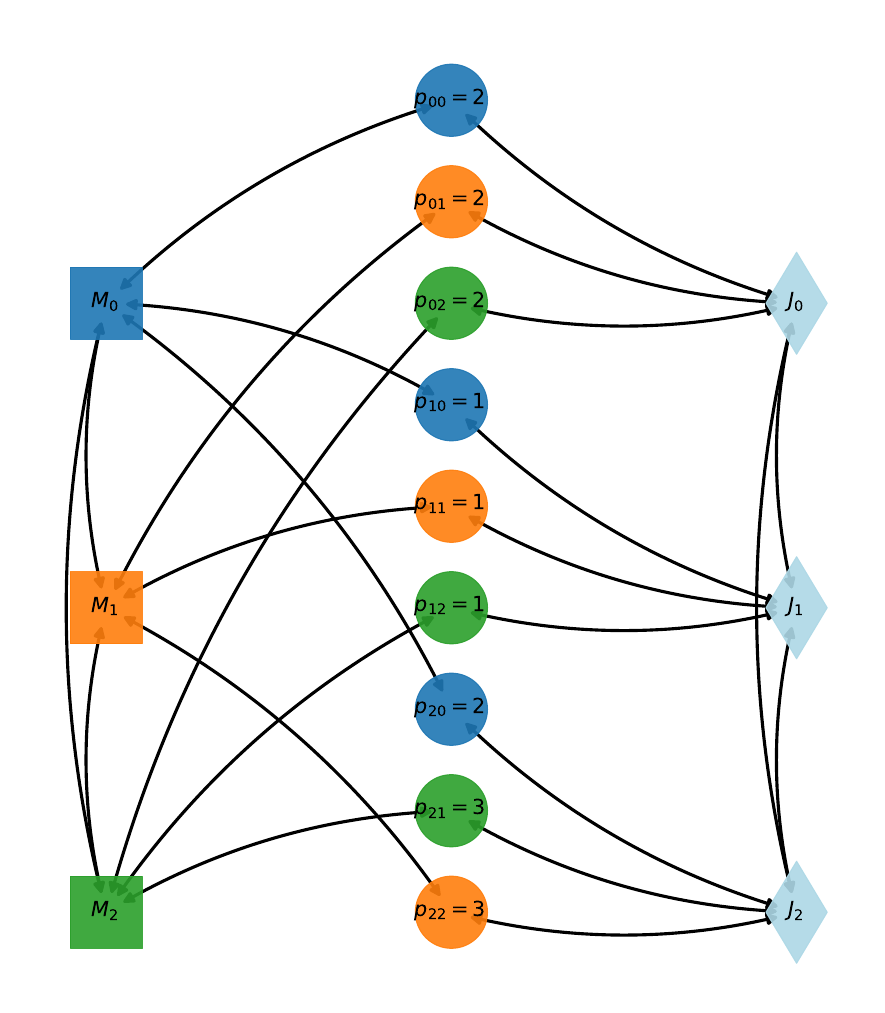}
            \caption{}
            \label{fig:resource_task_graph_with_jobs_example}
        \end{subfigure}
        \hfill
        \begin{subfigure}[b]{0.32\textwidth}
            \centering
            \includegraphics[height=7cm]{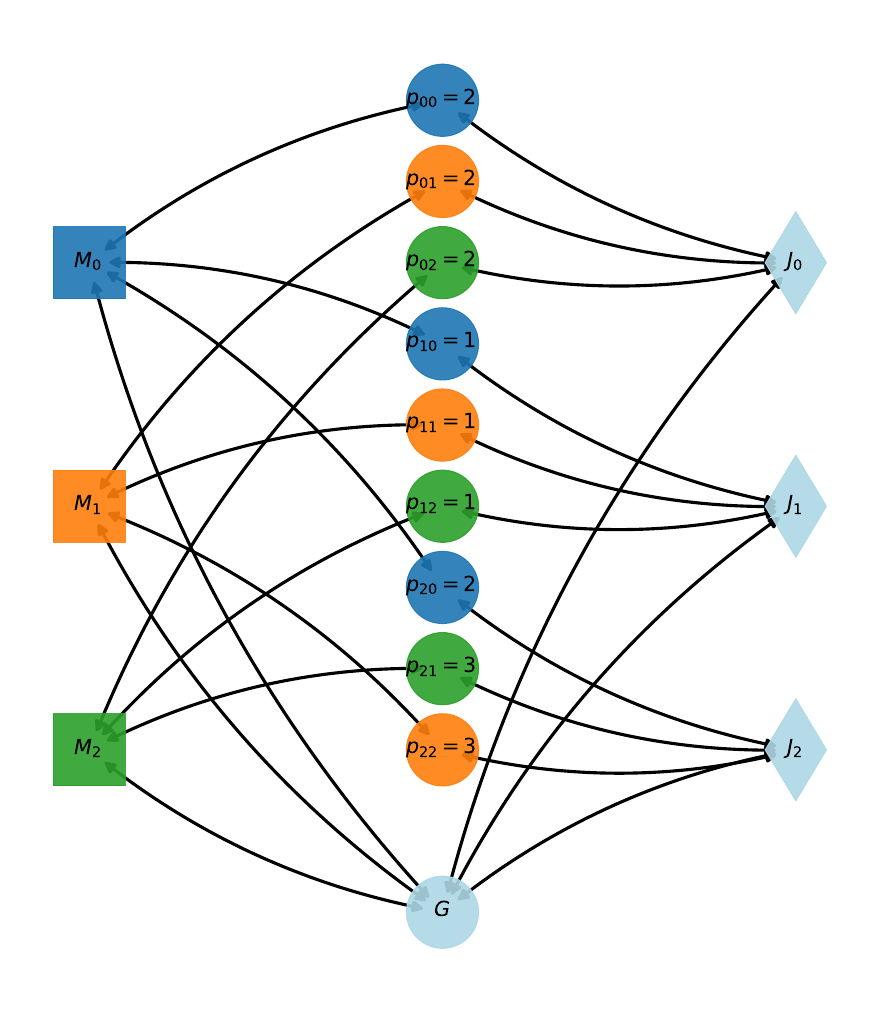}
            \caption{}
            \label{fig:complete_resource_task_graph_example}
        \end{subfigure}
    \end{minipage}
    }
    \caption[Resource-task graph representations of Example \ref{ejemplo1} (Table \ref{tab:ejemplo_introduccion}).]{Resource-task graph representations of Example \ref{ejemplo1} (Table \ref{tab:ejemplo_introduccion}). Square and diamond nodes represent machine and job nodes, respectively. (a) shows the graph representation used by \cite{park2021schedule_net}, and (b) represents how it can be extended with job nodes. In (c), an additional global node (G) removes the cliques formed by machine and job nodes.}
    \label{fig:resource_task_graphs_variants}
\end{figure}
For a problem with $|\mathcal{J}|$ jobs and $|\mathcal{M}|$ machines (resulting in $|\mathcal{J}| \times |\mathcal{M}|$ operations), the resource-task graph approach offers computational advantages over disjunctive graphs for GNN processing. On the one hand, assuming no recirculation of operations, disjunctive graphs have $O(|\mathcal{M}| \times |\mathcal{J}|^2)$ disjunctive edges in the worst case since each machine can have up to $|\mathcal{J}|$ operations, forming a clique with $O(|\mathcal{J}|^2)$ edges, across $|\mathcal{M}|$ machines.

Resource-task graphs reduce this number to $O(|\mathcal{J}| \times |\mathcal{M}|)$ edges between operations and resources, plus additional edges depending on the variant chosen. These additional edges can be $O(|\mathcal{J}|^2) $ and $O(|\mathcal{M}|^2) $ if there are machine or job node cliques, respectively. Even in these cases, however, the complexity is reduced for graphs with $|\mathcal{J}| \leq|\mathcal{M}|^2$. For large job shop problems, reducing edge count can significantly improve computational efficiency while maintaining or enhancing the graph's expressive power by explicitly modeling resource relationships.

However, there are many differences between \cite{park2021schedule_net} and \cite{ho2023residual} with other approaches. These differences include the \ac{SMDP} formulation, the model architecture used, the learning method employed for training, and which extra edges were added to the disjunctive graph. Therefore, the graph representation's impact on performance is unclear.

Additionally, the job and global node variants were introduced theoretically in this work as possible future directions, but their effectiveness has yet to be proved with a rigorous ablation study. For example, the Complete Resource-Task Graph might suffer from ``over-squashing'' \citep{alon2021oversquahing} because all information between machines or jobs must flow through the single global node, creating a potential bottleneck in message passing.

We introduce these variants to note how a flexible \ac{RL} environment should support heterogeneous graph representations with different types of nodes and edges. 

\subsection{Residual Scheduling}
\label{subsec:residual_scheduling}
It should also be noted that an additional mechanism for updating the graph \citep{lee2022imitation_jssp, ho2023residual} has become standard. It consists of removing completed nodes and adjusting their features accordingly. The key idea comes from the Markov property mentioned in Chapter \ref{ch4}: the future state depends only on the current state and action, not on the history of previous ones. For example, the duration of the completed operation that preceded the current available one does not matter when deciding whether to dispatch it. Even the scalar value of the current time is irrelevant. The best schedule will be the same regardless of the hour of the day, for instance. We need to know this information for reconstructing the schedule once we have scheduled each operation, but not for representing the state.

\begin{figure}[H]
    \makebox[\textwidth][c]{%
    \begin{minipage}{1\textwidth}
        \begin{subfigure}[b]{0.48\textwidth}
            \centering
            \includegraphics[width=\textwidth]{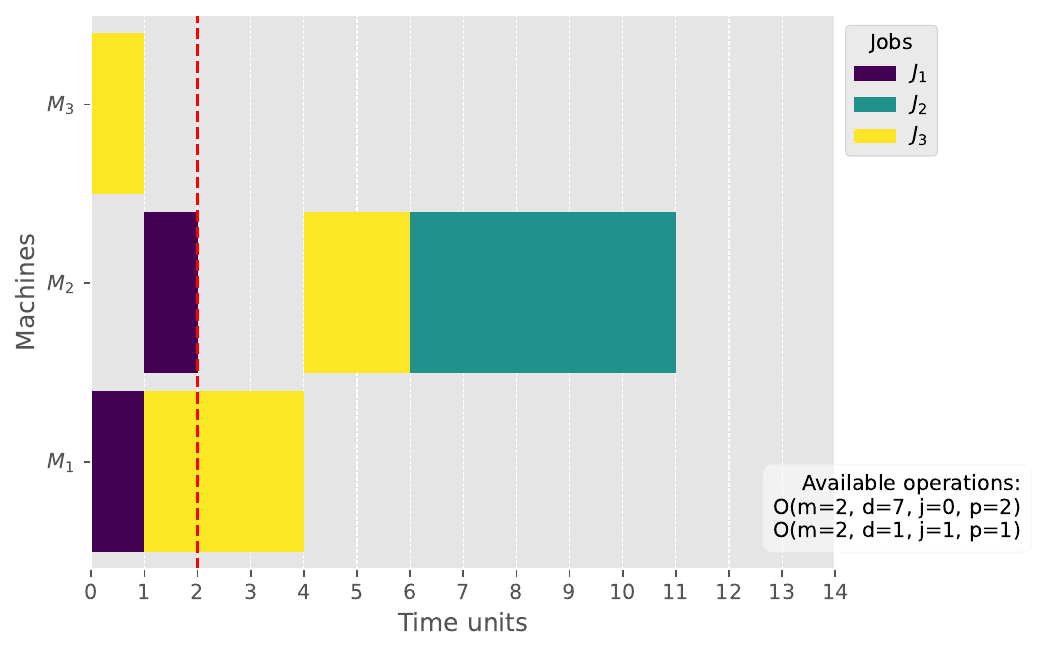}
            \caption{}
            \label{fig:partial_gantt_chart_1}
        \end{subfigure}
        \hfill
        \begin{subfigure}[b]{0.48\textwidth}
            \centering
            \includegraphics[width=\textwidth]{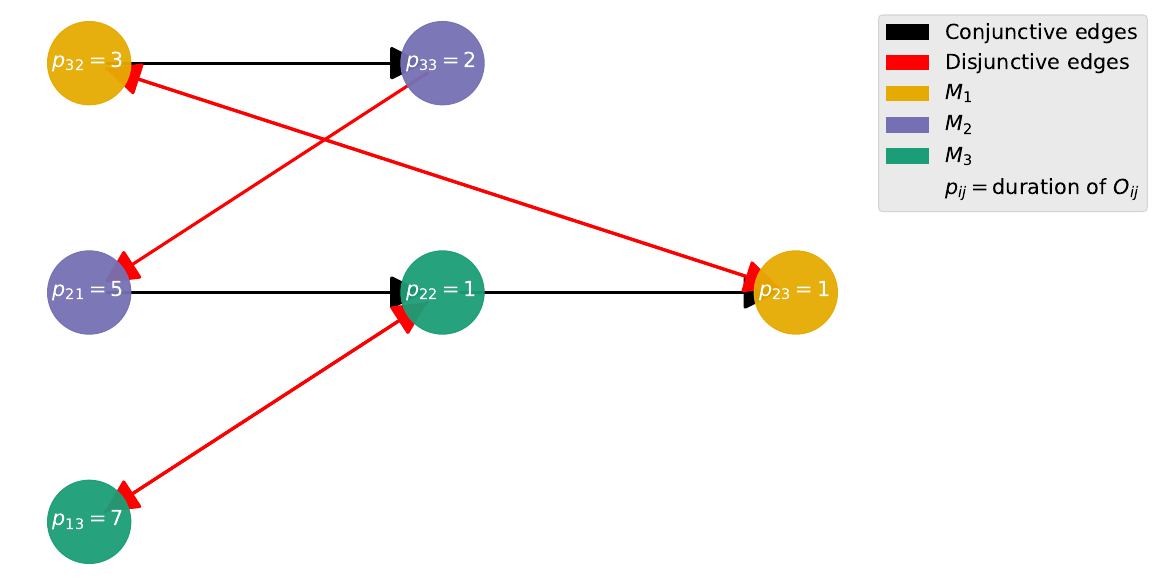}
            \caption{}
            \label{fig:partial_graph_1}
        \end{subfigure}
    \end{minipage}
    }
    \caption[Example of a disjunctive graph (b) representing a partial schedule (a) of the instance defined in Table \ref{tab:ejemplo_introduccion}]{Example of a disjunctive graph (b) representing a partial schedule (a) of the instance defined in Table \ref{tab:ejemplo_introduccion}. Operation $O_{33}$ is scheduled before $O_{21}$ and, thus, has the direction of their disjunctive edge defined. Nodes representing completed nodes have been removed. The red dotted line in (a) represents the current time, defined as the earliest start time of the available operations. These operations are shown in the legend, where ``m'' and ``j'' represent their machine or job id, respectively (starting from zero), ``d'' represents their duration or processing time ($p_{ij}$), and ``p'' represents their position or index in the job (starting from zero).}
    \label{fig:combined_disjunctive_graph_partial}
\end{figure}

Thus, removing completed nodes does not remove relevant information, helps prevent overfitting, and reduces computation during the network's forward and backward passes. Similarly, adjusting the node features creates a more accurate representation of the current state. One example would be updating the scheduled operations' duration to consider only their remaining processing time. For instance, in the state represented in Figure \ref{fig:combined_disjunctive_graph_partial}, the processing time of $O_{32}$ should be adjusted from 3 to 2. In other words, that situation would be equivalent to just starting an operation with a duration of 2 at the current time. This strategy was also employed in \cite{lee2024il_jssp}.

\subsection{The Feature Matrix}
\label{subsec:feature_matrix}
The feature matrix $\mathbf{X}_k$ contains relevant information representing each node. For example, one of the attributes used by all the aforementioned studies is the processing time of each operation $p_{ij}$, or the remaining processing time $p^{(k)}_{ij} = C_{ij} - t_k$ at $s_k$ if using residual scheduling. The work that introduces the most informative features at the time of writing is \citep{lee2024il_jssp}. Some of these features are:
\begin{itemize}[itemsep=0pt, topsep=0pt]
    \item The ready time to start of operation $O_{ij}$ at step $k$, denoted as $R^{(k)}_{ij}$. This feature is the earliest start time $S^{*(k)}_{ij}$ possible that respects the problem constraints adjusted by the current time: $R^{(k)}_{ij} = S^{*(k)}_{ij} - t_k$. This adjustment is made because ongoing operations can be considered just started, as discussed in the previous section.
    \item The machine load $L^{(k)}_{ij}$ of the machine $M_u$ associated with the operation $O_{ij}$. This load is computed as the sum of all the remaining processing times of the operations associated with machine $M_u$. It can then be normalized by the total sum of remaining processing times of the problem:
    $$\sum_{O_{i'j'} \in \mathcal{O}^{(k)}} p^{(k)}_{i'j'}$$
    where $\mathcal{O}^{(k)}$ is the set of uncompleted operations at step $k$).
    \item The number of remaining operations in the job $J_i$ of operation $O_{ij}$.
\end{itemize}

All the features they extracted respected the reasoning behind residual scheduling. Additionally, well-defined features should also respect the symmetry of the problem; they must not depend on the arbitrary indices used to label machines or jobs. Another important consideration is that scheduling decisions should remain constant between states to apply the learned dispatcher in real-time settings. For example, in \cite{zhang2020l2d} they adjust previous scheduling decisions to keep tight schedules. While effective in reducing the makespan, this strategy impedes constructing the schedule sequentially because start times are not fixed.

\section{Action Definition}
\label{sec:action_def}
\begin{figure}[H]
    \makebox[\textwidth][c]{%
    \begin{minipage}{\textwidth}
        \begin{subfigure}[b]{0.48\textwidth}
            \centering
            \includegraphics[height=5.25cm]{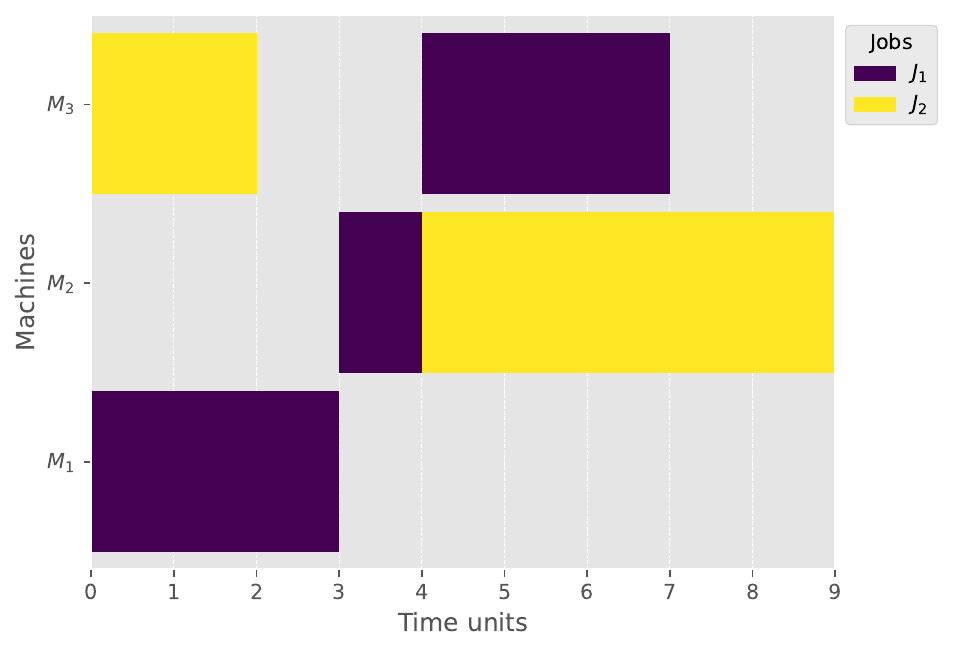}
            \caption{}
            \label{fig:image1}
        \end{subfigure}
        \hfill
        \begin{subfigure}[b]{0.48\textwidth}
            \centering
            \includegraphics[height=5.25cm]{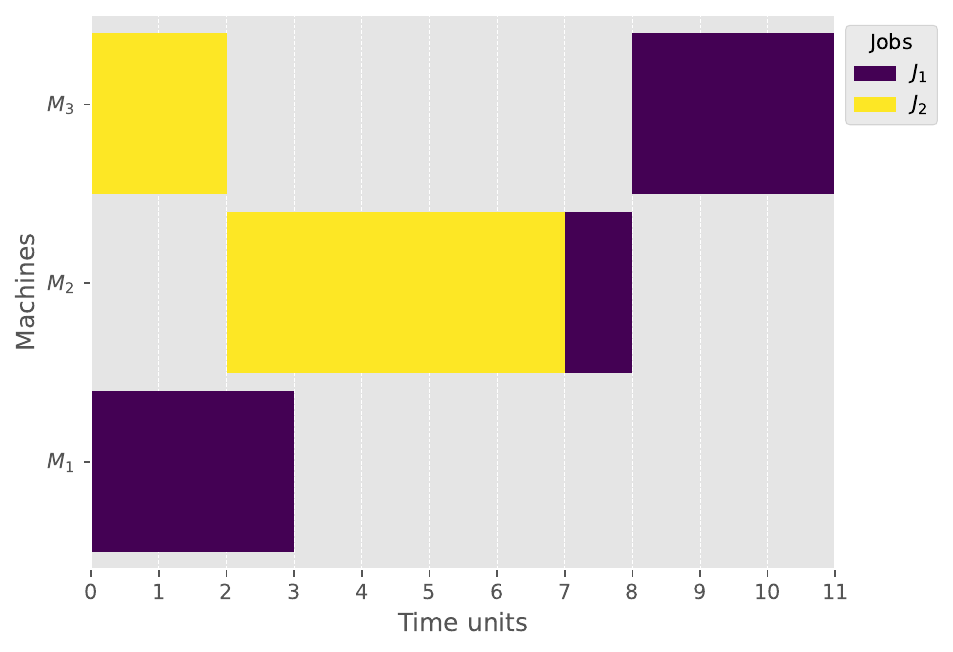}
            \caption{}
            \label{fig:image2}
        \end{subfigure}
    \end{minipage}
    }
    \caption[Comparison of two possible solutions to the instance defined in Table \ref{tab:ejemplo_jssp}.]{Comparison of two possible solutions to the instance defined in Table \ref{tab:ejemplo_jssp}. Both solutions have been obtained by applying the SPT dispatching rule. The difference is that schedule (b) has been obtained by defining available operations as those that can start immediately. In contrast, schedule (a) has been created with a more flexible definition that allows reserving operations. The latter reaches the optimal solution.}
    \label{fig:combined}
\end{figure}

Initial methodologies for utilizing dispatching rules in scheduling, such as those presented by \cite{zhang2020l2d} and \cite{Park2021l2s}, adopted a strategy where only operations that can start at the current time can be selected. They copy the definition employed by dispatching rules defining scheduling decisions ``when a machine becomes available.'' Consequently, this approach is limited to generating solely non-delay schedules as described in Section \ref{sec:representing_solutions}. Let's look at an example to illustrate this concept:

\begin{table}[H]
\begin{center}
\begin{tabular}{c|c|c|c}
\textbf{Job} & \textbf{Operation} & \textbf{Machine} & \textbf{Duration} \\
\hline
& $O_{11}$ & $M_1$ & 3 \\
$J_1$ & $O_{12}$ & $M_2$ & 1 \\
& $O_{13}$ & $M_3$ & 3 \\
\hline
& $O_{21}$ & $M_3$ & 2 \\
$J_2$ & $O_{22}$ & $M_2$ & 5 \\
\end{tabular}
\caption{Example of a \ac{JSSP} problem with two jobs. Operations must be completed in the order presented in the table (from top to bottom).}
\end{center}
\end{table}
\label{tab:ejemplo_jssp}
In the example shown in Table \ref{tab:ejemplo_jssp}, it is impossible to obtain a solution different from the one shown on the left of Figure \ref{fig:combined}, regardless of the dispatching rule used. This situation occurs because, at each step, there is only one available operation. It is impossible to wait for operation $O_{21}$ to be ready; instead, $O_{22}$ must be dispatched immediately, resulting in a less efficient solution.

In contrast, other works \citep{park2021schedule_net, lee2022imitation_jssp, lee2024il_jssp} allow reserving operations. The more flexible definition possible is to define an operation as ready to be dispatched if the previous operation of its job has been scheduled. However, this definition can be improved by removing dominated operations from the available action set \citep{lee2022imitation_jssp}. We consider an operation $O_{ij}$ to be dominated by $O_{i'j'}$ if the latter can be completed before the earliest start time of the first (i.e., if $C_{i'j'} \leq S_{ij}$) and both operations are ready to be processed on the same machine.

Excluding dominated operations acts as a heuristic that limits the space of available actions to operations that can begin immediately. In fact, the available actions definition used by \cite{zhang2020l2d} and \cite{Park2021l2s} can also be seen as a heuristic that filters operations that cannot start immediately.

\section{Reward Functions}
\label{sec:reward_funcs}
Since the objective is to minimize the makespan, some studies have directly used negative makespan as the reward ($-C_{max}$) \citep{park2021schedule_net, infantes2024wheatley}. This reward is defined as zero for every step, except the last one $T$, which has the negative makespan as the reward: $$R(s_{T-1}, a_{T-1}, s_T) = -t_T.$$ Note that we need to change the makespan's sign to negative because the reward is always maximized.

While the objective on which we are focusing in this work is the makespan, simply returning the makespan at the end of an episode would result in a sparse reward. A sparse reward is one that occurs infrequently during an agent's learning process. This is problematic because agents must explore extensively before receiving any meaningful feedback, making it difficult to correlate actions with eventual rewards (the credit assignment problem). This delays learning, decreases sample efficiency, and often leads to suboptimal policies as the agent struggles to identify which actions contributed to success. As noted by \cite{Sutton1998}, dense reward shaping that provides more frequent feedback can help guide exploration and accelerate learning.

The designs of dense reward functions have been very diverse. \cite{zhang2020l2d} define it as the difference in the quality measure between $s_{k}$ and $s_{k+1}$: $$R(s_{k}, a_{k}, s_{k+1}) = H(s_{k}) - H(s_{k+1}).$$
Here, $H(s)$ represents the quality measure of a state $s$, defined as the lower bound of the makespan $C_{max}$ achievable from that state. This lower bound is computed as the maximum estimated completion time considering only precedence constraints for unscheduled operations. With a discount factor of $\gamma=1$, the total cumulative reward equals $H(s_0) - C_{max}$. Since the initial lower bound $H(s_0)$ is constant for a given problem instance, maximizing the cumulative reward corresponds to minimizing the final makespan.

\cite{ho2023residual} simplify this reward by defining it as ``a negative of the additional processing time for the dispatched action.'' However, it is not clear whether it represents: $$R(s_{k}, a_{k}, s_{k+1}) = t_k-t_{k+1}$$ or $$R(s_{k}, a_{k}, s_{k+1}) = C_{\text{max}}^{(k)} - C_{\text{max}}^{(k+1)}.$$ In both interpretations, maximizing the cumulative rewards is equivalent to minimizing the makespan.

On the other hand, other works have aimed to minimize the makespan by minimizing a highly correlated metric. For instance, \cite{Park2021l2s} proposes a ``waiting job reward function," which defines the reward at $R(s_{k}, a_{k}, s_{k+1})$ as the negative number of jobs waiting at step $k$.

Another alternative is to define it based on the ``scheduled area" or idle time \citep{tassel2021rl_env}. After each action, the difference between the duration of the scheduled operations and the introduced idle times is computed. Thus, the scheduled-area-based reward can be defined as: $$R(s_{k}, a_{k}, s_{k+1}) = p_{ij} - \sum_{M \in \mathcal{M}}{\texttt{empty}_{M}}(s_k, s_{k+1})$$ where $p_{ij}$ is the processing time of the operation dispatched with the action $a_k$, and $\texttt{empty}_{M}(s_k, s_{k+1})$ is a function that returns the idle time of machine $M$ while transitioning from $s_k$ to $s_{k+1}$. 

\section{Reinforcement Learning Environments}
\label{sec:rl_envs}

Some open-source \ac{RL} environments to model the \ac{JSSP} already exist. However, they all force users to use a subset of the abovementioned possibilities.
For example, the environment presented by  \cite{tassel2021rl_env} was not designed to support graph representations and the necessary customization. The state is represented by a matrix where each row corresponds to a job and columns represent scaled features like remaining processing time, completion percentage, machine availability time, and idle times. The definition of available operations to schedule is also fixed. Legal actions are masked based on job/machine availability and completion status. The environment incorporates heuristics like ``non-final prioritization'' (prioritizing non-final operations) and rules to restrict the reservation of operations. Moreover, the only reward function supported is the scheduled-area-based one described in the previous section. Another limitation derived from not using graphs is that a different model needs to be created for each instance size.

To the best of our knowledge, the most customizable open-source \ac{RL} environment is being developed by \cite{infantes2024wheatley}. Their implementation allows customizing the reward function and supports \ac{GNNs}. However, some of the critical components we identified previously are non-customizable, including node features, the graph representation, and the definition of available actions. Additionally, modifying the current implementation components is hard due to the lack of documentation.

Other works have also developed \ac{RL} environments to support their experiments \cite{zhang2020l2d, song2023flexible_jssp_drl}. Nevertheless, they only support their design choices (e.g., a specific set of node features or graph representation) as well. Similarly, they lack the modularity and documentation to customize the environment further. Other authors have not made their code public. 

\section{Summary of GNN-Based Dispatchers}
Summarizing the approaches mentioned in the following sections helps to have a better perspective for Chapter \ref{ch8}, in which we train our GNN-based dispatcher using \textit{JobShopLib}'s components.
Table \ref{tab:sota_comparison} shows this analysis. The results of these papers are discussed in Chapter \ref{ch8}.

\begin{table}[H]
\centering
\small
\begin{adjustwidth}{-0.5cm}{}
\begin{tabular}{p{2.4cm}|p{2.2cm}|p{1.8cm}|p{2.2cm}|p{2.8cm}|p{2.2cm}}
\textbf{Work} & \textbf{Graph representation} & \textbf{Residual scheduling} & \textbf{Operation reservation} & \textbf{Learning method} & \textbf{GNN} \\
\hline
\hline
\cite{zhang2020l2d} (L2D) & Disjunctive & No & No (Non-delay) & PPO (\ac{RL}) & GIN \\
\hline
\cite{Park2021l2s} (L2S) & Disjunctive & No & No (Non-delay) & PPO (\ac{RL}) & Custom \\
\hline
\cite{park2021schedule_net} (ScheduleNet) & Resource-Task & No & Yes & Clip-REINFORCE (\ac{RL}) & Type-aware attention \\
\hline
\cite{lee2022imitation_jssp} & Disjunctive & Yes & Yes & \ac{BC} (\ac{IL}) & Custom (similar to \cite{Park2021l2s}) \\
\hline
\cite{ho2023residual} (RS) & Resource-Task & Yes & Yes & REINFORCE (\ac{RL}) & Relational GIN \\
\hline
\cite{infantes2024wheatley} (Wheatley) & Disjunctive & Yes & Yes & PPO (\ac{RL}) & GATv2 + edge attributes \\
\hline
\cite{lee2024il_jssp} & Disjunctive & Yes & Yes & \ac{BC} (\ac{IL}) & Relational GATv2 \\
\end{tabular}
\end{adjustwidth}
\caption{Comparison of GNN-based dispatchers.}
\label{tab:sota_comparison}
\end{table}

%% file: chapters/06-JobShopLib.tex
\doublespacing 

\chapter{Introducing \textit{JobShopLib}}
\label{ch6}

\begin{spacing}{1} 
\minitoc 
\end{spacing} 
\thesisspacing 

A flexible \ac{RL} environment for the \ac{JSSP} should implement many functionalities. For example, it should be able to generate instances randomly, calculate the active or semi-active schedule sequentially, update the graph, or compute node features for each step. To support all these features in a reusable way, we have developed \textit{JobShopLib}. These tasks can be helpful on their own. Therefore, creating this library has the advantage of unlocking many more applications beyond solving the problem with \ac{RL}. For example, this project shows how these components can be easily used to generate an \ac{IL} dataset.

In this chapter, we introduce some of the library's core components and its philosophy. Specifically, all the classes and functions necessary to solve the \ac{JSSP} with \ac{PDRs} are described following a first-principles approach.

We begin by examining the foundational data structures (\texttt{JobShopInstance}, \texttt{Operation}, \texttt{Schedule}, and \texttt{ScheduledOperation}) that form the backbone of the library.

Next, we discuss file formats and benchmark instances, showing how problem instances are stored, retrieved, and shared. Then, we explain how to generate random instances. These features can be used to evaluate and compare different dispatchers, including \ac{PDRs}.

However, we still need to handle the core scheduling logic---which is encapsulated in the \texttt{Dispatcher} class. It determines operation start times while respecting all constraints. This class also allows you to use different ready operation filters that define available operations at each decision point. Subsequently, we explore the Observer pattern implementation that allows components to react to dispatching events without tight coupling.

Finally, we examine the implementation of \ac{PDRs} and how they interface with the \texttt{Dispatcher} through the \texttt{DispatchingRuleSolver} to create complete schedules. Throughout the chapter, we emphasize design choices that favor code reusability and extensibility.

\section{Main Data Structures}
In his famous book \textit{Clean Code} \citep{robert2008clean_code}, Robert C. Martin distinguishes between ``data structures'' and ``objects''\footnote{This use of ``object'' should not be confused with Python's definition, which defines everything, including data structures or even integers, as objects.} as fundamental but opposite abstractions. Data structures expose their data and have no meaningful behavior—they are simply containers that hold information, with public variables and few (or no) methods. Objects, on the other hand, hide their data behind abstractions and expose methods that operate on that data, without revealing how the data is stored or manipulated internally. This difference creates a fundamental dichotomy: Data structures make it easy to add new functions without changing existing classes. In contrast, objects make it easy to add new classes without changing existing methods.

In research, we are not just interested in finding a solution for a particular \ac{JSSP} (the object approach), but in testing different algorithms and comparing them. For this reason, in \textit{JobShopLib}, we have opted to favor the creation of new functions by developing core data structures that are agnostic to the solving method. These main data structures are:

\begin{itemize}
    \item The \href{https://job-shop-lib.readthedocs.io/en/stable/api/job_shop_lib.html#job_shop_lib.JobShopInstance}{\texttt{JobShopInstance}}. It serves as the central data structure for representing \ac{JSSP} instances. It maintains a list of jobs, each composed of an ordered sequence (another list) of operations, along with a name and optional metadata. The class provides numerous properties that derive useful information from the instance, such as matrix representations of operation durations and machine assignments, machine loads, and job durations. These properties are cached for performance when they require expensive computations.

    \item The \href{https://job-shop-lib.readthedocs.io/en/stable/api/job_shop_lib.html#job_shop_lib.Operation}{\texttt{Operation}}. It is the atom of the library. Each operation represents a task with a specific duration that must be processed on one or more machines. In other words, the class supports both classical job shop problems, where each operation can only be processed on one specific machine (the focus of this project), and flexible problems, where operations may have multiple possible machines. Operations maintain references to their containing job through attributes like \texttt{job\_id}, \texttt{position\_in\_job}, and \texttt{operation\_id}, which are typically set by the \texttt{JobShopInstance} class after initialization.

    \item The \href{https://job-shop-lib.readthedocs.io/en/stable/api/job_shop_lib.html#job_shop_lib.Schedule}{\texttt{Schedule}}. It represents a solution (complete or partial) for a particular \ac{JSSP} instance. It stores a list of lists of \texttt{ScheduledOperation} objects, where each list represents the sequence of operations assigned to a specific machine. The class provides methods to calculate the makespan, check solution completeness, and add new \texttt{ScheduledOperations} to the schedule.

    \item The \href{https://job-shop-lib.readthedocs.io/en/stable/api/job_shop_lib.html#job_shop_lib.ScheduledOperation}{\texttt{ScheduledOperation}}. It encapsulates the assignment of an \texttt{Operation} to a specific machine at a particular start time. It maintains references to the original operation while adding the scheduling information (i.e., the start time and assigned machine ID). The class provides properties to derive useful information, such as end time, job ID, and ``position in job,'' from the underlying operation.
\end{itemize}
Note, for instance, that the \texttt{Schedule} class is agnostic to the logic used for setting the start time of a \texttt{ScheduledOperation}. Its \texttt{add} method receives an already initialized \texttt{ScheduledOperation}. This design choice decouples the schedule from the scheduling logic.

Having this logic decoupled also facilitates extending these data structures to support more realistic definitions of the \ac{JSSP}. For example, the library already supports the flexible \ac{JSSP} by allowing an \texttt{Operation} to be completed on multiple machines. Other features, such as setup times for machines or due dates, can be incorporated by extending the original \texttt{Operation} through inheritance or by adding this information into the \texttt{JobShopInstance}'s \texttt{metadata} attribute. For example:
\newpage
\begin{python}[frame=single, breaklines=true]
class CoatingOperation(Operation):
    def __init__(self, machines, duration, color="white", due_date=None):
        super().__init__(machines, duration)
        self.color = color
        self.due_date = due_date
\end{python}
Another feature of these data structures is the use of \texttt{\_\_slots\_\_} for storing their attributes. When a class defines \texttt{\_\_slots\_\_}, it explicitly declares which attributes instances of that class may possess, replacing the default dictionary-based attribute storage mechanism (\texttt{\_\_dict\_\_}) with a more efficient array-like structure. This approach achieves memory savings and reduces attribute access latency by avoiding the overhead associated with hash table implementations required by Python's dictionaries. This reduction in latency is particularly noticeable when solvers need to access these classes' attributes frequently. The \texttt{JobShopInstance} class is the only class that uses the \texttt{\_\_dict\_\_} attribute. The reason is that it needs this dictionary representation to store its cached attributes. 
\begin{figure}[H]
\centering
\includegraphics[width=1.05\textwidth]{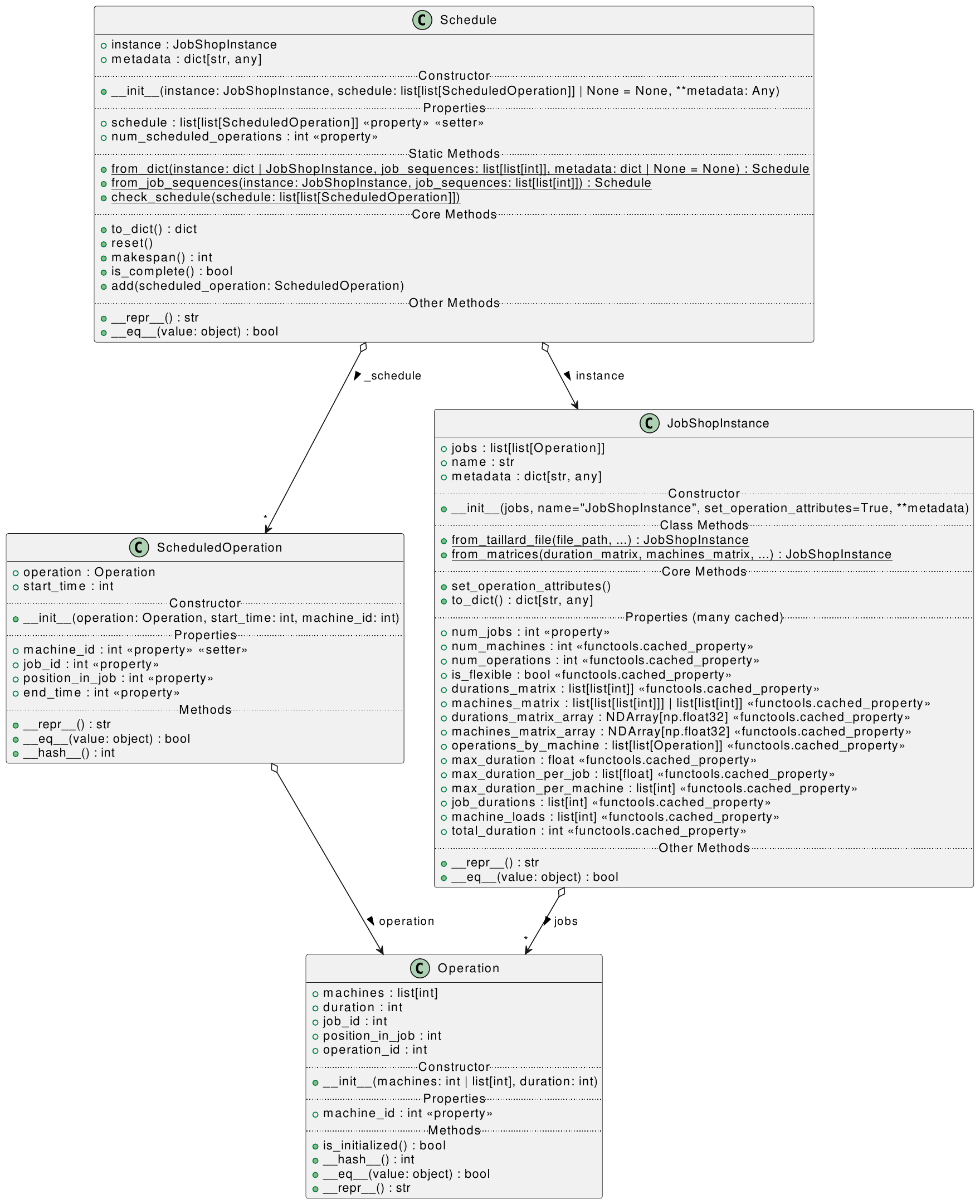}
\caption[Main \textit{JobShopLib}'s data structures.]{Main \textit{JobShopLib}'s data structures. The arguments of \texttt{JobShopInstance}'s \texttt{from\_taillard\_file} and \texttt{from\_matrices} methods have been simplified. They can be seen in \href{https://job-shop-lib.readthedocs.io/en/stable/api/job\_shop\_lib.html\#job\_shop\_lib.JobShopInstance}{\texttt{JobShopInstance}'s documentation}. Private methods and attributes of all the classes are also omitted for simplicity.}
\label{fig:data_structures}
\end{figure}

\section{Reading and Saving JSSP Instances and Solutions}
Comparing the performance of an algorithm to solve the \ac{JSSP} with other methods requires evaluating them with the same set of \ac{JSSP} instances. This set, then, constitutes a benchmark. To create these benchmarks, we need a way to save a \ac{JSSP} instance to a file that other practitioners can read.

\subsection{The Taillard Format}
Classic \ac{JSSP} instances are typically stored using the Taillard file format. It is named after Éric Taillard, who introduced benchmark instances for various scheduling problems in the early 1990s \citep{taillard1993benchmarks}.
The general format consists of a header section and a matrix representation of machines and processing times. The header typically includes the number of machines and jobs. For example, the \ac{JSSP} instance introduced in the introductory chapter (Example \ref{tab:ejemplo_introduccion}) would be represented as:
\begin{python}
# We can also add comments using "#"
3 3
0 2 1 2 2 2
0 1 1 1 2 1
0 2 2 3 1 3
\end{python}
In this case, \texttt{3 3} denotes that there are three jobs and three machines in the scheduling problem. Each subsequent line represents a job, consisting of a sequence of operations, where each operation is defined by a machine-processing time pair. The first number in each pair indicates the machine ID (indexed from 0), and the second number represents the processing time required for that operation. For example, the sequence \texttt{0 2 1 2 2 2} can be interpreted as follows: the first operation uses machine 0 (Cutting machine) for 2 hours, the second operation requires machine 1 (Sanding machine) for 2 hours, and the third operation needs machine 2 (Assembly station) for 2 hours.

\textit{JobShopLib}'s \texttt{JobShopInstance} class provides a class method (\href{https://job-shop-lib.readthedocs.io/en/stable/api/job\_shop\_lib.html#job_shop\_lib.JobShopInstance.from\_taillard\_file}{\texttt{from\_taillard\_file}}) to load an instance from this type of file.

\subsection{\textit{JobShopLib}'s JSON-Based Format}
Typically, benchmark instances are accompanied by metadata, such as lower and upper bounds or the optimum makespan if known. Parsing this information from the header format of a Taillard file can be cumbersome because there are no standard ``keys'' for this type of information. Additionally, while Taillard's format allows us to store \ac{JSSP} instances, it does not support storing solutions nor \ac{JSSP} variants.

In this project, we argue that the JSON format—introduced in the early 2000s \citep{crockford-jsonorg2006json}, a decade after the Taillard file—constitutes a more flexible and easy-to-use alternative for storing both \ac{JSSP}s and schedules. JSON files can be directly mapped to Python objects. To save an instance to a JSON file, we can first create its dictionary representation through \texttt{JobShopInstance}'s \href{https://job-shop-lib.readthedocs.io/en/stable/api/job_shop_lib.html#job_shop_lib.JobShopInstance.to_dict}{\texttt{to\_dict}} method:
\begin{python}[frame=single]
class JobShopInstance:
    def to_dict(self) -> dict[str, Any]:
        return {
            "name": self.name,
            "duration_matrix": self.durations_matrix,
            "machines_matrix": self.machines_matrix,
            "metadata": self.metadata,
        }
\end{python}
The \texttt{"duration\_matrix"} and \texttt{"machines\_matrix"} contain the same information that was intertwined in the above Taillard's matrix. The \texttt{"metadata"} field contains a dictionary with arbitrary information about the instance, such as lower bounds or the best-known makespan. 

Similarly, we can create a dictionary representation of a schedule using the same philosophy. Its dictionary representation has the following keys:

\begin{itemize}
    \item \textbf{\texttt{"instance"}}: The dictionary representation of the instance mentioned before.
    \item \textbf{\texttt{"job\_sequences"}}: A list of lists of job IDs. Each list
    The number of job IDs represents the order in which operations are processed on the machine. The implicit index of the list determines the machine to which the list corresponds. This attribute is the matrix representation $\mathbf{Y}$ introduced in Chapter \ref{ch2}.
    \item \textbf{\texttt{"metadata"}}: A dictionary with additional information
    about the schedule (e.g., the solver used, or the time required to reach that solution).
\end{itemize}
Having these dictionary representations allows us to store these data structures efficiently in a JSON file. Additionally, this format is generally well-understood and known among computer scientists. Therefore, the learning curve for switching from the previous Taillard standard to this one is minimal.

\subsection{Benchmark Instances}
\label{subsec:benchmark_instances}
Using this storage mechanism, \textit{JobShopLib} provides access to a custom benchmark dataset with the following instances:
\begin{table}[H]
\centering
\begin{tabular}{l|c|c|c}
\textbf{Benchmark Instances} & \textbf{\# Instances} & \textbf{Min size} & \textbf{Max size} \\
\hline
\textbf{\texttt{ta01-80}} \citep{taillard1993benchmarks} & 80 & $15 \times 15$ & $100 \times 20$ \\
\textbf{\texttt{abz5-9}} \citep{adams1988shifting_bottleneck} & 5 & $10 \times 10$ & $20 \times 15$ \\
\textbf{\texttt{ft06, ft10, ft20}} \citep{muth1963industrial_scheduling} & 3 & $6 \times 6$ & $20 \times 5$ \\
\textbf{\texttt{orb01-10}} \citep{applegate1991computational_study} & 10 & $10 \times 10$ & $10 \times 10$ \\
\textbf{\texttt{yn1-4}} \citep{yamada1992genetic_algorithm} & 4 & $20 \times 20$ & $20 \times 20$ \\
\textbf{\texttt{swv01-20}} \citep{storer1992new_search_spaces} & 20 & $20 \times 10$ & $50 \times 10$ \\
\textbf{\texttt{la01-40}} \citep{lawrence1984resource_constrained} & 40 & $10 \times 5$ & $30 \times 10$ \\
\end{tabular}
\caption{Benchmark instances available in \textit{JobShopLib}}
\label{tab:benchmarks_overview}
\end{table}
This dataset can be accessed through the \href{https://job-shop-lib.readthedocs.io/en/stable/api/job_shop_lib.benchmarking.html#module-job_shop_lib.benchmarking}{\texttt{benchmarking}} module without needing to download additional files; the JSON file containing the instances is imported when installing the library. For example, to get a single \ac{JSSP} instance, we can use the following code:
\begin{python}[frame=single]
from job_shop_lib.benchmarking import load_benchmark_instance

ft06 = load_benchmark_instance("ft06")
\end{python}
These instances contain relevant metadata such as lower and upper bounds and a reference to the article that introduced them:
\newpage
\begin{python}[frame=single, breaklines=true]
>>> ft06.metadata
{'optimum': 55,
 'upper_bound': 55,
 'lower_bound': 55,
 'reference': "J.F. Muth, G.L. Thompson. 'Industrial scheduling.', Englewood Cliffs, NJ, Prentice-Hall, 1963."}
\end{python}
This information has been obtained from \cite{Weise2025JSSP_Instances_Results}.

\section{Generating Random Instances}
\begin{sloppypar}
These benchmark instances were generated using random procedures. In \texttt{JobShopLib}, we provide an abstraction (\href{https://job-shop-lib.readthedocs.io/en/stable/api/job_shop_lib.generation.html#job_shop_lib.generation.InstanceGenerator}{\texttt{InstanceGenerator}}) and a built-in class to generate random instances (\href{https://job-shop-lib.readthedocs.io/en/stable/api/job_shop_lib.generation.html#job_shop_lib.generation.GeneralInstanceGenerator}{\texttt{GeneralInstanceGenerator}}).
\end{sloppypar}

The \texttt{InstanceGenerator} serves as an abstract base class that defines a common interface for all instance generators in \textit{JobShopLib}. This abstraction ensures consistency and allows for different generation strategies to be implemented and used interchangeably. In particular, \texttt{InstanceGenerator} implements the iterator protocol, facilitating the generation of multiple instances, for example, within loops. Subclasses are expected to implement the core \texttt{generate} method, which is responsible for producing a single \texttt{JobShopInstance}.

\begin{sloppypar}
The library provides a versatile, built-in implementation: the \href{[https://job-shop-lib.readthedocs.io/en/stable/api/job_shop_lib.generation.html#job_shop_lib.generation.GeneralInstanceGenerator](https://job-shop-lib.readthedocs.io/en/stable/api/job_shop_lib.generation.html#job_shop_lib.generation.GeneralInstanceGenerator)}{\texttt{GeneralInstanceGenerator}}. This class is designed to cover a wide range of standard random instance generation needs, including Taillard's methodology \citep{taillard1993benchmarks}. It operates based on specified ranges for the number of jobs (\texttt{num\_jobs\_range}) and machines (\texttt{num\_machines\_range}), as well as a range for operation durations (\texttt{duration\_range}).

When the \texttt{generate()} method is called (either directly or through iteration), \texttt{{GeneralInstanceGenerator}} first determines the exact number of jobs and machines for the instance, typically by randomly selecting values within the provided ranges. It then proceeds to:
\end{sloppypar}
\begin{sloppypar}
\begin{enumerate}
    \item \textbf{Generate the duration matrix}: It calls a utility function (e.g., \texttt{{generate\_duration\_matrix}}) that creates a matrix where each element $(i, j)$ represents the duration of the $j$-th operation of the $i$-th job. Durations are typically sampled uniformly from the specified \texttt{duration\_range}.
    \item \textbf{Generate the machine matrix}: \textit{JobShopLib} supports two common strategies for machine assignment through helper functions like \texttt{generate\_machine\_matrix\_with\_recirculation} and \texttt{generate\_machine\_matrix\_without\_recirculation}.
    \begin{itemize}
        \item Without recirculation: Each job visits each machine exactly once, ensuring a permutation of machines for each job's operations (as seen in many classic benchmarks like Taillard's).
        \item With recirculation: Operations of a job can be assigned to the same machine multiple times.
    \end{itemize}
    \item \textbf{Instantiate \texttt{JobShopInstance}}: Finally, it uses the generated duration and machine matrices, along with an auto-incrementing name (based on the \texttt{name\_suffix} attribute), to create and return a \texttt{JobShopInstance} object using the \texttt{from\_matrices()} class method.
\end{enumerate}    
\end{sloppypar}
The generator also includes flags, such as \texttt{allow\_less\_jobs\_than\_machines}, to control specific constraints during generation, adding to its flexibility. Reproducibility is ensured through an optional \texttt{seed} parameter.

Here is an example of how to use the \texttt{GeneralInstanceGenerator} to create a random 3x3 instance with durations between 1 and 10, and no recirculation:

\begin{python}[frame=single]
from job_shop_lib.generation import GeneralInstanceGenerator

generator = GeneralInstanceGenerator(
    num_jobs=3, # Fixed number of jobs
    num_machines=3, # Fixed number of machines
    duration_range=(1, 10),
    allow_recirculation=False, # Ensure each job visits each machine once
    iteration_limit=10,  # Stop after 10 instances if using it as an iterator
    seed=42,
)
random_instance = generator.generate()
\end{python}
Now, we can check that the instance was indeed created following the given indications:
\begin{python}[frame=single]
>>> random_instance
JobShopInstance(name=classic_generated_instance_1, num_jobs=3, num_machines=3)
>>> random_instance.duration_matrix_array
array([[1., 8., 7.],
       [5., 5., 9.],
       [1., 7., 3.]], dtype=float32)
>>> random_instance.machines_matrix_array
array([[2., 0., 1.],
       [0., 2., 1.],
       [2., 1., 0.]], dtype=float32)
\end{python}
As mentioned, we can also use a \texttt{for} loop to generate one instance per iteration:
\begin{python}[frame=single]
instances = []
for instance in generator:
    instances.append(instance)
>>> len(instances)
10
\end{python}
Or simply:
\begin{python}[frame=single]
>>> len(list(generator))
10
\end{python}
This example is available at \href{https://colab.research.google.com/drive/1BcZLeE4Cgw4Rwp-JFSiHW8P_Trk9ncDN?usp=sharing}{Google colab}.

\section{The \texttt{Dispatcher} Class}
Once we have the necessary data structures to store both instances and results, we can start developing solvers that use them. To solve the \ac{JSSP} using dispatching rules, we first need an engine able to compute the start time of a scheduled operation given the current state (remember that the \texttt{Schedule} class is agnostic to this).

The object responsible for this behavior is the \href{https://job-shop-lib.readthedocs.io/en/stable/api/job_shop_lib.dispatching.html#job_shop_lib.dispatching.Dispatcher}{\texttt{Dispatcher}} class, defined within the \texttt{dispatching} module. Its primary role is to determine the earliest possible start time for the dispatched operation, respecting the problem's constraints: precedence constraints within a job and resource constraints on the machines. This ensures that all the schedules generated are, at least, semi-active.

The core method is \href{https://job-shop-lib.readthedocs.io/en/stable/api/job_shop_lib.dispatching.html#job_shop_lib.dispatching.Dispatcher.dispatch}{\texttt{dispatch}}. It accepts an \texttt{Operation} object and calculates its earliest possible start time. This start time must be no earlier than the completion time of the preceding operation in the same job and no earlier than the time the required machine becomes available. Once calculated, it creates and returns a \texttt{ScheduledOperation} containing the operation, its assigned machine, and the computed start time. This \texttt{ScheduledOperation} is then added to the main \texttt{Schedule} object.

\subsection{Ready Operations Filters}
\label{subsec:ready_op_filters}
\begin{sloppypar}
Besides the core \texttt{dispatch} method, the \texttt{Dispatcher} offers several helper methods to query the current state of the scheduling process. These include methods like \texttt{current\_time}, \texttt{scheduled\_operations}, \texttt{unscheduled\_operations}, \texttt{ongoing\_operations}, and \texttt{completed\_operations}. They can be useful, for example, for computing later node features from the current state.
\end{sloppypar}

However, \href{https://job-shop-lib.readthedocs.io/en/stable/api/job\_shop\_lib.dispatching.html\#job\_shop\_lib.dispatching.Dispatcher.available\_operations}{\texttt{available\_operations}} is arguably the most critical method for implementing different scheduling approaches. It determines the set of operations from which a dispatching rule can choose to dispatch next.

Internally, \texttt{available\_operations} first identifies all operations whose predecessors in their respective jobs have already been completed. This basic set of potentially schedulable operations can be accessed directly via the \texttt{raw\_ready\_operations} method. These operations correspond to the most flexible definition mentioned in Chapter \ref{ch5}, where an operation is ready simply because its job sequence allows it.
\begin{sloppypar}
Crucially, if a \texttt{ready\_operations\_filter} function was provided when the \texttt{Dispatcher} was initialized, \texttt{available\_operations} applies this filter to the raw list of ready operations. This filter function takes the \texttt{Dispatcher} instance and the list of raw ready operations and returns a potentially smaller list. This filtering step is where different definitions of ``available operations'' are implemented:

\begin{itemize}
    \item \textbf{No Filter:} If no filter is provided, \texttt{available\_operations()} returns the same list as \texttt{raw\_ready\_operations()}. This definition gives the dispatcher the maximum freedom. Still, it requires the decision-making logic (the dispatching rule) to handle potentially more complex choices, including reserving operations.
    \item \href{https://job-shop-lib.readthedocs.io/en/stable/api/job_shop_lib.dispatching.html#job_shop_lib.dispatching.filter_non_immediate_machines}{\textbf{\texttt{filter\_non\_immediate\_operations}}} (NIO): This filter removes any operation whose earliest possible start time is later than the current time (the minimum earliest start time among all raw ready operations). In essence, it keeps only those operations that can start right now. It directly implements the non-delay scheduling strategy discussed in Chapter \ref{ch5} and used by works like \cite{zhang2020l2d} and \cite{Park2021l2s}. As shown in Figure \ref{fig:combined}, this can restrict choices and potentially lead to suboptimal schedules because it forces an operation to be scheduled if its machine is free and it can start immediately, even if waiting might be better. However, it can also be a good heuristic that removes suboptimal choices. This is the definition classically employed when defining \ac{PDRs} \citep{haupt1989survey_pdr}.
    \item \href{https://job-shop-lib.readthedocs.io/en/stable/api/job_shop_lib.dispatching.html#job_shop_lib.dispatching.filter_dominated_operations}{\textbf{\texttt{filter\_dominated\_operations}}} (DO): This filter implements the concept described by \cite{lee2022imitation_jssp}. It removes an operation if another operation exists that requires the same machine and can finish before the first operation can even start. In other words, it removes operations that are dominated by at least another one. This heuristic prunes the set of available operations, eliminating suboptimal choices without restricting the schedule to be strictly non-delayed. Thus, it ensures generating active schedules at least.
    
    Although this definition always removes suboptimal choices, it should be noted that this does not mean that better choices will always be made after applying this restriction. For instance, let's suppose that the priority that a dispatching rule assigns to operations A, B, and C is: $\text{A} > \text{B} > \text{C}$. At the same time, the optimal preferences in this example should be: $\text{C} > \text{A} > \text{B}$ (i.e., dispatching C will result in a lower makespan than dispatching A, and so on). If C dominates A, then A will be removed from the set of available actions by this filter. However, now the PDR will select to dispatch B instead of A, resulting in a worse schedule.
    \item \textbf{Other Filters:} The library also provides filters like \href{https://job-shop-lib.readthedocs.io/en/stable/api/job_shop_lib.dispatching.html#job_shop_lib.dispatching.filter_non_idle_machines}{\texttt{filter\_non\_idle\_machines}} (NIDM), which removes operations whose machines are currently busy; and \href{https://job-shop-lib.readthedocs.io/en/stable/api/job_shop_lib.dispatching.html#job_shop_lib.dispatching.filter_non_immediate_machines}{\texttt{filter\_non\_immediate\_machines}} (NIM), which keeps only operations assignable to machines where at least one operation can start immediately. These offer alternative heuristics to prune the action space. Filters can also be combined using \href{https://job-shop-lib.readthedocs.io/en/stable/api/job_shop_lib.dispatching.html#job_shop_lib.dispatching.create_composite_operation_filter}{\texttt{create\_composite\_operation\_filter}}, which applies the given filters sequentially.
\end{itemize}
\end{sloppypar}
As discussed, we cannot be certain of which definition will result in better performance in practice. To test this aspect, we ran an experiment to evaluate the average performance of the basic \ac{PDRs} discussed in Subsection \ref{ch2:pdr} on the aforementioned benchmark dataset. Table \ref{tab:filter_impact} shows these results: 

\begin{table}[H]
   \centering
   \begin{tabular}{l|c}
       \textbf{Filters} & \textbf{Makespan Improvement (\%)} \\
       \hline
       DO & 37.41  \\
       NIM & 46.44 \\
       NIDM & 42.94  \\
       NIO (classical definition) & \textbf{53.85} \\
       DO, NIM & 50.56 \\
       DO, NIDM & 50.03 \\
   \end{tabular}
   \caption[Impact of different filter combinations on PDR performance showing improvement over using no filter.]{Impact of different filter combinations on PDR performance, showing improvement over using no filter. The combination of some filters is not shown because it would be redundant. For example, the NIO filter already filters out dominated operations (DO), since it keeps only those that can start immediately. Similarly, all the machines that have at least one operation that can start immediately (NIM) are also idle (NIDM). The code of the experiment is available at \href{https://colab.research.google.com/drive/1UDRLSfCv51m2gCcGXXS97aE7rRuP-lLq?usp=sharing}{Google colab}.}
   \end{table}
   \label{tab:filter_impact}
These results show that the best 'available operations' definition when using these basic dispatching rules is NIO (i.e., considering operations that can start immediately exclusively). In other words, when using these basic dispatchers, reserving operations with the hope of finding a better schedule is generally a bad strategy. Interestingly, it seems that the more we restrict choices, the better performance we get when applying basic dispatching rules.

However, whether a GNN-based dispatcher would benefit from restrictions in the action space is still unclear. The more intelligent this dispatcher is, the more it will distinguish those states in which reserving operations is a suboptimal strategy. Similarly, the model could rapidly learn these heuristics (e.g., always dispatching immediate operations), which would reduce the benefit from applying the corresponding filter. The results of the experiments that test this idea (and more) are presented in Chapter \ref{ch8}.

Users can also define their own filters. However, it is important to note that a filter can alter the current time $t_k$ computation since it is defined as the earliest start time of the \textit{available} operations. Thus, preserving the property $t_{k+1} \geq t_k$ is recommended when designing filters. For example, a filter that only considers operations belonging to the machine with the lowest ID could cause the current time to move backward. In particular, this could happen if there is an operation available at $M_1$ with start time $S_1 = t_k$, and in the next step, only operations of another machine are available. The lowest start time of these operations $S_2$ will be the new $t_{k+1}$, but there is no guarantee that $S_1 \leq S_2$.

\subsection{The \texttt{DispatcherObserver} Pattern}
\label{subsec:dispatcher_observer}

As the \texttt{Dispatcher} dispatches operations, its internal state changes. Several components might need to react to these state changes. For instance:
\begin{itemize}
    \item A logging mechanism might want to record each dispatching decision.
    \item A feature extractor for a machine learning model needs to update its representation of the system state after each step.
    \item A visualization tool might want to update a Gantt chart incrementally.
\end{itemize}
Implementing the logic for all these potential reactions directly within the \texttt{Dispatcher} class would make it overly complex, tightly coupled to specific functionalities, and difficult to extend.

To address this, \textit{JobShopLib} employs the Observer design pattern \citep{gamma1995design_patterns}. In this pattern, the \texttt{Dispatcher} acts as the ``Subject'' (or ``Observable''), and other objects that need to react to its changes act as ``Observers''. The \texttt{Dispatcher} maintains a list of its subscribed observers and notifies them automatically whenever a relevant event occurs, primarily when a new operation is dispatched.

The core of this implementation is the abstract base class \href{https://job-shop-lib.readthedocs.io/en/stable/api/job_shop_lib.dispatching.html#job_shop_lib.dispatching.DispatcherObserver}{\texttt{DispatcherObserver}}. Any class that needs to observe a \texttt{Dispatcher} must inherit from \texttt{DispatcherObserver} and implement its abstract methods:
\begin{itemize}
    \item \texttt{update()}: It is called by the \texttt{Dispatcher} immediately after an operation has been successfully dispatched and added to the schedule. The observer receives the newly \texttt{ScheduledOperation} as an argument, allowing it to react accordingly.
    \item \texttt{reset()}: It is called when the associated \texttt{Dispatcher}'s \texttt{reset()} method is invoked. This allows the observer to clear its internal state and synchronize with the reset state of the \texttt{Dispatcher}.
\end{itemize}
Each \texttt{DispatcherObserver} instance holds a reference to the \texttt{Dispatcher} it observes (the \texttt{dispatcher} attribute).

This pattern decouples the \texttt{Dispatcher} from the specific logic of the components that depend on its state. The \texttt{Dispatcher} only needs to know about the \texttt{DispatcherObserver} interface, not the concrete observer classes.
\begin{sloppypar}
\textit{JobShopLib} provides several built-in observers leveraging this pattern. Those available within the \texttt{dispatching} module are:
\begin{itemize}
    \item \href{https://job-shop-lib.readthedocs.io/en/stable/api/job_shop_lib.dispatching.html#job_shop_lib.dispatching.HistoryObserver}{\texttt{HistoryObserver}}: Records the sequence of \texttt{ScheduledOperation} objects as they are dispatched. This observer can be used by some visualization functions (e.g., \href{https://job-shop-lib.readthedocs.io/en/stable/api/job_shop_lib.visualization.gantt.html#job_shop_lib.visualization.gantt.create_gantt_chart_video}{\texttt{create\_gantt\_chart\_video}}) to reproduce the exact sequence of actions taken to create a schedule. 
    \item \href{https://job-shop-lib.readthedocs.io/en/stable/api/job_shop_lib.dispatching.html#job_shop_lib.dispatching.UnscheduledOperationsObserver}{\texttt{UnscheduledOperationsObserver}}: Efficiently maintains the set of operations that have not yet been scheduled, updating itself each time an operation is dispatched. While the \texttt{Dispatcher} class already has a cached \texttt{unscheduled\_operations} method, this observer shows how this pattern can also be more efficient than these methods. \texttt{Dispatcher}'s \texttt{unscheduled\_operations()} creates the list of unscheduled operations after each step. This observer is more efficient because it only needs to update the previous set of unscheduled operations. For a performance comparison, see \href{https://job-shop-lib.readthedocs.io/en/stable/examples/11-UnscheduledOperationsObserver.html}{this example} from \textit{JobShopLib}'s documentation.
\end{itemize}
\end{sloppypar}
In the next chapter, we introduce the observers that extract relevant node features for a GNN-based dispatcher. 

Users can easily implement custom observers for specific needs, such as feature extraction, logging, or visualization, without modifying the core \texttt{Dispatcher} logic. For example, a simple custom observer to print dispatched operations:
\newpage
\begin{python}[frame=single]
from job_shop_lib.dispatching import DispatcherObserver, Dispatcher

class PrintObserver(DispatcherObserver):
    def update(self, scheduled_operation):
        op = scheduled_operation.operation
        print(f"Time {self.dispatcher.current_time()}: "
              f"Dispatched Op({op.job_id},{op.position_in_job}) on"
              f"Machine {scheduled_operation.assigned_machine_id} "
              f"at time {scheduled_operation.start_time}")

    def reset(self):
        print("Observer reset.")
\end{python}
With this class, whenever its \texttt{dispatcher.dispatch} is called, the \texttt{PrintObserver}'s update method will be invoked and the specified message will be printed.

\begin{figure}[H]
\centering
\includegraphics[width=1.15\textwidth]{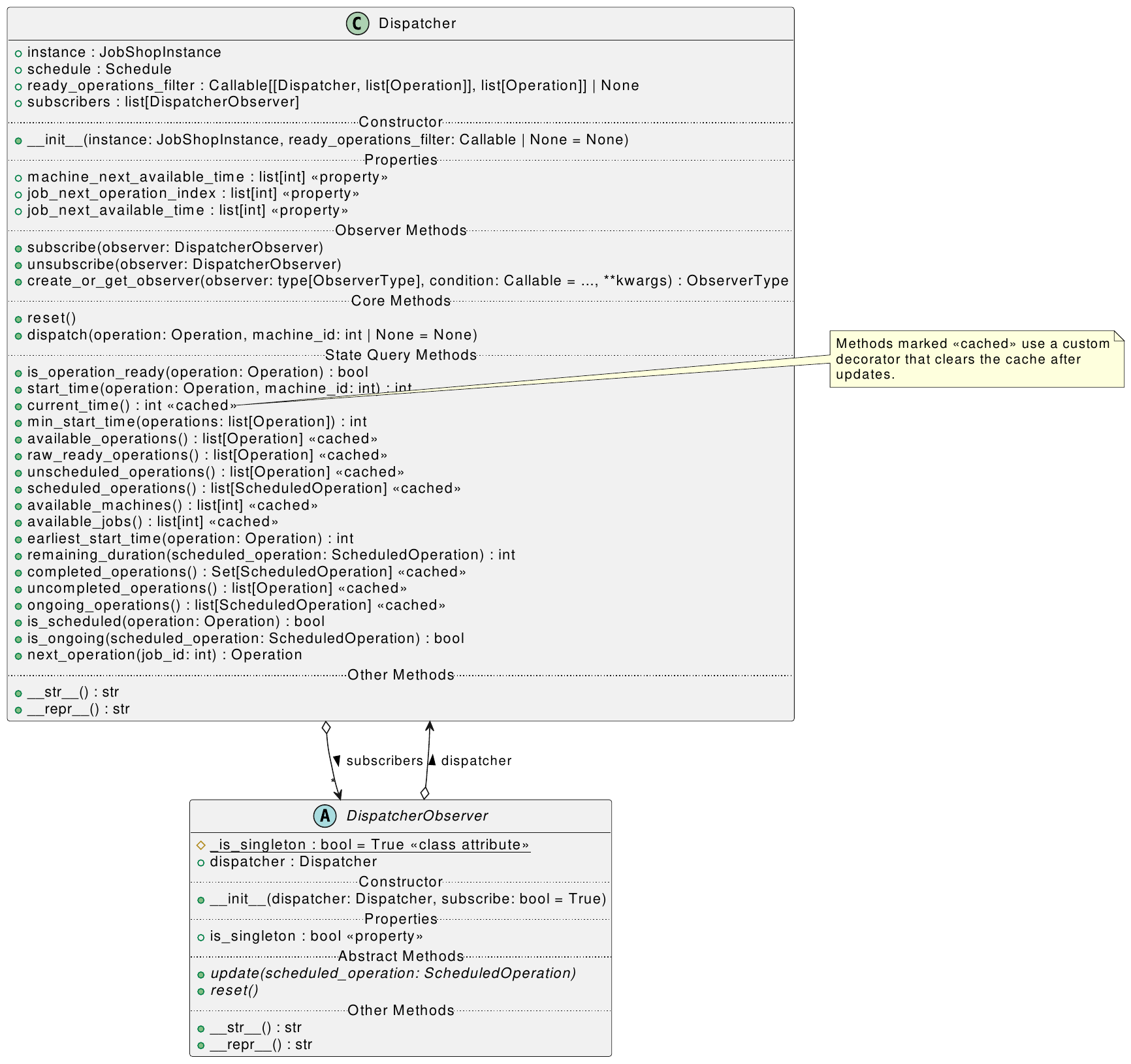}
\caption[\texttt{Dispatcher} and \texttt{DispatcherObserver} class diagram.]{\texttt{Dispatcher} and \texttt{DispatcherObserver} class diagram. Private methods and attributes of all the classes are omitted for simplicity.}
\label{fig:dispatcher_uml}
\end{figure}


\section{Dispatching Rules}
\label{sec:dispatching_rules}

Using the \texttt{Dispatcher} class, creating basic dispatching rules is a straightforward task. However, before writing the first PDR, we need to decide on a common interface for these functions. The interface chosen for this project is a callable that receives a \texttt{Dispatcher} and returns the chosen \texttt{Operation}. We do not need to pass it a list of available actions since they are already accessible through the \texttt{Dispatcher.available\_operations} method.

\textit{JobShopLib} includes several basic, built-in \ac{PDRs} within its \texttt{dispatching.rules} module. Here are a few examples:
\begin{itemize}
    \item \textbf{\ac{SPT}:} Selects the available operation with the minimum duration.
    \begin{python}[frame=single]
def shortest_processing_time_rule(dispatcher: Dispatcher):
    return min(
        dispatcher.available_operations(),
        key=lambda operation: operation.duration,
    )
    \end{python}

    \item \textbf{\ac{FCFS}---based on job sequence:} Selects the operation appearing earliest in its job sequence (minimum \texttt{position\_in\_job}).
    \begin{python}[frame=single]
def first_come_first_served_rule(dispatcher):
    return min(
        dispatcher.available_operations(),
        key=lambda operation: operation.position_in_job,
    )
    \end{python}

    \item \textbf{\ac{MWKR}:} Selects the operation whose job has the largest sum of durations for its remaining unscheduled operations.
    \begin{python}[frame=single]
def most_work_remaining_rule(dispatcher):
    job_remaining_work = [0.0] * dispatcher.instance.num_jobs
    for op in dispatcher.unscheduled_operations():
        job_remaining_work[op.job_id] += op.duration
    return max(
        dispatcher.available_operations(),
        key=lambda op: job_remaining_work[op.job_id],
    )
    \end{python}
\end{itemize}

\subsection{Score-Based Rules}
\label{subsec:score_based_rules}

A versatile approach involves assigning a numerical score to each job based on the current state. The rule then selects an available operation belonging to the job with the highest score. The \href{https://job-shop-lib.readthedocs.io/en/stable/api/job_shop_lib.dispatching.rules.html#job_shop_lib.dispatching.rules.score_based_rule}{\texttt{score\_based\_rule}} factory creates such rules from a scorer function. Example: scoring jobs by the number of remaining operations (\texttt{most\_operations\_remaining\_score}):
\begin{python}[frame=single,basicstyle=\ttfamily\small]
from job_shop_lib.dispatching import Dispatcher
from job_shop_lib.dispatching.rules import score_based_rule

def most_operations_remaining_score(dispatcher: Dispatcher) -> list[int]:
    scores = [0] * dispatcher.instance.num_jobs
    for operation in dispatcher.uncompleted_operations():
        scores[operation.job_id] += 1
    return scores

most_ops_rule = score_based_rule(most_operations_remaining_score)
\end{python}

Certain rules require state information that is computationally expensive to derive repeatedly. The observer pattern (Section \ref{subsec:dispatcher_observer}) allows rules to use \texttt{DispatcherObserver} instances that efficiently maintain specific state representations, updating incrementally as operations are dispatched. See the \href{https://job-shop-lib.readthedocs.io/en/stable/api/job_shop_lib.dispatching.rules.html#job_shop_lib.dispatching.rules.MostWorkRemainingScorer}{\texttt{MostWorkRemainingScorer}} class for an example.





\subsection{The \texttt{DispatchingRuleSolver}}
\label{subsec:dispatching_rule_solver}

An important observation is that while those functions provide a way to choose the next operation to dispatch, they are not a solver on their own. The class \href{https://job-shop-lib.readthedocs.io/en/stable/api/job_shop_lib.dispatching.rules.html#job_shop_lib.dispatching.rules.DispatchingRuleSolver}{\texttt{DispatchingRuleSolver}} allows us to create a solver from a dispatching rule and a machine chooser.

Its \texttt{solve()} method takes a \texttt{JobShopInstance}, initializes a \texttt{Dispatcher}, and iteratively calls its \texttt{step()} method until the schedule is complete (\texttt{dispatcher.schedule.is\_complete()} is true). Each \texttt{step()} performs one dispatch cycle:
\begin{enumerate}
    \item Invokes the \texttt{dispatching\_rule} to select an \texttt{Operation}.
    \item Invokes a \texttt{machine\_chooser} to select a machine ID (not needed in classical problems, the focus of this project). A \texttt{machine\_chooser} is a callable (e.g., a function) that takes the \texttt{Dispatcher} and the chosen \texttt{Operation}, returning a machine ID.
    \item Calls \texttt{dispatcher.dispatch()} to schedule the operation on the chosen machine.
\end{enumerate}
The final \texttt{Schedule} is then returned. In other words, this class provides a standard way to execute scheduling using any configured dispatching strategy.

%% file: chapters/07-TheRLEnv.tex
\doublespacing 

\chapter{The Reinforcement Learning Environment}
\label{ch7}

\begin{spacing}{1} 
\minitoc 
\end{spacing} 
\thesisspacing 

Once the basic components in \textit{JobShopLib} have been explained, we are in a position to introduce the \ac{RL} environment. In contrast to the previous environments mentioned in Section \ref{sec:rl_envs}, \textit{JobShopLib}'s environment allows us to customize all the aspects of the \ac{SMDP} mentioned in Chapter \ref{ch5}. They include:
\begin{itemize}[itemsep=0pt, topsep=0pt]
    \item The state: How to define the set of vertices $V_k$, edges $E_k$, and the node feature matrix $\mathbf{X}_k$ at each step $k$?
    \item The action: What constitutes an available operation?
    \item The reward function: What is the reward $r_k$ at step $k$?
\end{itemize}
At a high level, the state $s_k=G_k$ is updated using \texttt{DispatcherObserver} subclasses that track these features. In particular, new abstract classes are introduced:
\begin{itemize}[itemsep=0pt, topsep=0pt]
    \item A \texttt{GraphUpdater} for managing the graph topology.
    \item A \texttt{FeatureObserver} to track relevant node features. \textit{JobShopLib} provides several built-in feature observers, whose features can be concatenated into a single matrix $\mathbf{X}_k$ using the \texttt{CompositeFeatureObserver} class.
    \item Finally, a \texttt{RewardObserver} that will compute the reward at every step.
\end{itemize}
By inheriting from these classes, we support any definition of state space, transition function, and reward function. The action space is defined by choosing a \texttt{ready\_operations\_filter} as mentioned in Subsection \ref{subsec:ready_op_filters}.

This environment, despite being designed for \ac{RL}, also supports \ac{IL} workflows. In particular, it can be used to create a dataset of observations and optimal actions. We demonstrated this capacity with the experiments described in the next chapter.

Unlike existing environments with fixed designs, this flexibility is essential because different state representations, action spaces, and reward functions can significantly impact performance. As we mentioned in \ref{ch5}, no single approach has proven universally superior for all \ac{JSSP} variants.

\begin{figure}[H]
    \centering
    \includegraphics[width=\linewidth]{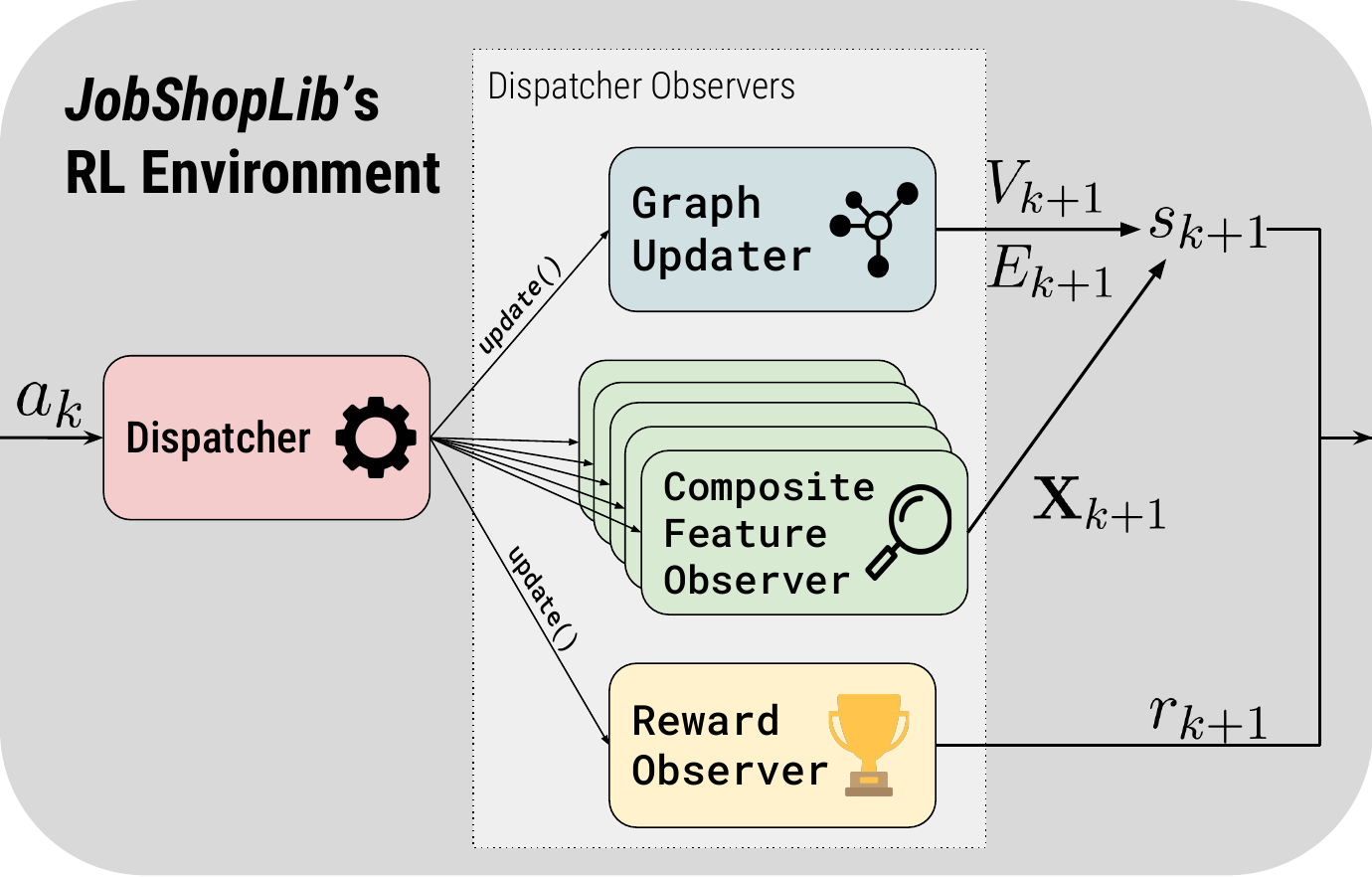}
    \caption{High-level overview of \textit{JobShopLib}'s \ac{RL} environment's architecture.}
    \label{fig:env-high-level-diagram}
\end{figure}

\section{Computing Node Features}
\label{sec:computing_node_features}
As we mentioned, nodes can represent operations, machines, jobs, or even more abstract concepts such as the state of \ac{JSSP} instance itself (e.g., with a global node). Thus, each node contains a user-defined set of raw features that represent each entity. We need to update these features dynamically to take advantage of the benefits of residual scheduling discussed in Subsection \ref{subsec:residual_scheduling}. For example, we need to consider only the remaining duration of ongoing operations. This need to update features at each step makes using the observer pattern very convenient.

An important observation is that node features $\mathbf{X_k}$ can be computed independently of the graph structure $(V_k, E_k)$ because the graph structure itself is constructed from these features. For this reason, we can define feature observer objects that listen directly to the \texttt{Dispatcher} and compute job, machine, or operation-related features.

\subsection{The \texttt{FeatureObserver} Class}
The abstract class \href{https://job-shop-lib.readthedocs.io/en/stable/api/job_shop_lib.dispatching.feature_observers.html#job_shop_lib.dispatching.feature_observers.FeatureObserver}{\texttt{FeatureObserver}} provides a common interface for all these feature observers. It is a subclass of \texttt{DispatcherObserver} that specifically observes features related to operations, machines, or jobs within the \texttt{Dispatcher}.

Attributes observed by a \texttt{FeatureObserver} are stored in NumPy arrays \citep{harris2020array}. These arrays typically have a shape of (\textit{num\_entities}, \textit{feature\_size}), where \textit{num\_entities} corresponds to the count of entities being monitored (operations, machines, or jobs) and \textit{feature\_size} is the number of feature values recorded for each entity. Using NumPy arrays offers efficiency, as updates can be performed in a vectorized manner (i.e., the tracked attributes can be updated in parallel rather than sequentially). The arrays are stored in a dictionary called \texttt{features}, where keys are \texttt{FeatureType} enums (OPERATIONS, MACHINES, JOBS).

\noindent Key methods include:
\begin{itemize}[itemsep=0pt, topsep=0pt]
    \item \texttt{initialize\_features()}: Initializes the features based on the dispatcher's current state. This is called automatically upon observer initialization and also by the default \texttt{update} method. Subclasses should implement the logic for calculating initial feature values here.
    \item \texttt{update(scheduled\_operation)}: Updates features when an operation is scheduled. The default implementation simply calls \texttt{initialize\_features()} again, but subclasses can override this for more incremental and efficient updates.
    \item \texttt{reset()}: Resets all features to zero and then calls \texttt{initialize\_features()} to re-establish the initial state.
\end{itemize}

\subsection{Built-in Feature Observers}
\textit{JobShopLib} provides a comprehensive list of feature observers that follow the principles described in Subsection \ref{subsec:feature_matrix}. Some of them were already mentioned in this subsection, too, such as the remaining processing time $p^{(k)}_{ij} = C_{ij} - t_k$. In this subsection, we revisit these features, mentioning their associated \texttt{FeatureObserver}, and introduce new ones.

\subsubsection{Status Features}
\begin{sloppypar}
The status features are binary features that indicate the current status of a job, machine, or operation (ready, scheduled, or completed). They are tracked by the \href{https://job-shop-lib.readthedocs.io/en/stable/api/job_shop_lib.dispatching.feature_observers.html#job_shop_lib.dispatching.feature_observers.IsReadyObserver}{\texttt{IsReadyObserver}}, \href{https://job-shop-lib.readthedocs.io/en/stable/api/job_shop_lib.dispatching.feature_observers.html#job_shop_lib.dispatching.feature_observers.IsScheduledObserver}{\texttt{IsScheduledObserver}}, and \href{https://job-shop-lib.readthedocs.io/en/stable/api/job_shop_lib.dispatching.feature_observers.html#job_shop_lib.dispatching.feature_observers.IsCompletedObserver}{\texttt{IsCompletedObserver}} objects, respectively. The definitions of these statuses are the following:
\begin{itemize}[itemsep=0pt, topsep=0pt]
    \item An operation is ready if it belongs to the ``available operations set'' defined by \texttt{Dispatcher.available\_operations()} as described in Subsection \ref{subsec:ready_op_filters}. Similarly, a machine or job is ready if it contains at least one available operation.
    \item An operation is considered scheduled if it has already been dispatched. A machine or job is considered scheduled if it contains at least one operation scheduled but not completed.
    \item Finally, an operation is completed when the current time $t_k$, defined as the earliest start time of the available operations, is greater than or equal to the completion time of this operation (i.e., $C_{ij} \leq t_k$). A machine or job is considered completed once all its operations have been completed. Note that a machine or job can still be considered uncompleted even if all its operations have been scheduled. Another important consideration is that this feature observer is not usually used as a feature of $\mathbf{X}_k$ because all completed entities are removed from the graph following residual scheduling's principles.
\end{itemize}
\end{sloppypar}

\subsubsection{Earliest Start Time}
As described in Subsection \ref{subsec:feature_matrix}, the ``earliest start time'' feature measures the earliest start time $S^{*(k)}_{ij}$ possible that that respects the problem constraints adjusted by the current time: $R^{(k)}_{ij} = S^{*(k)}_{ij} - t_k$. This feature is a good example of how the Observer pattern can save us computation by allowing us to take advantage of the features computed in the previous state. To illustrate this, let's see how this feature is computed internally by the \href{https://job-shop-lib.readthedocs.io/en/stable/api/job_shop_lib.dispatching.feature_observers.html#job_shop_lib.dispatching.feature_observers.EarliestStartTimeObserver}{\texttt{EarliestStartTimeObserver}}.

Initially, the observer establishes a baseline for every operation's absolute earliest start times. This is stored internally in a two-dimensional NumPy array $\mathbf{S}^*$, where $S^*_{ij}$ represents the earliest start time of the $j$-th operation of the $i$-th job. The first operation of each job $S^*_{i0}$ is set to 0. For subsequent operations within the same job its initial earliest start time $S^*_{ij}$ is calculated as the sum of the processing times of all preceding operations $O_{il}$ (where $l < j$) in that job, added to the earliest start time of its immediate predecessor $S^*_{i,j-1}$. This array captures the earliest points at which operations could begin if there were no machine conflicts, based solely on their sequence within each job.

When an operation $O_{s\_op}$ is scheduled and thus its completion time is known, the internal $\mathbf{S}^*$ array is updated. This update has two main considerations. First, for the job to which $O_{s\_op}$ belongs, the earliest start time of the next operation in that job $O_{next\_job\_op}$ is re-evaluated. $S^*_{next\_job\_op}$ is set to the maximum of: (a) the completion time of $O_{s\_op}$; (b) its own previously calculated $S^*_{next\_job\_op}$ (based on job precedence); and (c) the earliest time its required machine becomes available. If this re-evaluation pushes $S^*_{next\_job\_op}$ later, this delay (or gap) is propagated by adding it to the $S^*$ values of $O_{next\_job\_op}$ and all subsequent operations in that same job.

Second, the machine that $O_{s\_op}$ utilized is considered. For every other unscheduled operation $O_{other}$ that is also assigned to this machine, its earliest start time $S^*_{other}$ is updated to be no earlier than the completion time of $O_{s\_op}$. If this constraint forces $S^*_{other}$ to be later than its previous value, the resulting delay is applied to $O_{other}$. As before, this delay needs to be propagated through $O_{other}$'s own job sequence by adding the delay to the $S^*$ values of all subsequent operations in its job.

After these internal absolute earliest start times $S^*_{ij}$ are updated, the actual feature value $R^{(k)}_{ij}$ that can be incorporated into the feature matrix $\mathbf{X}_k$ is computed. For any operation $O_{ij}$, this is done by subtracting the current time $t_k$ from its updated absolute earliest start time: $R^{(k)}_{ij} = S^{*(k)}_{ij} - t_k$. Similar normalized earliest start time features are computed for machines (based on the minimum $S^{*(k)}_{ij} - t_k$ of unscheduled operations on that machine) and for jobs (based on the $S^{*(k)}_{ij} - t_k$ of the next unscheduled operation in that job). This normalization ensures the feature represents the earliest start time relative to the current decision point $k$.

While this feature can be more computationally expensive than simpler ones, we mitigate this fact by using this incremental approach. Instead of recalculating all earliest start times from scratch at each step k, the observer updates only the necessary parts of the internal $\mathbf{S}^*$ array affected by the last scheduled operation. This targeted update significantly reduces computational overhead. Furthermore, because these earliest start times are stored in NumPy arrays, the update operations, such as propagating delays through job sequences or identifying minimums across machine operations, can often be performed using vectorized operations.

\subsubsection{Remaining Processing Time}
The \href{https://job-shop-lib.readthedocs.io/en/stable/api/job_shop_lib.dispatching.feature_observers.html#job_shop_lib.dispatching.feature_observers.DurationObserver}{\texttt{DurationObserver}} tracks the remaining processing time for operations, and the aggregate sum of durations for unscheduled operations on machines and within jobs. For an individual operation, its feature initially represents its full processing time. Once scheduled, this feature is updated to reflect its actual remaining duration ($C_{ij} - t_k$). 

For machines and jobs, the feature represents the sum of durations of their constituent operations that have not been scheduled yet. When an operation is scheduled, its original full duration is subtracted from the total duration tracked for its assigned machine and its parent job, thereby dynamically updating the remaining unscheduled workload for these entities. Note that considering uncompleted scheduled operations would also be a valid definition. We chose this one because it is significantly more efficient to compute. In particular, the other definition requires computing these features from scratch, because it is not possible to efficiently know which scheduled operations have been completed, given the current scheduled one. On the other hand, the definition we propose can be updated with just one subtraction.

\subsubsection{Position in Job}
This feature, computed by the \href{https://job-shop-lib.readthedocs.io/en/stable/api/job_shop_lib.dispatching.feature_observers.html#job_shop_lib.dispatching.feature_observers.PositionInJobObserver}{\texttt{PositionInJobObserver}} class, simply adjusts the original position ($j$) of an operation $O_{ij}$ by subtracting from $j$ the number of already scheduled operations in that job. Naturally, this feature cannot be computed for machines or jobs.

\subsubsection{Remaining Operations}
The \href{https://job-shop-lib.readthedocs.io/en/stable/api/job_shop_lib.dispatching.feature_observers.html#job_shop_lib.dispatching.feature_observers.RemainingOperationsObserver}{\texttt{RemainingOperationsObserver}} tracks the number of unscheduled operations for jobs and machines. It is initialized by counting all unscheduled operations for each entity. When an operation is scheduled, the counts for that operation's job and machine are decremented. It does not compute features for individual operations.

\subsubsection{Composite Feature Observer}
The \href{https://job-shop-lib.readthedocs.io/en/stable/api/job_shop_lib.dispatching.feature_observers.html#job_shop_lib.dispatching.feature_observers.CompositeFeatureObserver}{\texttt{CompositeFeatureObserver}} combines features from multiple individual \texttt{FeatureObserver} instances into a unified representation. This observer is particularly useful for constructing the complete node feature matrix $\mathbf{X}_k$ by allowing various distinct features to be managed by specialized observers and then merged. It takes the feature matrices generated by a list of specified observers (or all observers subscribed to the dispatcher if none are specified) and concatenates them horizontally (along \texttt{axis=1}) for each \texttt{FeatureType} (OPERATIONS, MACHINES, JOBS). This process results in a wider feature matrix for each entity type, where the columns represent the combined features from all constituent observers.

\begin{figure}[H]
    \centering
    \includegraphics[width=1.06\linewidth]{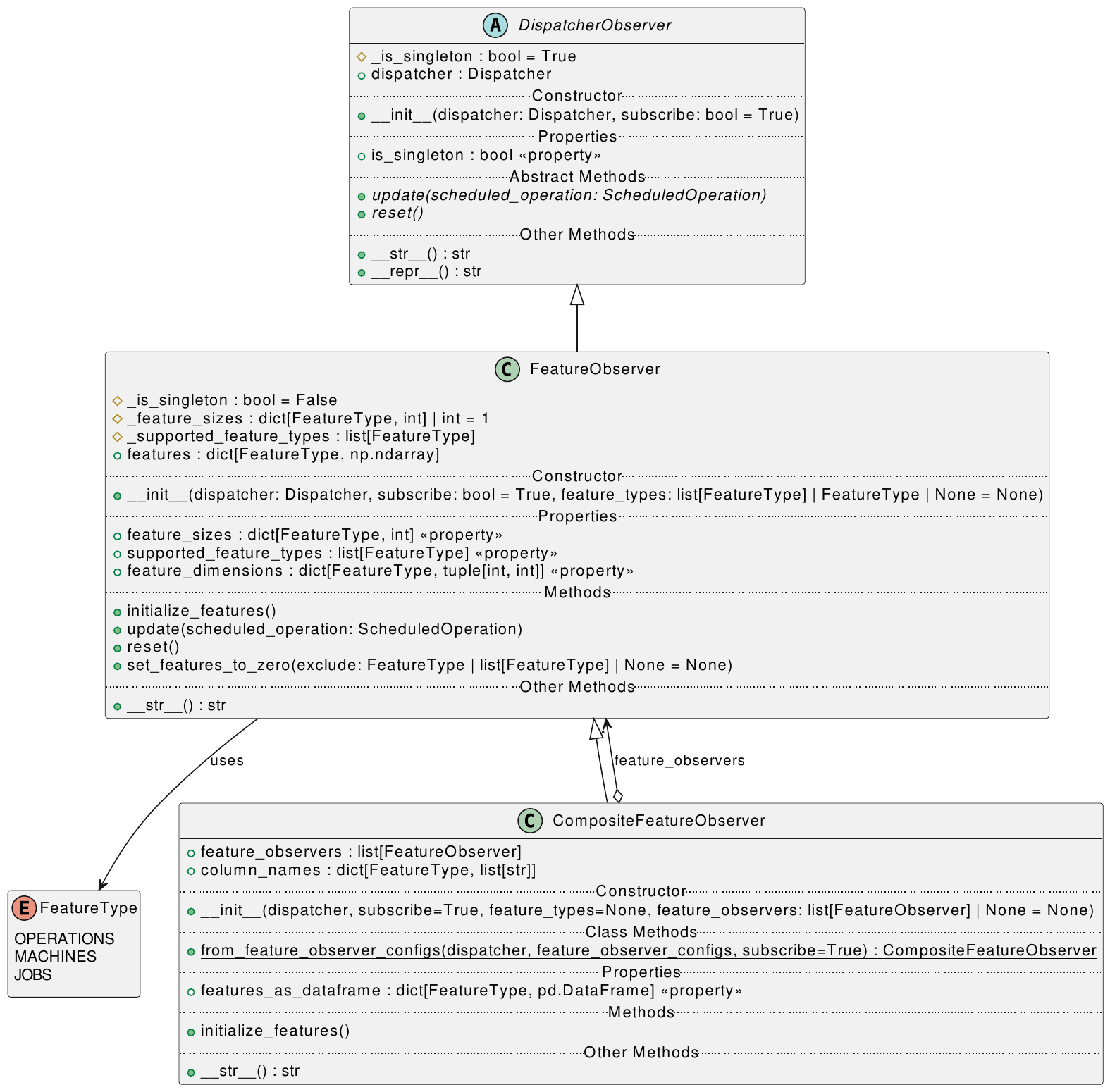}
    \caption[\texttt{FeatureObserver} and \texttt{CompositeFeatureObserver} class diagram.]{\texttt{FeatureObserver} and \texttt{CompositeFeatureObserver} class diagram. The classes' private methods and attributes (except class attributes) are omitted for simplicity. Some type hints were also omitted for size constraints.}
    \label{fig:composite_feature_obs_class_diagram}
\end{figure}

\section{The Graph Representation}
\label{sec:graph_representation}
Before explaining how the \texttt{GraphUpdater} works, we need to introduce an additional data structure: the \texttt{JobShopGraph}. This class allows us to store a graph representation of a \texttt{JobShopInstance} that will be updated by the \texttt{GraphUpdater} while running the \ac{RL} environment.

The \texttt{JobShopGraph} uses the \href{https://networkx.org/}{\texttt{networkx}} library internally to manage a directed graph (\href{https://networkx.org/documentation/stable/reference/classes/digraph.html}{\texttt{nx.DiGraph}}). As we mentioned, the nodes in this graph represent different entities within the JSSP, such as operations, machines, and jobs. These nodes are instances of the \href{https://job-shop-lib.readthedocs.io/en/stable/api/job_shop_lib.graphs.html#job_shop_lib.graphs.Node}{\texttt{Node}} class, which stores information about the node's type and associated data, such as an \texttt{Operation} object for operation nodes or a machine or job ID for machine or job nodes, respectively. The \texttt{Node} class defines several types using the \href{https://job-shop-lib.readthedocs.io/en/stable/api/job_shop_lib.graphs.html#job_shop_lib.graphs.NodeType}{\texttt{NodeType}} enum, including \texttt{OPERATION}, \texttt{MACHINE}, and \texttt{JOB}. Each node is assigned a unique integer ID (\texttt{node\_id}) when it is added to the graph, making it hashable.

\begin{figure}[H]
    \centering
    \includegraphics[width=1\linewidth]{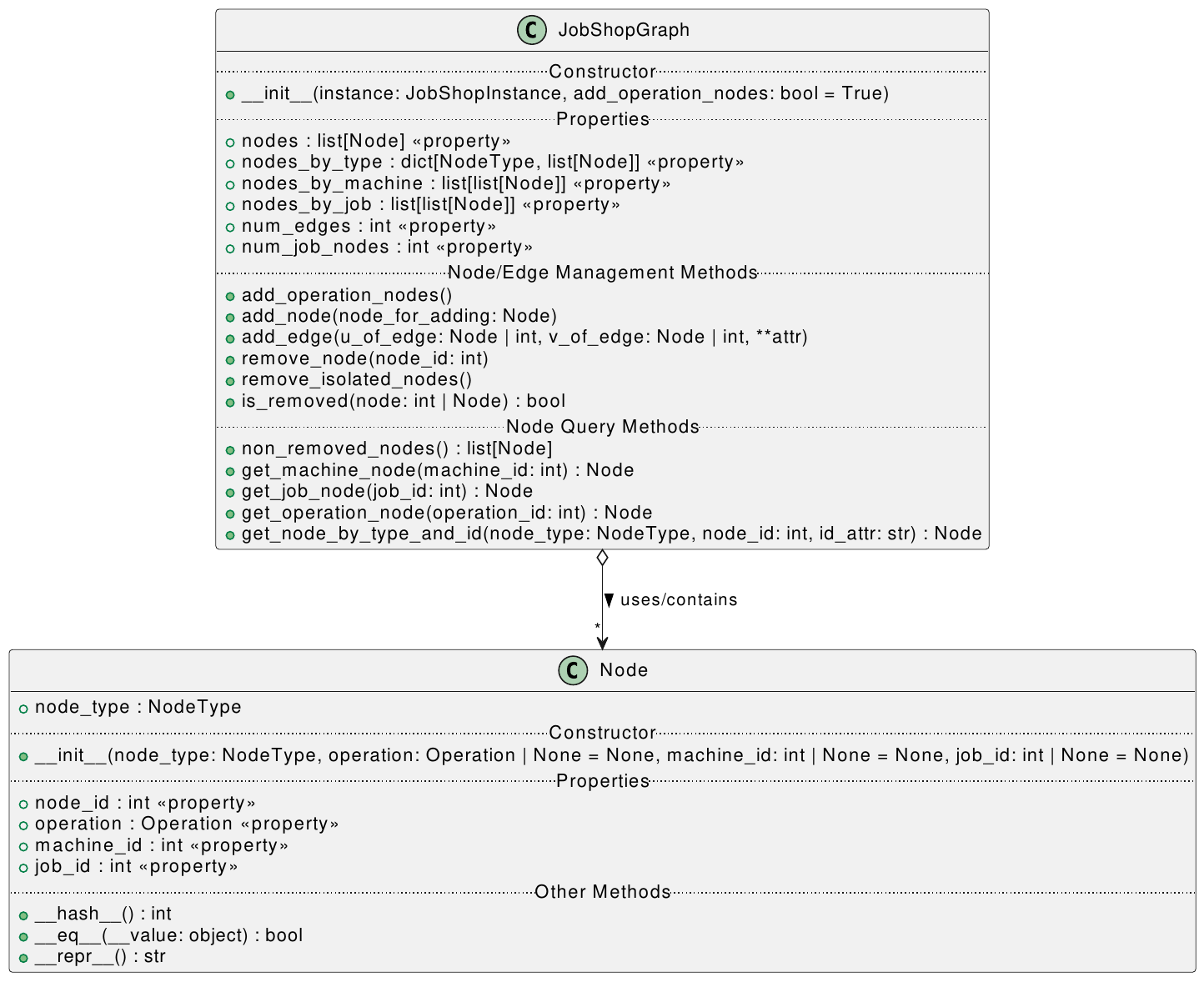}
    \caption[\texttt{JobShopGraph} and \texttt{Node} class diagram.]{\texttt{JobShopGraph} and \texttt{Node} class diagram. Private methods and attributes of all the classes are omitted for simplicity.}
    \label{fig:graph_and_node_class_diagram}
\end{figure}

\subsection{Defining the Initial Graph}
The initial graph $(V_0, E_0)$ needs to be created outside the \texttt{GraphUpdater} class, which only updates the graph. In other words, we need a function that, given a \texttt{JobShopInstance} as an argument, returns a \texttt{JobShopGraph}. 

Nodes and edges are added to the graph primarily through the \texttt{add\_node()} and \texttt{add\_edge()} methods.
\begin{itemize}[itemsep=0pt, topsep=0pt]
    \item \texttt{add\_node(node\_for\_adding: Node)}: Assigns the next available ID to the input \texttt{Node} object, adds it to the \texttt{networkx} graph (storing the \texttt{Node} object itself as a node attribute), and updates its internal lists (\texttt{\_nodes}, \texttt{\_nodes\_by\_type}, etc.) accordingly. By default, when a \texttt{JobShopGraph} is initialized, it automatically adds nodes for all operations in the provided \texttt{JobShopInstance}.
    \item \texttt{add\_edge(u\_of\_edge, v\_of\_edge, **attr)}: Adds a directed edge between the specified nodes (identified by their \texttt{Node} object or integer ID). It automatically determines the edge type (e.g., \texttt{('operation', 'to', 'operation')}) based on the node types, unless it is explicitly provided.
\end{itemize}

\textit{JobShopLib} provides builders for the two primary initial graph structures mentioned in Section \ref{sec:graph_representations}. The first is the classic disjunctive graph, constructed using \href{https://job-shop-lib.readthedocs.io/en/stable/api/job_shop_lib.graphs.html#job_shop_lib.graphs.build_disjunctive_graph}{\texttt{build\_disjunctive\_graph}}. This graph includes operation nodes, along with optional special source ($S$) and sink ($T$) nodes. Conjunctive edges link operations within the same job sequentially, while bidirectional disjunctive edges connect operations that require the same machine but belong to different jobs.

\begin{sloppypar}
The second approach involves building a resource-task graph using functions like \href{https://job-shop-lib.readthedocs.io/en/stable/api/job_shop_lib.graphs.html#job_shop_lib.graphs.build_resource_task_graph}{\texttt{build\_resource\_task\_graph}}. Instead of direct disjunctive edges, this representation introduces explicit machine nodes. Operation nodes are connected to the machine node they require (see Subsection \ref{subsec:resource_task_graphs} for more information). Variations are also supported, such as adding job nodes connected to their respective operations (\href{https://job-shop-lib.readthedocs.io/en/stable/api/job_shop_lib.graphs.html#job_shop_lib.graphs.build_resource_task_graph_with_jobs}{\texttt{build\_resource\_task\_graph\_with\_jobs}}) or a global node connecting all machine and job nodes (\href{https://job-shop-lib.readthedocs.io/en/stable/api/job_shop_lib.graphs.html#job_shop_lib.graphs.build_complete_resource_task_graph}{\texttt{build\_complete\_resource\_task\_graph}}).
\end{sloppypar}
\subsection{Updating the Graph}
\begin{sloppypar}
Nodes can be removed using \texttt{JobShopGraph.remove\_node()}, which removes the node from the \texttt{networkx} graph and marks it as removed in the \texttt{removed\_nodes} list. The class also provides a method \texttt{JobShopGraph.remove\_isolated\_nodes()} to clean up nodes that no longer have any connections.
\end{sloppypar}

Potentially using these methods, the \href{https://job-shop-lib.readthedocs.io/en/stable/api/job_shop_lib.graphs.graph_updaters.html#job_shop_lib.graphs.graph_updaters.GraphUpdater}{\texttt{GraphUpdater}} abstract base class dynamically modifies the \texttt{JobShopGraph} during an \ac{RL} episode, as we mentioned. As a subclass of \texttt{DispatcherObserver}, it automatically receives notifications from the dispatcher whenever an operation is scheduled.

The primary responsibility of a \texttt{GraphUpdater} is defined by its abstract \texttt{update} method. The logic within \texttt{update} dictates how the graph's nodes ($V_k$) and edges ($E_k$) should be modified to reflect the new state of the scheduling process. Additionally, the \texttt{GraphUpdater} provides a \texttt{reset} method, which restores the \texttt{job\_shop\_graph} attribute to its initial configuration.

\textit{JobShopLib} provides concrete implementations of \texttt{GraphUpdater} that correspond to the graph update strategies discussed in Chapter \ref{ch5}:
\begin{itemize}[itemsep=0pt, topsep=0pt]
    \item \href{https://job-shop-lib.readthedocs.io/en/stable/api/job_shop_lib.graphs.graph_updaters.html#job_shop_lib.graphs.graph_updaters.ResidualGraphUpdater}{\texttt{ResidualGraphUpdater}}: Implements the residual scheduling approach described in Subsection \ref{subsec:residual_scheduling}. It primarily removes completed operation nodes from the graph using the \texttt{remove\_completed\_operations} utility function. Optionally, based on the boolean flags \texttt{remove\_completed\_machine\_nodes} and \texttt{remove\_completed\_job\_nodes}, it can also remove machine and job nodes once all their associated operations are finished. To track completion status, it utilizes an \texttt{IsCompletedObserver} (explained in the next section).
    \item \href{https://job-shop-lib.readthedocs.io/en/stable/api/job_shop_lib.graphs.graph_updaters.html#job_shop_lib.graphs.graph_updaters.DisjunctiveGraphUpdater}{\texttt{DisjunctiveGraphUpdater}}: Extends \texttt{ResidualGraphUpdater} to specifically handle disjunctive graphs (Subsection \ref{subsec:disjunctive_graphs_for_gnns}). In addition to removing completed nodes, it implements the ``Removing-arc'' strategy discussed in \cite{Park2021l2s}. After an operation is scheduled, it removes the disjunctive edge pointing from the newly scheduled operation to the previously scheduled operation on the same machine, as well as all disjunctive edges connected to the previously scheduled operation that involve operations not yet scheduled on that machine.
\end{itemize}
This modular design allows users to easily select standard graph update mechanisms or implement their own custom \texttt{GraphUpdater} subclass to experiment with novel state representations and transition dynamics within the \ac{RL} environment.
\begin{figure}
    \centering
    \includegraphics[width=1.1\linewidth]{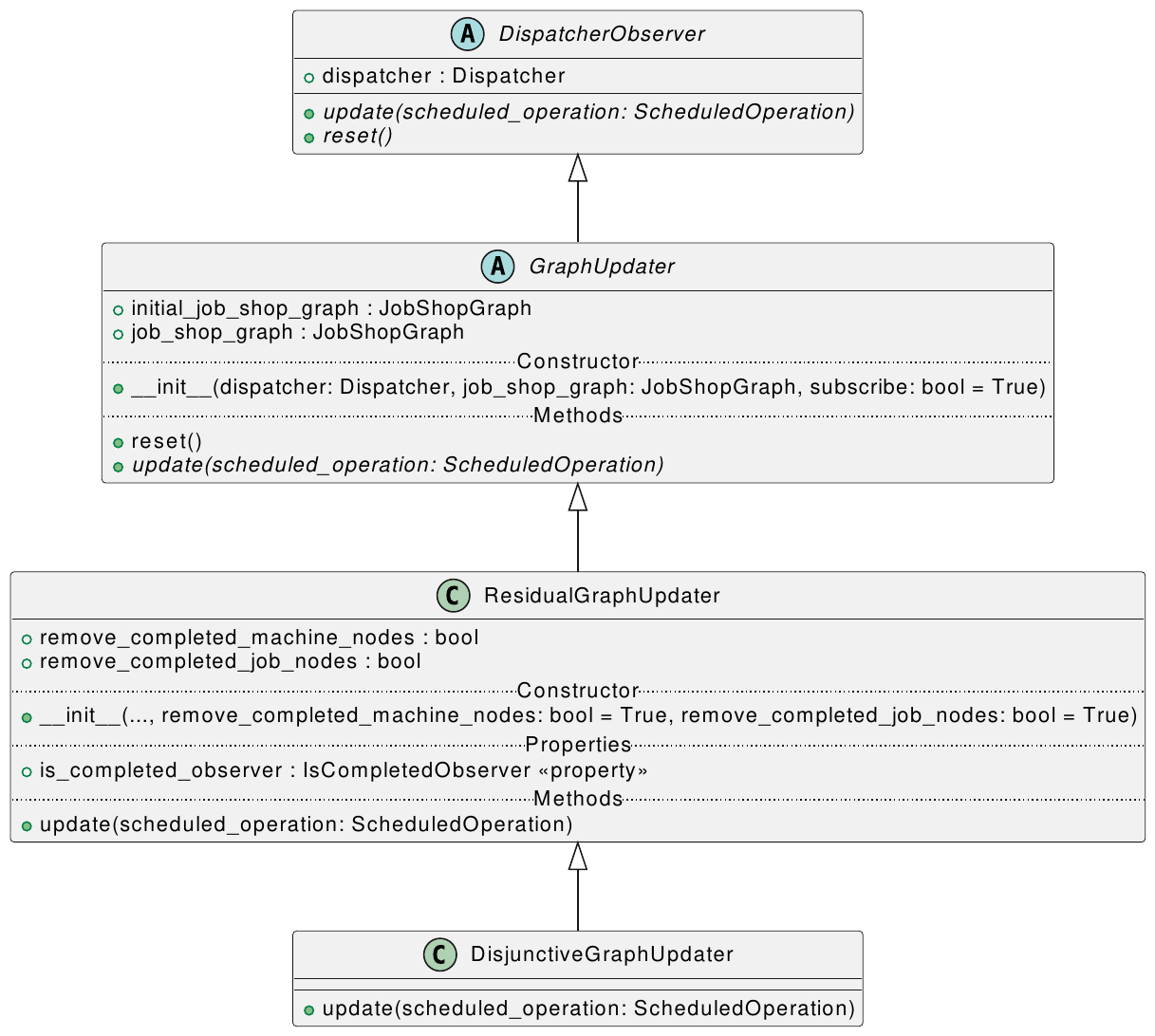}
    \caption[Class hierarchy of \textit{JobShopLib}'s graph updaters.]{Class hierarchy of \textit{JobShopLib}'s graph updaters. The omitted arguments of \texttt{ResidualGraphUpdater}'s constructor are those present in all \texttt{GraphUpdater}'s subclasses (\texttt{dispatcher}, \texttt{job\_shop\_graph}, and \texttt{subscribe}).  Private methods and attributes of all the classes are also omitted for simplicity.}
    \label{fig:graph_updaters}
\end{figure}

\section{Reward Observers}
\label{sec:reward_observers}
The final component of the \ac{RL} environment to be defined is the reward function $R$. In \textit{JobShopLib}, rewards are computed by subclasses of the \href{https://job-shop-lib.readthedocs.io/en/stable/api/job_shop_lib.reinforcement_learning.html#job_shop_lib.reinforcement_learning.RewardObserver}{\texttt{RewardObserver}} class. These observers monitor the \texttt{Dispatcher} and calculate a reward value $r_k$ each time an operation is scheduled. While the ultimate goal is often to minimize a global metric like makespan (which could be represented as a sparse reward $-C_{\text{max}}$ at the end of an episode), dense rewards provided at each step are generally preferred. As discussed in Section \ref{sec:reward_funcs}, sparse rewards can lead to significant challenges in credit assignment and slow down learning. Dense rewards offer more frequent feedback, guiding the agent more effectively.

\subsection{The \texttt{RewardObserver} Class}
The abstract class \href{https://job-shop-lib.readthedocs.io/en/stable/api/job_shop_lib.reinforcement_learning.html#job_shop_lib.reinforcement_learning.RewardObserver}{\texttt{RewardObserver}} serves as the base for all reward computation modules. As a subclass of \texttt{DispatcherObserver}, it is automatically notified when an operation is scheduled.

Its primary attribute is \texttt{rewards}, a list that stores the sequence of rewards calculated at each step (i.e., after each operation is scheduled).

\noindent Key methods include:
\begin{itemize}[itemsep=0pt, topsep=0pt]
    \item \texttt{last\_reward()}: A property that returns the most recently calculated reward, or 0 if no rewards have been computed yet.
    \item \texttt{reset()}: Clears the \texttt{rewards} list, typically called at the beginning of a new episode.
    \item \texttt{update(scheduled\_operation)}: This abstract method must be implemented by subclasses. The dispatcher calls it after an operation is scheduled and should contain the logic to calculate the reward for the current step and append it to the \texttt{rewards} list.
\end{itemize}
By inheriting from \texttt{RewardObserver}, users can implement custom reward functions tailored to specific objectives or to experiment with different reward shaping techniques. This flexibility for defining custom reward functions is crucial, especially when facing problems with multiple objectives, which are common in real-world scenarios. For example, many problems require simultaneously minimizing makespan and delays (in problems with due dates). 

\subsection{Built-in Reward Observers}
\textit{JobShopLib} provides implementations of some of the dense reward functions discussed in Subsection \ref{sec:reward_funcs}.

\subsubsection{Makespan Reward}
The \href{https://job-shop-lib.readthedocs.io/en/stable/api/job_shop_lib.reinforcement_learning.html#job_shop_lib.reinforcement_learning.MakespanReward}{\texttt{MakespanReward}} observer calculates a dense reward based on the immediate change in makespan. At each step $k$, after scheduling an operation, the reward is $r_k = C_{\text{max}}^{(k-1)} - C_{\text{max}}^{(k)}$, where $C_{\text{max}}^{(k-1)}$ is the makespan before scheduling the current operation and $C_{\text{max}}^{(k)}$ is the makespan after. This approach is a form of difference reward, similar to $H(s_k) - H(s_{k+1})$ where $H(s)$ is a quality measure (like a makespan lower bound or the actual makespan) \citep{zhang2020l2d}, or the simpler $t_k - t_{k+1}$ formulation discussed in Section \ref{sec:reward_funcs}. 

With a discount factor $\gamma=1$, maximizing the sum of these rewards is equivalent to minimizing the final makespan, as the cumulative reward becomes $C_{\text{max}}^{(0)} - C_{\text{max}}^{(T)}$. Since $C_{\text{max}}^{(0)}$ (the initial makespan, often zero or based on initial conditions) is constant for a given problem start, this incentivizes actions that lead to a smaller final $C_{\text{max}}^{(T)}$.

\subsubsection{Idle Time Reward}
The \href{https://job-shop-lib.readthedocs.io/en/stable/api/job_shop_lib.reinforcement_learning.html#job_shop_lib.reinforcement_learning.IdleTimeReward}{\texttt{IdleTimeReward}} observer computes a dense reward aimed at minimizing machine idle time. As shown by \cite{tassel2021rl_env}, this objective is highly correlated with minimizing makespan.

When an operation $O_{ij}$ is scheduled, the reward is $r_k = -\text{idle\_time}_k$, where $\text{idle\_time}_k$ is the idle period on the operation's assigned machine immediately preceding its start. If it's the first operation on the machine, the idle time is its start time (typically zero); otherwise, it's the duration between the completion of the previous operation on that machine and the start of $O_{ij}$. This provides a direct penalty for idleness.

This formulation is related to the ``scheduled area'' reward proposed by \cite{tassel2021rl_env}, which is defined as $R(s_{k}, a_{k}, s_{k+1}) = p_{ij} - \text{idle\_time}_k$, where $p_{ij}$ is the processing time of the dispatched operation $O_{ij}$. Both rewards utilize a similar calculation for $\text{idle\_time}_k$. The key difference is the inclusion of the $p_{ij}$ term in \cite{tassel2021rl_env} formulation.

The term $p_{ij}$ in the ``scheduled area'' reward can be understood as a potential-based shaping term \citep{ng1999policy}. Specifically, $p_{ij} = \Phi(s_{k+1}) - \Phi(s_k)$, where $\Phi(s)$ is a potential function representing the sum of processing times of operations scheduled up to state $s$. Adding such a term does not change the set of optimal policies in terms of minimizing the total sum of idle times, because the cumulative sum $\sum p_{ij}$ is constant for a given problem instance ($P_{\text{total}}$). Thus, maximizing $\sum (p_{ij} - \text{idle\_time}_k) = P_{\text{total}} - \sum \text{idle\_time}_k$ is equivalent to maximizing $\sum (-\text{idle\_time}_k)$.

However, the per-step dense rewards differ, which can influence learning dynamics. The \texttt{IdleTimeReward} ($-\text{idle\_time}_k$) offers a pure penalty for idleness. In contrast, Tassel et al.'s reward ($p_{ij} - \text{idle\_time}_k$) provides an immediate positive incentive proportional to the ``work done'' ($p_{ij}$) in addition to penalizing idleness. This might affect exploration and convergence speed, as scheduling longer operations appears more immediately beneficial, provided the introduced idle time is not excessively large. Moreover, this change would significantly reduce the number of steps with no reward, since not introducing any gap is more common than introducing an idle time of exactly $p_{ij}$.

This potential-based shaping term (and any other) can be easily included by extending the \texttt{IdleTimeReward} class with a new update method:

\begin{python}[frame=single]
class ScheduledAreaReward(IdleTimeReward):
    def update(self, scheduled_operation: ScheduledOperation):
        # This will calculate idle_time and append `parent_reward = -idle_time`
        # to self.rewards.
        super().update(scheduled_operation)
        # We pop it because we will append the corrected reward.
        parent_reward = self.rewards.pop()
        actual_reward = parent_reward + scheduled_operation.duration
        self.rewards.append(actual_reward)
\end{python}

\begin{figure}[H]
\centering
\includegraphics[width=1.06\linewidth]{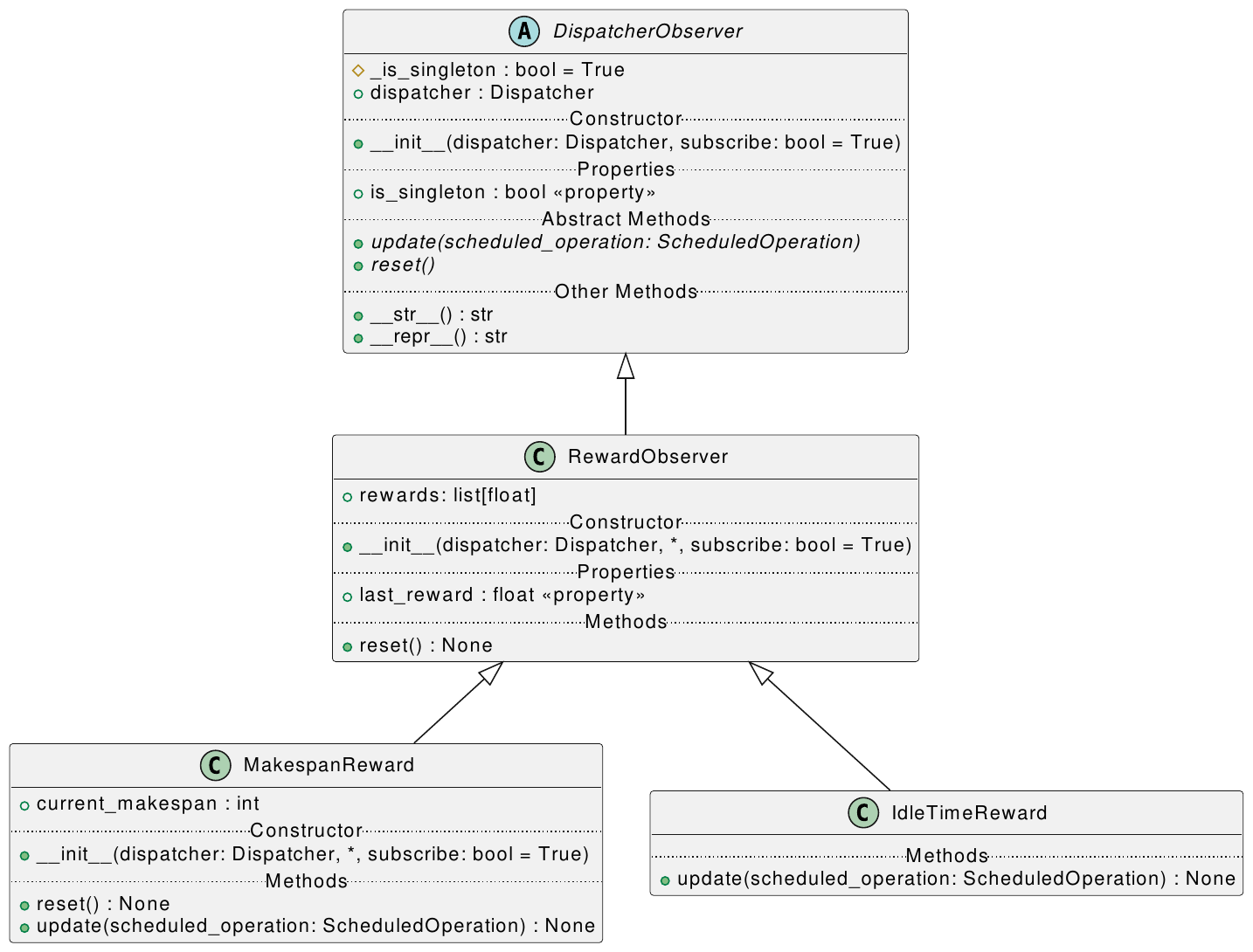}
\caption[Class hierarchy of \textit{JobShopLib}'s reward observers.]{Class hierarchy of \textit{JobShopLib}'s reward observers. Private methods and attributes (except class attributes) are omitted for simplicity.} \label{fig:reward_observers_class_diagram}
\end{figure}

\section{Gymnasium Environments in \textit{JobShopLib}}
\label{sec:gym_environments}

The preceding sections of this chapter have deconstructed the \ac{RL} problem for the \ac{JSSP} into its core components: methods for computing node features (Section \ref{sec:computing_node_features}), defining and updating graph representations (Section \ref{sec:graph_representation}), and formulating reward signals (Section \ref{sec:reward_observers}). This section now describes how \textit{JobShopLib} integrates these elements into fully functional \ac{RL} environments. These environments are designed to be compatible with the \href{https://gymnasium.farama.org/}{\texttt{gymnasium.Env}} interface \citep{towers2024gymnasium}, a widely adopted standard in the \ac{RL} community. This compatibility facilitates the use of established \ac{RL} algorithms and libraries for training agents to solve \ac{JSSP} instances.

\noindent \textit{JobShopLib} provides two primary environment classes for this purpose:
\begin{itemize}[itemsep=0pt, topsep=0pt]
    \item \texttt{SingleJobShopGraphEnv}: For a single \ac{JSSP} instance. Figure \ref{fig:env-high-level-diagram} illustrates the overall architecture of this environment.
    \item \texttt{MultiJobShopGraphEnv}:  For training agents that can generalize across a distribution of \ac{JSSP} instances, which can be dynamically generated during the training process. It uses the \texttt{SingleJobShopGraphEnv} internally.
\end{itemize}
The way these environments are initialized---by accepting configurations for the discussed observers (\texttt{FeatureObserver}s, \texttt{RewardObserver}) and updaters (\texttt{GraphUpdater})---is the primary mechanism for their customization. This allows researchers to flexibly define and experiment with the different \ac{SMDP} components discussed in Chapter \ref{ch5}.

\subsection{The \texttt{SingleJobShopGraphEnv}}
\label{subsec:single_job_shop_graph_env}
The \href{https://job-shop-lib.readthedocs.io/en/stable/api/job_shop_lib.reinforcement_learning.html#job_shop_lib.reinforcement_learning.SingleJobShopGraphEnv}{\texttt{SingleJobShopGraphEnv}} class provides a complete \ac{RL} environment for a specific \ac{JSSP} instance represented by a \texttt{JobShopGraph}. In particular, it manages the \ac{RL} interaction loop by defining the \ac{SMDP} for a particular problem. It is initialized with the following arguments:
\begin{sloppypar}
\begin{itemize}[itemsep=0pt, topsep=0pt]
    \item \texttt{job\_shop\_graph}: A pre-constructed \texttt{JobShopGraph} object representing the specific \ac{JSSP} instance to be solved. This graph contains the initial set of nodes $V_0$ and edges $E_0$.
    \item \texttt{feature\_observer\_configs}: A sequence of configurations (instances of \texttt{DispatcherObserverConfig} or identifiers for built-in observers) that define which \texttt{FeatureObserver}s will be used to compute the node features $\mathbf{X}_k$.
    \item \texttt{reward\_function\_config}: A \texttt{DispatcherObserverConfig} specifying the type and parameters of the \texttt{RewardObserver} subclass that will calculate the reward $r_k$.
    \item \texttt{graph\_updater\_config}: A \texttt{DispatcherObserverConfig} that defines the \texttt{GraphUpdater} subclass responsible for managing the graph topology $(V_k, E_k)$ transitions.
    \item \texttt{ready\_operations\_filter} (optional): A callable that filters the set of operations considered available for scheduling at each step, influencing the action space $A(s_k)$. Defaults to \texttt{filter\_dominated\_operations}. See Subsection \ref{subsec:ready_op_filters} for more information.
\end{itemize}
Internally, the environment instantiates a \texttt{Dispatcher} to manage the scheduling process. The provided \texttt{feature\_observer\_configs} are used to create a \texttt{CompositeFeatureObserver}, which aggregates all specified node features. Similarly, the \texttt{reward\_function\_config} and \texttt{graph\_updater\_config} are used to instantiate the chosen \texttt{RewardObserver} and \texttt{GraphUpdater} respectively, linking them to the \texttt{Dispatcher}. These configurations effectively allow the user to define the core components of the \ac{SMDP}.
\end{sloppypar}
\subsubsection{Observation Space}
\label{subsec:single_env_state_and_obs}
In the context of the classical \ac{JSSP}---which we are tackling, the true state $s_k$ of the environment can be thought of as the observation the \ac{RL} agent receives. Unless the graph representation or features are flawed, no information is hidden from the agent. In the \texttt{SingleJobShopGraphEnv}, this observation is a dictionary adhering to the \texttt{gymnasium.spaces.Dict} structure. \footnote{This observation is not meant to be used as it is, but wrapped to accommodate the needs of the deep learning framework of choice (e.g., \ac{PyG} \citep{pyg}). The reason is that we want our library to be framework-agnostic. More information about this wrapper is available in the next chapter.}. Its keys and typical contents are:
\begin{sloppypar}
\begin{itemize}[itemsep=0pt, topsep=0pt]
    \item \texttt{"removed\_nodes"}: A NumPy array of booleans, indicating which nodes from the initial \texttt{JobShopGraph} are currently considered inactive or removed (e.g., completed operations). The length of this array corresponds to the total number of nodes in the original graph.
    \item \texttt{"edge\_index"}: A $2 \times N_e^{(k)}$ NumPy array of integers, representing the current graph's edge list in COO (Coordinate) format, where $N_e^{(k)}$ is the number of active edges at step $k$. Each column $(u, v)$ denotes a directed edge from node $u$ to node $v$. This represents $E_k$.
    \item \texttt{"operations"}, \texttt{"jobs"}, \texttt{"machines"}: These (optional) keys map to NumPy arrays containing the feature matrices for operations, jobs, and machines, respectively. They are computed by the \texttt{CompositeFeatureObserver} and collectively form the future node feature matrix $\mathbf{X}_k$. Their shapes are (current number of active entities of that type, number of features for that type). Handling these matrices separately is helpful. For example, if using \ac{PyG}, we 
\end{itemize}
\end{sloppypar}

One of the key differences between graph environments and traditional ones is the dynamic size of the data. Most \ac{RL} frameworks, such as Stable Baselines3 \citep{2021stable-baselines3}, are designed to work with static-size data. To help with these requirements, we provide the option to pad the observation's tensors with the \texttt{use\_padding=True} argument. When active, all variable-size arrays in the observation (like \texttt{"edge\_index"} and the feature matrices) are padded (e.g., with -1, or `True` for \texttt{"removed\_nodes"} representing non-existent padded nodes) to match the maximum possible dimensions derived from the initial graph state. This ensures a consistent observation shape, although it will need to be removed in a subsequent step to be fed into a GNN.

\subsubsection{Action Space}
\begin{sloppypar}
The action space $\mathcal{A}(s_k)$ available to the agent at state $s_k$ is defined by the set of operations that can be scheduled next. The \texttt{action\_space} attribute of the environment is a \texttt{gymnasium.spaces.MultiDiscrete([num\_jobs, num\_machines])}. An action $a_k$ selected by the agent is a tuple $(j, m)$, where:
\begin{itemize}[itemsep=0pt, topsep=0pt]
    \item $j$ is the index of the job from which an operation is to be scheduled.
    \item $m$ is the index of the machine on which this operation is to be performed. A special value of $m = -1$ can be used if the chosen operation has only one candidate machine.
\end{itemize}
\end{sloppypar}

\subsubsection{Environment Dynamics}
At the beginning of an episode, the system calls the \texttt{reset()} method, which reinitializes the environment to its starting state $s_0$ for the current \ac{JSSP} instance. This process involves resetting the \texttt{Dispatcher} by clearing any partial schedule and reinitializing all subscribed observers (including those that track features, graph configurations, and rewards). The \texttt{JobShopGraph} is also restored to its initial configuration. The method returns both the initial observation $s_0$ and an information dictionary containing feature names and a list of initially available actions.

The dynamics of the environment, i.e., how the state transitions and how rewards are generated, are primarily managed by the \texttt{step(action)} method.
When an action $a_k = (j,m)$ is taken:
\begin{enumerate}[itemsep=0pt, topsep=0pt]
    \item The \texttt{Dispatcher} attempts to schedule the corresponding operation. This involves updating its internal representation of the schedule. This is a core part of the state transition $s_k \rightarrow s_{k+1}$.
    \item The configured \texttt{GraphUpdater} is notified of the scheduled operation and modifies the \texttt{JobShopGraph} accordingly (e.g., by removing nodes or edges, as per residual scheduling principles). This updates the graph structure from $(V_k, E_k)$ to $(V_{k+1}, E_{k+1})$.
    \item The \texttt{FeatureObserver}s (within the \texttt{CompositeFeatureObserver}) update their respective feature values based on the scheduled operation and the current time, producing the new feature matrix $\mathbf{X}_{k+1}$.
    \item The configured \texttt{RewardObserver} calculates the reward $r_k$ for the taken action $a_k$ leading to state $s_{k+1}$. This reward is then available via the observer's \texttt{last\_reward} property.
\end{enumerate}
The \texttt{step()} method returns a tuple: (\texttt{observation} $o_{k+1}$, \texttt{reward} $r_k$, \texttt{terminated}, \texttt{truncated}, \texttt{info}). The \texttt{observation} $o_{k+1}$ is the new observation dictionary reflecting $s_{k+1}$.

The \texttt{terminated} boolean flag, returned by the \texttt{step()} method, becomes \texttt{True} when the episode reaches its natural conclusion—specifically when all operations in the \ac{JSSP} instance have been scheduled. The \texttt{truncated} boolean flag, also returned by \texttt{step()}, indicates whether external conditions prematurely ended the episode, such as time or step limits. Since the current implementation does not support step limits, this flag always remains \texttt{False}. This second indicator was added to maintain compatibility with the Gymnasium interface.

\subsubsection{Rendering}
The \texttt{SingleJobShopGraphEnv} supports visualization of the scheduling process through its \texttt{render()} method. This is primarily useful for debugging and understanding agent behavior. Supported rendering modes include:
\begin{itemize}[itemsep=0pt, topsep=0pt]
    \item \texttt{"human"}: Displays the current Gantt chart of the partial or complete schedule.
    \item \texttt{"save\_video"}: Saves a video of the entire scheduling episode as a Gantt chart animation.
    \item \texttt{"save\_gif"}: It is similar to the \texttt{"save\_video"}, but saves a GIF instead. An example of the result of using this format can be seen in the \href{https://github.com/Pabloo22/job_shop_lib?tab=readme-ov-file#solve-an-instance-with-a-dispatching-rule-solver}{\texttt{README.md} file of the \textit{JobShopLib}'s GitHub page.}
\end{itemize}
These rendering capabilities are facilitated by the \href{https://job-shop-lib.readthedocs.io/en/stable/api/job_shop_lib.visualization.gantt.html#job_shop_lib.visualization.gantt.GanttChartCreator}{\texttt{GanttChartCreator}} class (detailed in the API documentation), which typically uses a \texttt{HistoryObserver} attached to the \texttt{Dispatcher} to record the sequence of scheduled operations. The appearance and output paths for these visualizations can be configured using a \texttt{RenderConfig} dictionary passed during environment initialization.

\subsection{The \texttt{MultiJobShopGraphEnv}}
\label{subsec:multi_job_shop_graph_env}
While \texttt{SingleJobShopGraphEnv} is suited for a single problem instance, the \href{https://job-shop-lib.readthedocs.io/en/stable/api/job_shop_lib.reinforcement_learning.html#job_shop_lib.reinforcement_learning.MultiJobShopGraphEnv}{\texttt{MultiJobShopGraphEnv}} is designed for training \ac{RL} agents that can generalize across a variety of \ac{JSSP} instances.

The primary role of \texttt{MultiJobShopGraphEnv} is to provide an environment where each episode can potentially involve a different \ac{JSSP} instance. This is crucial for developing robust agents that do not overfit to a specific problem structure.
Key initialization arguments include:
\begin{itemize}[itemsep=0pt, topsep=0pt]
    \item \texttt{instance\_generator}: An instance of a \texttt{InstanceGenerator} subclass (e.g., \texttt{InstanceGenerator}, responsible for generating new \texttt{JobShopInstance} objects when \texttt{reset()} is called. This generator also defines the maximum possible size of instances (e.g., max jobs, max machines), which is used to set up consistent observation and action spaces. This can be useful for the reasons mentioned in the previous subsection's footnote.
    \item \texttt{feature\_observer\_configs}, \texttt{reward\_function\_config}, \texttt{graph\_updater\_config}: Similar to \texttt{SingleJobShopGraphEnv}, these configure the observers and updaters that define the \ac{SMDP} components for each generated instance.
    \item \texttt{graph\_initializer}: A callable that takes a \texttt{JobShopInstance} and returns a \texttt{JobShopGraph} (e.g., the \texttt{build\_resource\_task\_graph} function mentioned in Section \ref{sec:graph_representation}).
\end{itemize}
These arguments ensure that while instances may vary, the way their states are represented, rewards are calculated, and graphs are updated follows a consistent (configurable) logic.

\subsubsection{Handling Multiple Instances and the SMDP}
As we mentioned, the \texttt{MultiJobShopGraphEnv} internally manages an instance of \texttt{SingleJobShopGraphEnv}. The core difference in its operation, particularly concerning the \ac{SMDP}, lies in the \texttt{reset()} method.

When \texttt{reset()} is called, the \texttt{instance\_generator} is invoked to create a new \texttt{JobShopInstance}. This defines a new initial state for a new scheduling problem. The \texttt{graph\_initializer} function is then used to build the initial \texttt{JobShopGraph} for this newly generated instance. Finally, the internal \texttt{SingleJobShopGraphEnv} is reconfigured with this new graph and instance data, along with the same observer configurations.

Within each episode (i.e., between calls to \texttt{reset()}), the \ac{SMDP} elements (state $s_k$, action $a_k$, reward $r_k$, transition $P(s_{k+1}|s_k, a_k)$) are handled by the internal \texttt{SingleJobShopGraphEnv} for the currently active \ac{JSSP} instance, as described in Section \ref{subsec:single_job_shop_graph_env}. 

The only difference in behavior with the underlying \texttt{SingleJobShopGraphEnv} is that additional padding may be needed to ensure shape consistency across instances of varying sizes (e.g., different numbers of jobs, machines, or operations). The observation and action spaces of \texttt{MultiJobShopGraphEnv} are defined based on the maximum possible instance size, as specified by the \texttt{instance\_generator} (e.g., \texttt{max\_num\_jobs}, \texttt{max\_num\_machines}). Thus, if using the \texttt{use\_padding=True} mechanism, observations from smaller instances are padded to match the dimensions of the largest possible instance. This consistency in shape is added for the same reasons described in Subsection \ref{subsec:single_env_state_and_obs}.

%% file: chapters/08-ExperimentsAndResults.tex
\doublespacing 

\chapter{Experiments and Results}
\label{ch8}

\begin{spacing}{1} 
\minitoc 
\end{spacing} 
\thesisspacing 

This chapter contains experiments that prove the environment's capabilities\footnote{The code employed for conducting these experiments is available in the \href{https://github.com/Pabloo22/gnn_scheduler}{\texttt{gnn\_scheduler}} GitHub repository.} that we mentioned in the previous chapter. These experiments also serve as examples to illustrate how \textit{JobShopLib}'s components can be used to train a GNN-based dispatcher. We also mention some data about the community's adoption of \textit{JobShopLib}.

We conducted this project's experiments using behavioral cloning---a form of \ac{IL} mentioned in Section \ref{sec:il}. This learning method offers us a simpler way to train a deep learning model than \ac{RL} does. For example, training is usually more stable than with \ac{RL} methods. Additionally, the state-of-the-art at the time of writing \citep{lee2024il_jssp} uses this learning method. In fact, they showed that using \ac{RL} yielded slightly worse results in its ablation study.

\begin{sloppypar}
Because \ac{BC} uses supervised learning, we must create a dataset of observation-action pairs. This process is achieved in two steps. First, we use a \ac{CP} solver to obtain optimal schedules $\mathbf{Y}^*$ for small randomly generated \ac{JSSP} instances. The hope is that the deep learning model will generalize from small instances to bigger ones, in which using this kind of exact solver becomes prohibitively expensive.

Once we have solved these instances, we use the \texttt{SingleJobShopGraphEnv} described in the previous chapter to create the pairs of observations and optimal actions needed to train the model using traditional supervised learning. In this case, the action space for each state is the set of available operations to be dispatched. We need to create a mapping from these operations to one or zero, depending on whether the operation is optimal or not. This mapping is created by checking if dispatching each operation maintains the same order of operations of the optimal schedule $\mathbf{Y}^*$ found by the \ac{CP} solver.
\end{sloppypar}

The first experiment shows that the features computed by the \textit{JobShoplib} built-in feature observers are expressive enough to explain most of the performance of a GNN-based dispatcher. To prove this, we removed the message passing mechanism from the GNN architecture. Thus, the model processes each available operation independently. In other words, only its features are considered; the graph structure is a fully disconnected graph. This experiment also serves as a baseline for evaluating the advantage of using a graph representation of the problem.

In the second experiment, we trained a GNN, particularly the GATv2 model described in Subsection \ref{subsec:gat}. The graph representation used is the standard Resource-Task graph used by \cite{park2021schedule_net}. The results show an improvement over their results, especially in large instances. 




\section{Dataset Generation}
\label{sec:dataset_generation}

As we mentioned, before training a deep learning model to predict whether an operation is optimal to dispatch, we first need to generate a dataset of (observation, action label) pairs. This process leverages several \textit{JobShopLib}'s components and can be divided into several steps, which are explained in the following subsections.

\subsection{Generating Optimal Schedules}
\begin{sloppypar}
Initially, we generated a diverse set of \textit{small} \ac{JSSP} instances to construct our training dataset. For this, we utilized the \texttt{GeneralInstanceGenerator} class from \textit{JobShopLib} (see Chapter \ref{ch6}). Table \ref{tab:training_instance_distribution} details the specific configurations.
\end{sloppypar}
\begin{table}[H]
\centering
\begin{tabular}{c|c|c}
\textbf{Number of Jobs} & \textbf{Number of Machines }& \textbf{\# Instances}\\
\hline
10 & 5  & 175,000 \\
8 & 8  & 100,000 \\
10 & $\sim U(5, 10)$  & 100,000 \\
10 & 10  & 50,000 \\
$\sim U(10, 15)$ & $\sim U(5, 10)$  & 5,000 \\
\hline
\multicolumn{2}{c|}{\textbf{Total Instances}} & 430,000 \\
\end{tabular}
\caption[Distribution of instances used for training.]{Distribution of instances used for training. $\sim U(a, b)$ indicates that the integer used is sampled from a uniform distribution from the range $[a, b]$. The number of operations per job is always the number of machines. Additionally, recirculation is not allowed; each job visits each machine exactly once. Finally, operation's processing times are sampled from $\sim U(1, 99)$ for all instances.}
\label{tab:training_instance_distribution}
\end{table}
We compute the optimal schedule for each of these randomly generated \ac{JSSP} instances. This is achieved by employing an exact solver, specifically the \texttt{ORToolsSolver} provided within \textit{JobShopLib}'s constraint programming module. The resulting \texttt{Schedule} objects, which contain both the instance and its optimal solution, are saved into JSON files using their \texttt{to\_dict()} method.

\subsection{Defining the Observation}
\begin{sloppypar}
The second, and most crucial, step involves transforming these optimal schedules into a sequence of (observation, action label) pairs suitable for supervised learning. This is where the \texttt{SingleJobShopGraphEnv} (detailed in Chapter \ref{ch7}) plays a central role. The environment is configured with the specific \textit{solved} \ac{JSSP} instance, a graph builder function (e.g., the \texttt{build\_resource\_task\_graph}), and a collection of feature observers (as discussed in Section \ref{sec:computing_node_features}) to dynamically compute node features.
\end{sloppypar}

This project uses Resource-Task graphs and the residual scheduling strategy described in Subsection \ref{subsec:residual_scheduling}. As argued in Subsection \ref{subsec:resource_task_graphs}, Resource-Task graphs are typically more computationally efficient than disjunctive graphs. By explicitly modeling machines as nodes, they can also be more expressive than traditional disjunctive graphs. We also use residual scheduling updates to both node features and the graph structure to remove irrelevant or noisy information. This is needed since everything behind the current time cannot be changed and, thus, should not affect future scheduling decisions.

\noindent The features used to represent each node---computed using their respective feature observers---are:

\noindent \textbf{Operation nodes:}
    \begin{itemize}[itemsep=0pt, topsep=0pt]
        \item \textbf{Is scheduled}: Indicates if the operation has been dispatched.
        \item \textbf{Earliest start time:} The earliest possible start time for the operation relative to the current time ($R^{(k)}_{ij} = S^{*(k)}_{ij} - t_k$).
        \item \textbf{Remaining processing time:} The operation's full processing time, updated to reflect its actual remaining duration ($C_{ij} - t_k$) once scheduled.
        \item \textbf{Position in job:} The original position ($j$) of an operation $O_{ij}$ adjusted by subtracting the number of already scheduled operations in that job.
        \item \textbf{Job-related features:} Features of the job to which the operation belongs are also used. These include:
        \begin{itemize}[itemsep=0pt, topsep=0pt]
            \item Whether the job has an operation currently scheduled or not.
            \item Job earliest start time (based on the next unscheduled operation in that job).
            \item Job remaining processing time (sum of durations of its unscheduled operations).
            \item Job remaining operations (count of its unscheduled operations).
        \end{itemize}
    \end{itemize}

\noindent \textbf{Machine nodes:}
    \begin{itemize}[itemsep=0pt, topsep=0pt]
        \item \textbf{Is scheduled}: Indicates if the machine has at least one operation scheduled but not yet completed.
        \item \textbf{Earliest start time:} The minimum earliest start time ($S^{*(k)}_{ij} - t_k$) among unscheduled operations on that machine.
        \item \textbf{Remaining processing time:} The aggregate sum of durations of unscheduled operations on the machine.
        \item \textbf{Remaining operations:} The count of unscheduled operations for the machine.
    \end{itemize}
\noindent All these features are also normalized by dividing them by the maximum absolute value present in the observation for all the nodes of the same type. Since all features are positive, all features end in the $[0, 1]$ range. This normalization ensures a consistent scale across all instances and features.

\subsection{Wrapping \textit{JobShopLib}'s Observation for PyG Compatibility}
\label{subsec:wrapping_obs}
A crucial consideration at this stage is that the raw observation from \texttt{SingleJobShopGraphEnv} (a dictionary containing global node features, a global edge index, and a list of removed nodes, see Subsection \ref{subsec:single_env_state_and_obs}) may not be directly compatible with GNN libraries such as \ac{PyG}, especially when dealing with heterogeneous graphs. \ac{PyG}'s \texttt{HeteroData} objects, which are ideal for representing such graphs, expect node features and edge indices to be structured in separate dictionaries, keyed by node type and edge type (a tuple of source node type, relation type, and destination node type), respectively. Furthermore, these libraries often require node indices within each type to be contiguous and start from zero.

\begin{sloppypar}
We employ an observation wrapper (\href{https://job-shop-lib.readthedocs.io/en/stable/api/job_shop_lib.reinforcement_learning.html#job_shop_lib.reinforcement_learning.ResourceTaskGraphObservation}{\texttt{ResourceTaskGraphObservationDict}}) to bridge this gap. This wrapper takes the raw observation from the \texttt{SingleJobShopGraphEnv} and transforms it into the \texttt{ResourceTaskGraphObservationDict} format, which can be directly used by \ac{PyG}. Specifically, the \texttt{ResourceTaskGraphObservation.observation()} method performs the following key transformations:
\begin{itemize}[itemsep=0pt, topsep=0pt]
    \item It converts the global \texttt{edge\_index} (a single NumPy array) into an \texttt{edge\_index\_dict}. In this dictionary, keys are tuples representing edge types (e.g., \texttt{("operation", "to", "machine")}), and values are the edge index NumPy arrays for that specific relation.
    \item It organizes node features from the environment's observation into a \texttt{node\_features\_dict}. Here, keys are node type names (e.g., \texttt{"operation"}, \texttt{"machine"}), and values are the feature matrices (NumPy arrays) for the currently active nodes of that type.
    \item It handles removed nodes (e.g., completed operations) by filtering them out from the feature matrices, ensuring that only features of active nodes are included.
    \item Crucially, it remaps node identifiers. The original graph in \texttt{SingleJobShopGraphEnv} considers global node IDs. The wrapper ensures that nodes of each type are indexed locally within the processed observation, starting from 0 (e.g., machine and operation nodes will have indices independent from each other). It also maintains an \texttt{original\_ids\_dict} to keep track of the original indices of the active nodes relative to their type's initial set of nodes.
\end{itemize}
With these changes, the \texttt{ResourceTaskGraphObservation} wrapper ensures that the observations can be directly used by \ac{PyG}'s \texttt{HeteroData} objects.
\end{sloppypar}

\subsection{Sampling Observation-Label Pairs}
\label{subsec:sample-n-steps}
With the wrapped environment, we ``replay" each optimal schedule. At each step of this replay, we identify the optimal action(s) using the \href{https://job-shop-lib.readthedocs.io/en/stable/api/job_shop_lib.dispatching.html#job_shop_lib.dispatching.OptimalOperationsObserver}{\texttt{OptimalOperationsObserver}}, initialized with the complete optimal \texttt{Schedule} object. An available operation is considered optimal if dispatching it at the current step respects the sequence of jobs in each machine described by the optimal solution matrix $\mathbf{Y^*}$.

For each available operation in a step, we encode labels as binary values: 1 for optimal operations (those that respect the optimal solution's sequencing) and 0 for non-optimal operations. This binary encoding allows us to formulate the problem as a standard binary classification task.

An important observation, however, is that the graphs and labels observed from step $k$ to $k+1$ are often highly similar (i.e., $(G_k, \mathbf{y}_k) \simeq (G_{k+1}, \mathbf{y}_{k+1})$). Training on such codependent samples can lead the model to “memorize" the answer for step $k$, and then use this knowledge to predict the value for step $k+1$. Thus, the model will not be learning new information from $(G_{k+1}, y_{k+1})$, which could hinder learning by promoting overfitting. Furthermore, storing every step for all instances would create an impractically large dataset. For our 430,000 instances (Table~\ref{tab:training_instance_distribution}), this could amount to approximately 28 million samples\footnote{Given $\sim U(a, b)$, assuming an average instance size of $\frac{a + b}{2}$: $175,000 \cdot 10 \cdot 5 + 100,000 \cdot 8 \cdot 8 + 100,000 \cdot 10 \cdot 7.5 + 50,000 \cdot 10 \cdot 10 + 5,000 \cdot 12.5 \cdot 7.5 = 28,118,750$. Note, however, that observation-label pairs that only have one available operation are skipped, so the number would be lower in practice.}.

To address these issues, we adopt a sampling strategy: we store an observation-label tuple in our training dataset only after every $n$ replay steps. This reduces dataset size by a factor of $n$, often enabling it to fit in memory and thereby simplifying and speeding up training. More importantly, it mitigates the codependence between samples.

To further ensure diversity and prevent always sampling the same relative steps (e.g., always the initial steps) from all instances, we employ a global step counter, $C_{\text{global}}$. This counter is initialized to zero at the beginning of the dataset generation process and accumulates the total number of elementary steps processed across all preceding instances. For the current instance being replayed, an operation at its internal step $k$ (0-indexed) is considered for sampling if the condition $(k + C_{\text{global}}) \bmod n = 0$ is met. After the replay of an instance with $T$ total steps is completed, $C_{\text{global}}$ is incremented by $T$.

The value of $C_{\text{global}}$ means that the specific set of internal steps $k$ sampled from any given instance depends on $(\sum T_{\text{prior}}) \bmod n$, where $T_{\text{prior}}$ are the lengths of all preceding instances. The evolution of this sampling phase, which is key to ensuring diversity across the dataset, critically depends on the length $T$ of the instance just processed. Specifically, after $C_{\text{global}}$ is incremented by $T$, the sampling phase $C_{\text{global}} \bmod n$ for the \textit{subsequent} instance will differ from the current instance's phase if and only if $T \bmod n \neq 0$.
If $T \bmod n = 0$ (i.e., $n$ divides $T$), then that particular instance does not shift the sampling phase. Should several consecutive instances all have lengths that are multiples of $n$, the set of sampled internal steps could be identical for these instances relative to their own start, leading to a form of aliasing.
Therefore, the overall effectiveness of this strategy in promoting a well-distributed set of sampled steps across different stages of the schedules hinges on the instance lengths $T_j$ in the dataset frequently satisfying $T_j \bmod n \neq 0$. When this condition holds (e.g., by setting $n$ to be a prime number), it helps prevent concentrating samples on specific steps of all instances.

\noindent For example, let $n = 31$ and a dataset of two \ac{JSSP} instances of size $10\times10$ created with the procedure described in the previous section.
\begin{itemize}[itemsep=0pt, topsep=0pt]
    \item Consider the first instance, having $T_1 = 100$ steps to replay. Initially, $C_{\text{global}} = 0$. We sample internal steps $k$ where $(k + 0) \bmod{31} = 0$. Thus, steps $k \in \{0, 31, 62, 93\}$ are sampled. After processing this instance, $C_{\text{global}}$ becomes $0 + 100 = 100$. (Here $T_1 \bmod{31} = 7 \neq 0$, so the phase will shift).
    \item For the second instance, The inherited $C_{\text{global}}$ is $100$. We sample internal steps $k$ where $(k + 100) \bmod{31} = 0$. Since $100 \bmod{31} = 7$, this condition is equivalent to $(k+7) \bmod{31} = 0$. Thus, steps $k \in \{24, 55, 86\}$ are sampled. After this instance, $C_{\text{global}}$ becomes $100 + 100 = 200$.
\end{itemize}
\noindent Conversely, consider a scenario where $n=50$. If the first instance has $T_1=100$ steps, then $T_1 \bmod{50} = 0$. For this instance with $C_{\text{global}} = 0$, we sample steps $k$ where $(k+0)\bmod{50}=0$, yielding samples at $k \in \{0, 50\}$. After processing, $C_{\text{global}}$ becomes $100$. For the second instance, we sample steps $k$ where $(k+100)\bmod{50}=0$. Since $100 \bmod{50}=0$, this is equivalent to $(k+0)\bmod{50}=0$, resulting in samples at the same relative positions $k \in \{0, 50\}$. This shows how, when $T \bmod n = 0$, the sampling phase remains unchanged between consecutive instances, potentially leading to the aliasing problem mentioned earlier.

In short, when we ensure $T_j \bmod n \neq 0$, this mechanism effectively distributes the sampling points, ensuring that different parts of various instances contribute to the training dataset.

\noindent This sampling process ultimately yields a dataset where each entry consists of:
\begin{itemize}[itemsep=0pt, topsep=0pt]
\item A heterogeneous graph observation representing the current state of the JSSP instance (suitable for \ac{PyG}).
\item Binary labels for each available operation indicating whether it is optimal (1) or non-optimal (0) according to the pre-computed optimal schedule.
\end{itemize}
\noindent This dataset is significantly more manageable in size while preserving the diversity needed for effective model training.

\section{Model Architectures}
As we mentioned, we trained two \ac{DL}-based dispatchers. Except for the message-passing stage, both models undergo conceptually the same phases.

\subsection{Feature Embeddings}
After extracting the node features for our graph representation with the aforementioned feature observers, an important consideration is how to represent these numerical features within the neural network architecture. While traditional approaches might directly use these scalar values as inputs, we implement feature embeddings to enhance the expressiveness of our model.

Feature embeddings, initially proposed for applying deep learning to tabular data \citep{gorishniy2022feature_embeddings}, transform scalar values of numerical features into high-dimensional vector representations before processing them in the main backbone of the network. Formally, for each numerical feature $x_{i,j} \in \mathbf{x}_i \in \mathbb{R}^F$ of node $i$ with $F$ total features, we define an embedding function:
$$\mathbf{z}_{i,j} = f_j(x_{i,j}) \in \mathbb{R}^{d_j}$$
where $f_j: \mathbb{R} \rightarrow \mathbb{R}^{d_j}$ is the embedding function for the $j$-th feature, $\mathbf{z}_{i,j}$ is the resulting embedding vector, and $d_j$ is the dimensionality of the embedding.

In our implementation, we use the \textbf{periodic encoder} described in \cite{gorishniy2022feature_embeddings}. It transforms each scalar feature using sinusoidal functions with learnable frequency parameters:
$$f_j(x_{i,j}) = \left[\sin(2\pi c_{j1}x_{i,j}), \cos(2\pi c_{j1}x_{i,j}), \ldots, \sin(2\pi c_{jK}x_{i,j}), \cos(2\pi c_{jK}x_{i,j})\right]$$
where $c_{jk}$ are learnable frequency parameters for feature $j$ and frequency component $k$, $K$ is the number of frequency components used for each feature, and $[,]$ denotes concatenation. These periodic embeddings are then linearly transformed to the desired output dimension. The complete node representation after embedding becomes:
$$\mathbf{h}_i^{(0)} = \left[\mathbf{z}_{i,1}, \mathbf{z}_{i,2}, \ldots, \mathbf{z}_{i,F}\right]$$
where $\mathbf{h}_i^{(0)}$ represents the initial hidden state of node $i$, and $[,]$ denotes concatenation.

Our implementation creates separate periodic embeddings for each input feature and then concatenates them. For each feature, we use multiple frequency components initialized from a normal distribution with a configurable standard deviation $\sigma$ (we used $\sigma = 1$ in all the experiments). The number of frequency components is determined by the desired output embedding size divided by twice the number of input features (since each frequency component produces both sine and cosine values).

These feature embeddings are learned end-to-end along with the rest of the model parameters during training. After the embedding stage, these enhanced node representations are passed to the message-passing layers in our GNN architecture (or directly to the readout function in the case of our non-message-passing baseline).

\subsection{Message Passing}
\label{subsec:mp}
This is the stage in which the different types of architectures differ the most. In our experiments, we explore three distinct approaches. In this subsection, we describe the best configurations of hyperparameters after a manual fine-tuning.\footnote{More details on the specific hyperparameter tested are available in the \href{https://github.com/Pabloo22/gnn_scheduler/tree/main/gnn_scheduler/configs}{configs} folder from the \href{https://github.com/Pabloo22/gnn_scheduler/tree/main/gnn_scheduler}{gnn\_scheduler} repository.}

The message passing block described in this subsection consists of stacking GNN layers. First, the Resource-Task graph is divided into subgraphs that only contain one relationship type (e.g., (“operation," “to," “machine")). Each subgraph is processed by one of the classical GNN layers described in Section \ref{sec:message_passing}. Each of them uses a different set of weights. Then, if a node is connected to other nodes by more than one relationship type (e.g., an operation node is typically related to machine nodes and other operations), the messages coming from each relationship are aggregated.

During this block, we optionally remove randomly a percentage of edges during training---a concept known as “edge dropout." This removal can help us reduce overfitting by forcing the GNN not to rely on individual nodes.

This processing is followed by the \ac{ELU} activation function. Similarly to ReLU, this function mitigates the vanishing gradient problem\footnote{The vanishing gradient problem occurs when gradients become extremely small as they propagate backward through many layers, making earlier layers learn very slowly or not at all.} by not modifying positive values, but has the advantage of not setting negative values to zero. It transforms these negative values by $\alpha (\exp(x) - 1)$. In our case, we used $\alpha=1$ for all the experiments.

\begin{figure}
    \centering
    \includegraphics[width=0.75\linewidth]{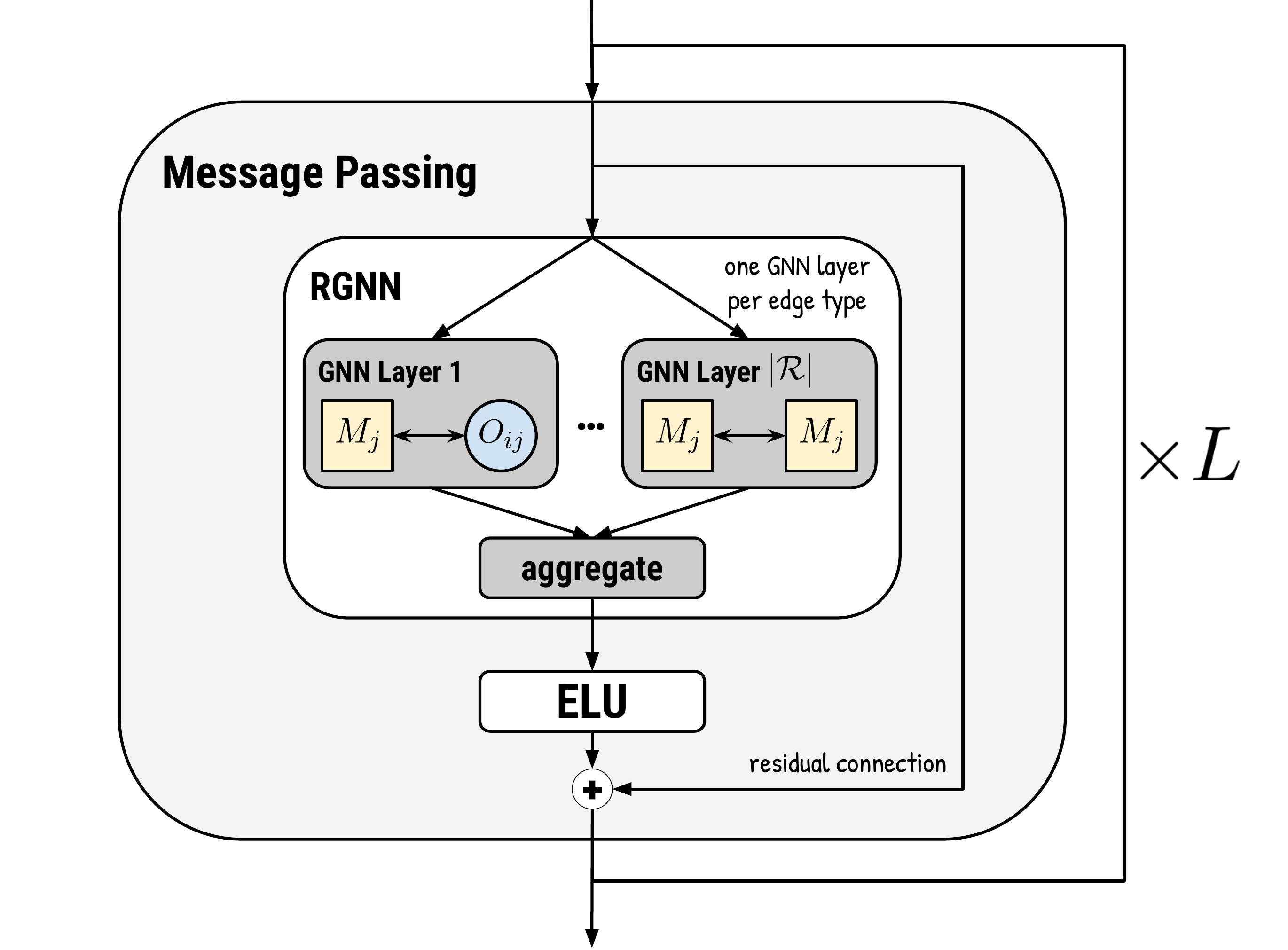}
    \caption[Diagram showing a message passing block.]{Diagram showing a message passing block.}
    \label{fig:mp-block}
\end{figure}

Finally, the vector before this computation is added to the final result (i.e., a residual connection). This trick, introduced in the ResNet architecture \citep{he2016residual}, also mitigates the vanishing gradient problem by allowing gradients to “skip" layers. Then, this message passing block is repeated $L$ times, as illustrated in Figure \ref{fig:mp-block}.

\subsubsection{Non-Message-Passing Baseline}
Our first approach serves as a baseline by completely removing the message passing mechanism. In this configuration, node features are processed independently without any information exchange between nodes in the graph. After the embedding stage described previously, the node representations are passed directly to the readout function. This configuration allows us to evaluate how much of the model's performance depends solely on the expressiveness of the node features versus the structural information captured through message passing.

\subsubsection{Relational Graph Isomorphism Network}
For our second approach, we implement a \ac{RGIN} to handle heterogeneous graphs with the following configuration:

The RGIN layer first transforms node features using type-specific MLPs, then performs message passing through its neighbors. In the heterogeneous context, this means each node type has its own MLP, and messages are passed and aggregated separately for each edge type. The aggregation combines all the aggregated messages a node receives using the specified function (in this case, element-wise maximum). Adjusting the computation detailed in Subsection \ref{subsec:relational_mp} to use the max as the final aggregator, we get:
$$ \mathbf{h}_i^{(l+1)} = \max_{r \in \mathcal{R}} \underbrace{\left( \text{MLP}_r^{(l)} \Bigg((1 + \epsilon_r^{(l)})\mathbf{h}_i^{(l)} + \sum_{j \in \mathcal{N}_i^r} \mathbf{h}_j^{(l)}\Bigg) \right)}_{\text{Output of GIN layer for relation } r}.$$

While in Subsection \ref{subsec:gin} we argued that the sum aggregator is the most expressive one, it is important to note that this is not always a good feature. In our case, we train with small problem instances, and then we apply the same GNN to bigger ones. Ideally, the GNN will learn generalizable patterns from small problems to bigger ones. The sum aggregator likely hinders this task because it can discriminate better between different neighbor sizes. The max aggregator, on the other hand, is less sensitive to the number of neighbors. It also aligns better with dynamic programming \citep{dudzik2022gnns_are_dps}, which can also help the GNN approximate this algorithm more easily. Thus, it is usually the recommended aggregator in tasks that require generalization to bigger graphs, as argued in \cite{dudzik2022gnns_are_dps}.

\subsubsection{Relational Graph Attention Network v2}
A second family of experiments uses the \ac{RGATv2} architecture, which was detailed in Chapter \ref{ch3}. We also used the max aggregator to combine the messages from each relationship type for the aforementioned reasons. Substituting the general aggregator $\bigoplus$ used in Equation \ref{eq:rgat}, we obtain:
$$\mathbf{h}_i^{(l+1)} = \max_{r \in \mathcal{R}} \underbrace{\left( \sigma_r\Bigg(\sum_{j \in \mathcal{N}_i^r \cup \{i\}} \alpha_{ij}^{(l,r)} \mathbf{W}_r^{(l)}\mathbf{h}_j^{(l)}\Bigg) \right)}_{\text{Output of GATv2 layer for relation } r}$$
where the attention coefficients $\alpha_{ij}^{(l,r)}$ are obtained using the GATv2 methodology described in Subsection \ref{subsec:gat}.

\noindent Table \ref{tab:mp_hparams} summarizes the hyperparameters used in the best architectures for each family of methods (“Non-Message-Passing", “\ac{RGIN}", “\ac{RGATv2}"). We consider the best architecture the one with the lowest average makespan across all benchmark instances described in Subsection \ref{subsec:benchmark_instances}.\footnote{Note that using this selection process biases the results to choose models that perform better in larger instances. For example, a performance difference of 10 in the \texttt{ft06} instance, a problem of size $6x6$, is significant in relative terms but it will probably have little impact on average makespan. On the other hand, small relative improvements in larger instances could mean a difference in makespan of 100 (i.e., ten times more impact). This is a conscious choice because the instances in which GNN-based dispatchers would have a greater advantage over other methods will be in large instances. Similarly, it is in large instances where even small improvements can be more noticeable, economically speaking. While the range of operation's processing times could also distort this metric, all benchmark instances have similar ranges (around $[1, 99]$).}

\begin{table}[hbtp]
\centering
\begin{tabular}{c|c|c|c}
\textbf{Feature} & \textbf{Non-Message-Passing} & \textbf{\ac{RGIN}} & \textbf{\ac{RGATv2}} \\
\hline
Hidden dimensions ($d_{\text{hidden}}$)& N/A & 32 & 128 \\

Number of layers ($L$)& N/A & 2 & 2 \\

Relationship types aggregator & N/A & Max & Max \\

Edge dropout & N/A & 0.0 & 0.1 \\
\end{tabular}
\caption[Comparison of model architectures used in the experiments.]{Comparison of model architectures used in the experiments.}
\end{table}
\label{tab:mp_hparams}
\subsection{Readout}
The final component of our model architectures is the readout function, which transforms the node embeddings obtained after message passing (or directly from feature embeddings in the Non-Message-Passing baseline) into scores (the computed action's optimality probability) for each available operation. This readout function follows a consistent process for all model architectures.

The readout process consists of several key steps. First, we identify all valid operation-machine pairs that are available for scheduling at the current state. Next, we concatenate the embeddings of the corresponding operation and machine nodes for each valid pair, creating a feature vector that represents the entire scheduling possibility. Finally, these concatenated features are passed through an \ac{MLP} to compute the scalar score for each available operation.

\begin{table}[H]
\centering
\begin{tabular}{c|c|c}
\textbf{Layer} & \textbf{Input Dimension} & \textbf{Output Dimension}\\
\hline
Linear Layer 1 & $d_{\text{concat}}$ & $d_{\text{hidden}}$ \\

\ac{ReLU} & $d_{\text{hidden}}$ & $d_{\text{hidden}}$ \\

Linear Layer 2 & $d_{\text{hidden}}$ & $\lfloor d_{\text{hidden}}/2 \rfloor$ \\

\ac{ReLU} & $\lfloor d_{\text{hidden}}/2 \rfloor$ & $\lfloor d_{\text{hidden}}/2 \rfloor$ \\

Linear Layer 3 & $\lfloor d_{\text{hidden}}/2 \rfloor$ & 1 \\
\end{tabular}
\caption{Structure of the \ac{MLP} used in the readout function, where $d_{\text{concat}}$ is the dimensionality of the concatenated node features (equal to the sum of embedding dimensions across all node types), and $d_{\text{hidden}}$ is the hidden dimension hyperparameter mentioned in Table \ref{tab:mp_hparams}.}
\end{table}
During training, the scalar scores output by this MLP are compared against the binary labels (optimal or non-optimal) to compute the loss function. During inference, the operation with the highest score is selected for dispatching.

\section{Training}
In this section, we describe the procedure and hyperparameters followed to train each of the selected models mentioned in the previous section.\footnote{More information about all the experiments run can be found in the \href{https://github.com/Pabloo22/gnn_scheduler}{\texttt{gnn\_scheduler}} GitHub repository. The selected experiments, based on the average makespan obtained across all benchmark instances, are referred to as experiments 5, 24, and 40 (Non-\ac{MP}, \ac{RGIN}, and \ac{RGATv2}, respectively).} An important consideration is that not all models were trained using the same methodology. The reason is that new features were introduced to the training pipeline incrementally, but some of the best results were obtained in earlier iterations that used different settings.

\noindent Nevertheless, there are some commonalities across all the experiments:
\begin{itemize}[itemsep=0pt, topsep=0pt]
    \item A batch size of 512.
    \item AdamW optimizer with a learning rate of 0.0001 and a weight decay of 0.01.
    \item Used early stopping. The Non-\ac{MP} and \ac{RGIN} experiments were stopped manually. \ac{RGATv2}, on the other hand, used an automatic callback. This callback stops the training if there is no improvement in the validation dataset after a predetermined number of epochs.
    \item They were evaluated allowing the GNN-based dispatcher to reserve operations for later (i.e., scheduling operations that cannot start immediately); only dominated operations were pruned as possible actions.
\end{itemize}

\subsection{Non-Message-Passing Baseline}
The best results for the Non-\ac{MP} baseline were obtained in the 5th experiment run. The “sampling each $n$ steps" methodology, mentioned in Subsection \ref{subsec:sample-n-steps}, had not yet been introduced. Therefore, we used all the samples generated (i.e., $n=1$). Not sampling meant that the generated dataset was too large to fit into memory. Fitting a dataset into memory makes loading data to the GPU significantly faster.

We used a hybrid approach to take advantage of this; we divided the training dataset into smaller datasets that we could load into memory individually. They have a size of 25,000-50,000 instances each (\~2 million observation graphs), depending on the instance's size. After processing the first subdataset, we unload it from memory and load the next one. Each dataset processed is considered a “step."

\begin{figure}[htbp]
    \centering
    \includegraphics[width=0.7\linewidth]{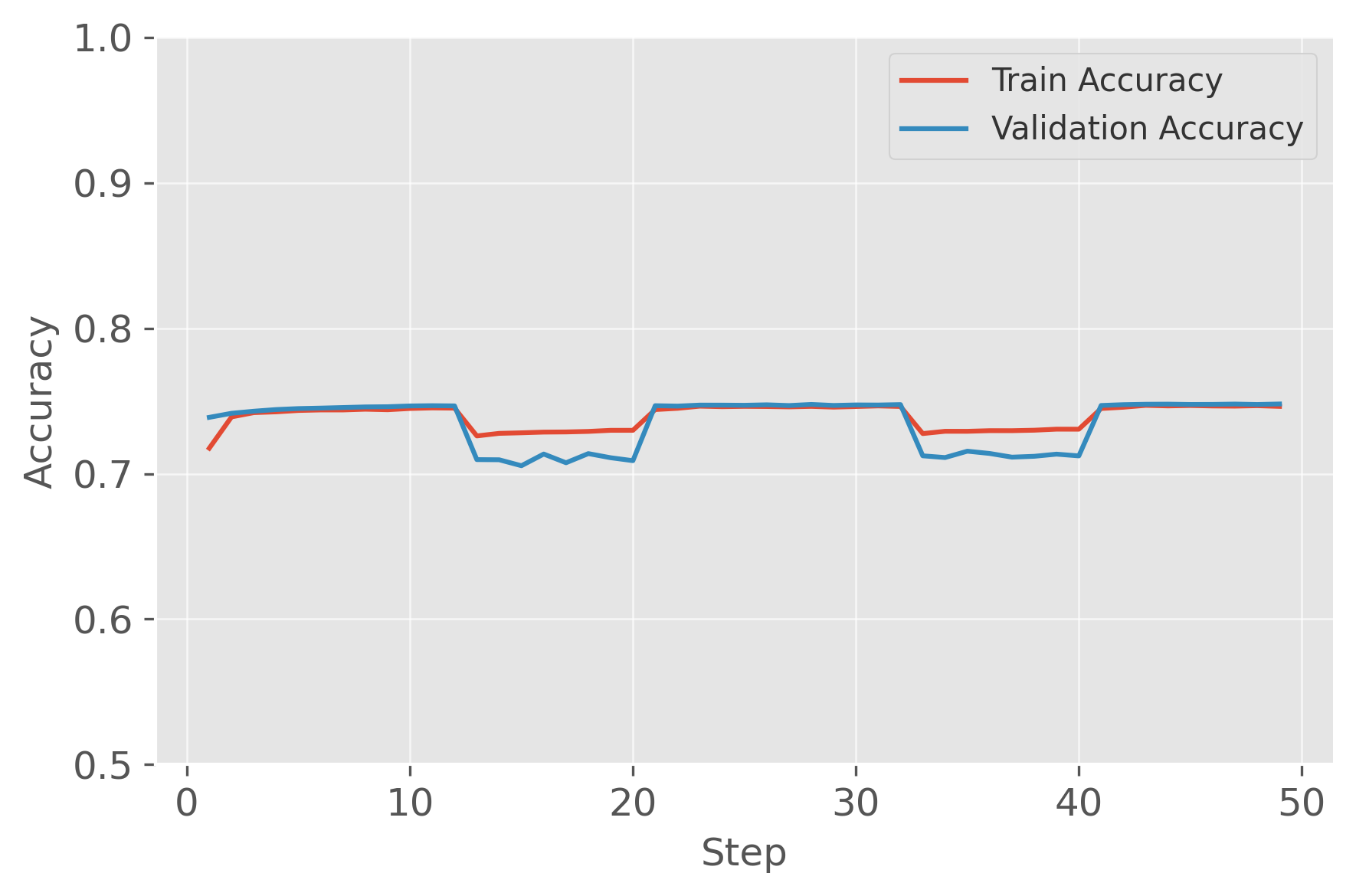}
    \caption[Accuracy in train and validation sets for the best Non-\ac{MP} model.]{Accuracy in train and validation sets for the best Non-\ac{MP} model.}
    \label{fig:exp5_acc}
\end{figure}

Figure \ref{fig:exp5_acc} shows training and validation accuracy (i.e., the percentage of correctly predicted operations as optimal or not) while training in these subdatasets. The validation dataset consisted of 1,000 instances of size $10\times10$. The sudden changes in performance are due to slight distribution shifts. In particular, the first subdatasets contain instances of $10\times10$ size, while subsequent ones contain ones with a number of machines ranging from 5 to 10. We did not use $8\times8$ instances for this experiment. 

\subsection{Relational Graph Isomorphism Network}
The best \ac{RGIN} model was trained using a sampling value of 11 ($n=11$). Since 11 is a prime number, we can ensure that the sampled observations will be evenly distributed across all step numbers, as argued in Subsection \ref{subsec:sample-n-steps}. As we mentioned, using this sample value allowed us to use a single dataset, which speeds up training, removes redundancy, and helps prevent overfitting.

We used the same validation dataset as in the previous experiment. However, fitting the entire dataset into memory with such a low sampling value forced us to use only $10\times10$ instances. Note that in this section, we are reporting the best experiments across many runs. We also tried to beat these results with a more refined methodology described in the next subsection that yielded the best results (average makespan across benchmarks) for the \ac{RGATv2} model.
\begin{figure}[htbp]
    \centering
    \includegraphics[width=0.7\linewidth]{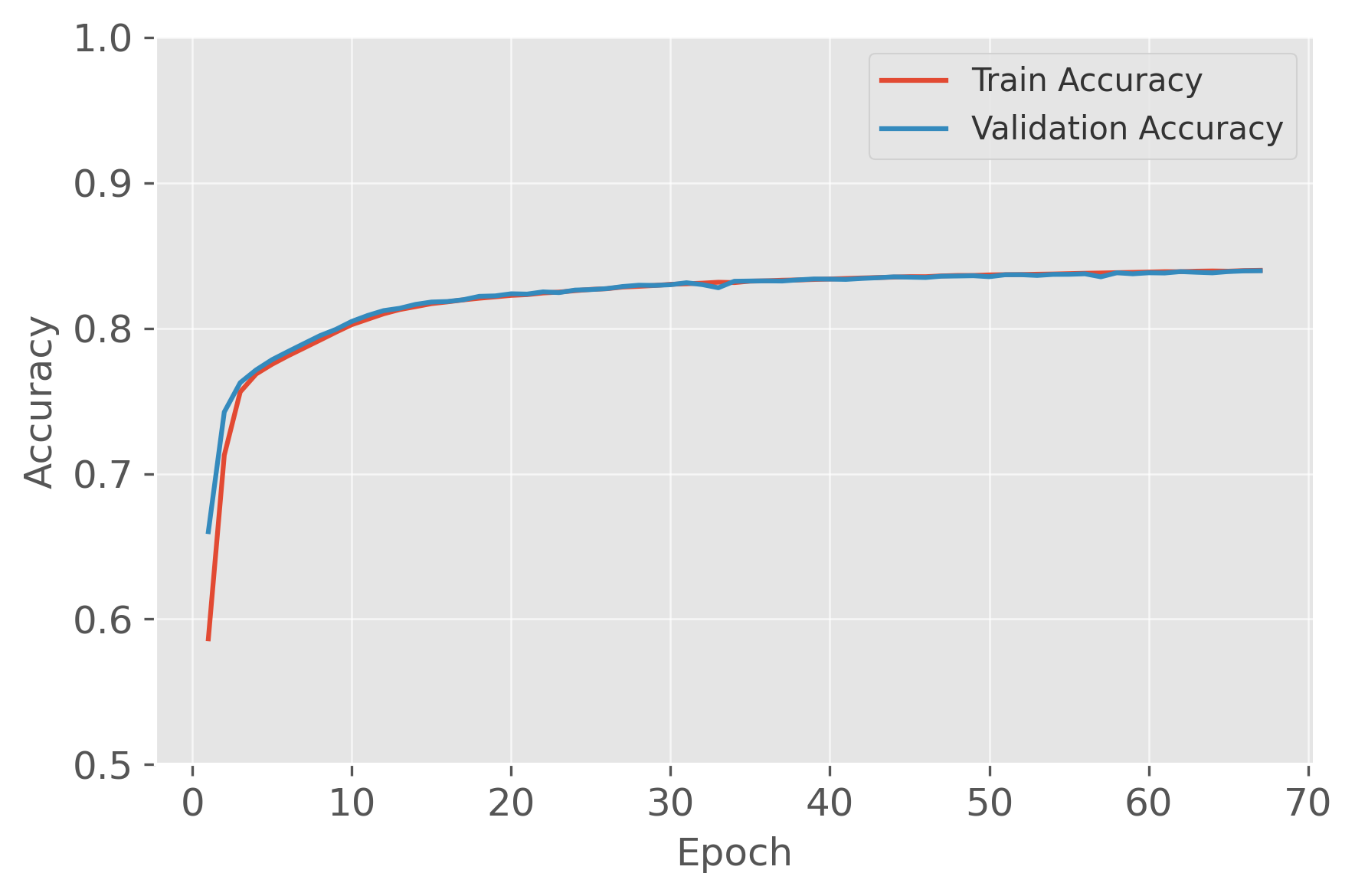}
    \caption{Accuracy in train and validation sets for the best RGIN model.}
    \label{fig:exp24_acc}
\end{figure}

\subsection{Relational Graph Attention Network v2}
For the best \ac{RGATv2}, we reduced the sampling value $n$ to 31 to use the entire dataset generated. Another difference was the evaluation dataset. Instead of accuracy, we solved the Taillard instances of size $10\times10$ and used the average optimality gap as the early stopping criterion, as done in \cite{lee2024il_jssp}. The optimality gap is the relative increase in makespan of the found solution with respect to the optimal makespan $C_{max}^*$:
$$\text{optimality gap} = \frac{C_{max} - C_{max}^*}{C_{max}^*}.$$

In particular, using \textit{JobShopLib}'s ready operation filters, we evaluated the model using the two competing available action definitions: Disabling reserving operations to only generate non-idle schedules, or only filtering out dominated operations. Then, we chose the lowest of these two average optimality gaps obtained as the early stopping's patience threshold; for an epoch to be considered an improvement, at least one of the two gaps computed for that epoch had to beat the lowest average optimality gap obtained so far.

We also tracked the model's accuracy in a validation dataset that consisted of 100 instances with the number of jobs and machines sampled randomly from $U(10, 15)$ and $U(5, 10)$, respectively. Because this data is more diverse than the previous $10\times10$ instances, it should give us a better estimate of how it will perform in unseen instances.

Interestingly, while validation accuracy and loss improved after each epoch, the same cannot be said of the validation optimality gap in the $15\times15$-sized Taillard instances. This suggests that the patterns the model was learning at this stage of learning did not generalize to bigger instances. In fact, when evaluating the model without operation reservations, the best optimality gap was obtained in the first epoch. However, the best result came at epoch 10 by allowing the model to reserve operations.

\begin{figure}[htbp]
    \makebox[\textwidth][c]{%
    \begin{minipage}{1\textwidth}
        \begin{subfigure}[b]{0.5\textwidth}
            \centering
            \includegraphics[width=\textwidth]{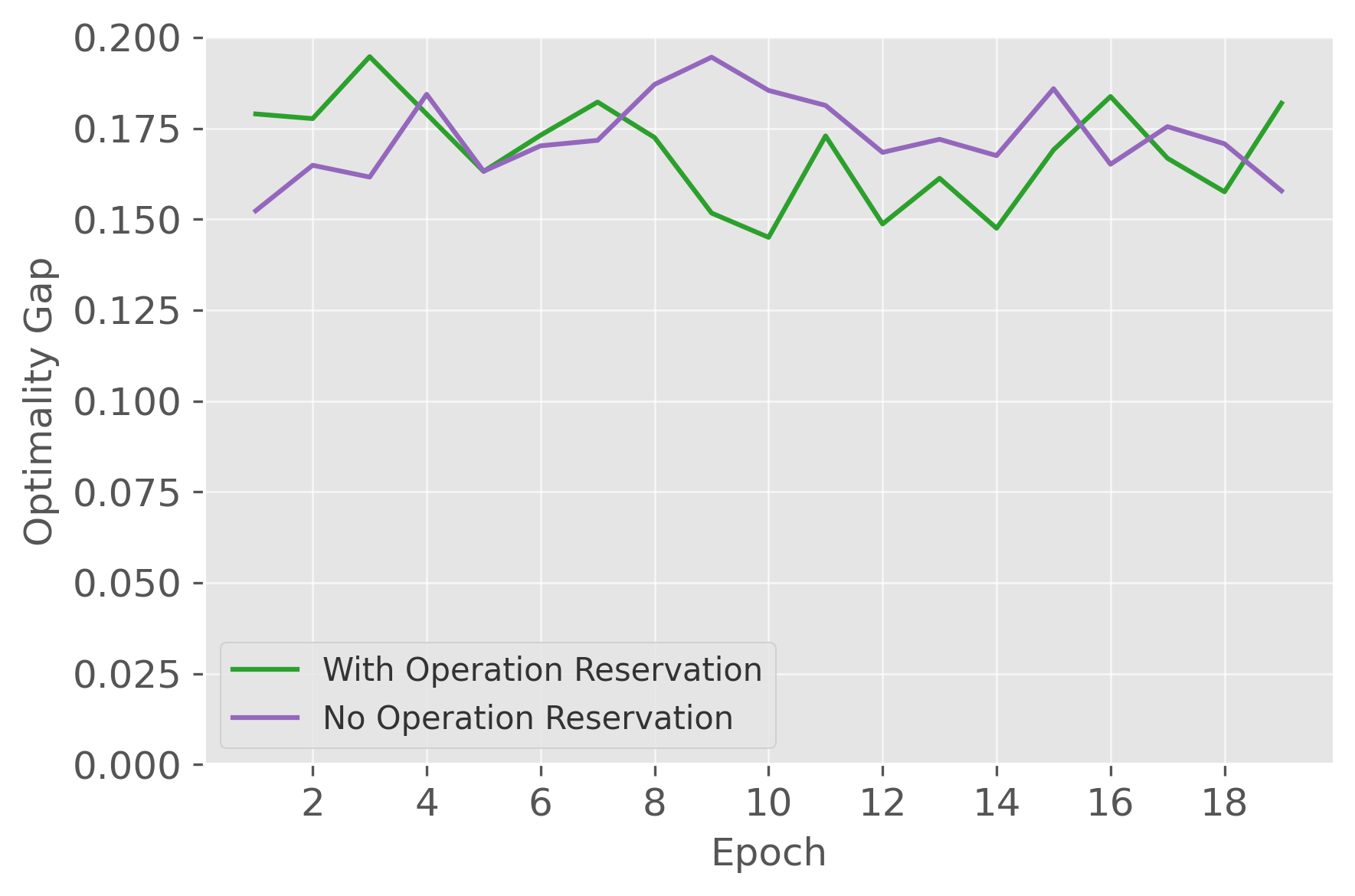}
            \caption{}
            \label{fig:experiment_40_gap}
        \end{subfigure}
        \hfill
        \begin{subfigure}[b]{0.5\textwidth}
            \centering
            \includegraphics[width=\textwidth]{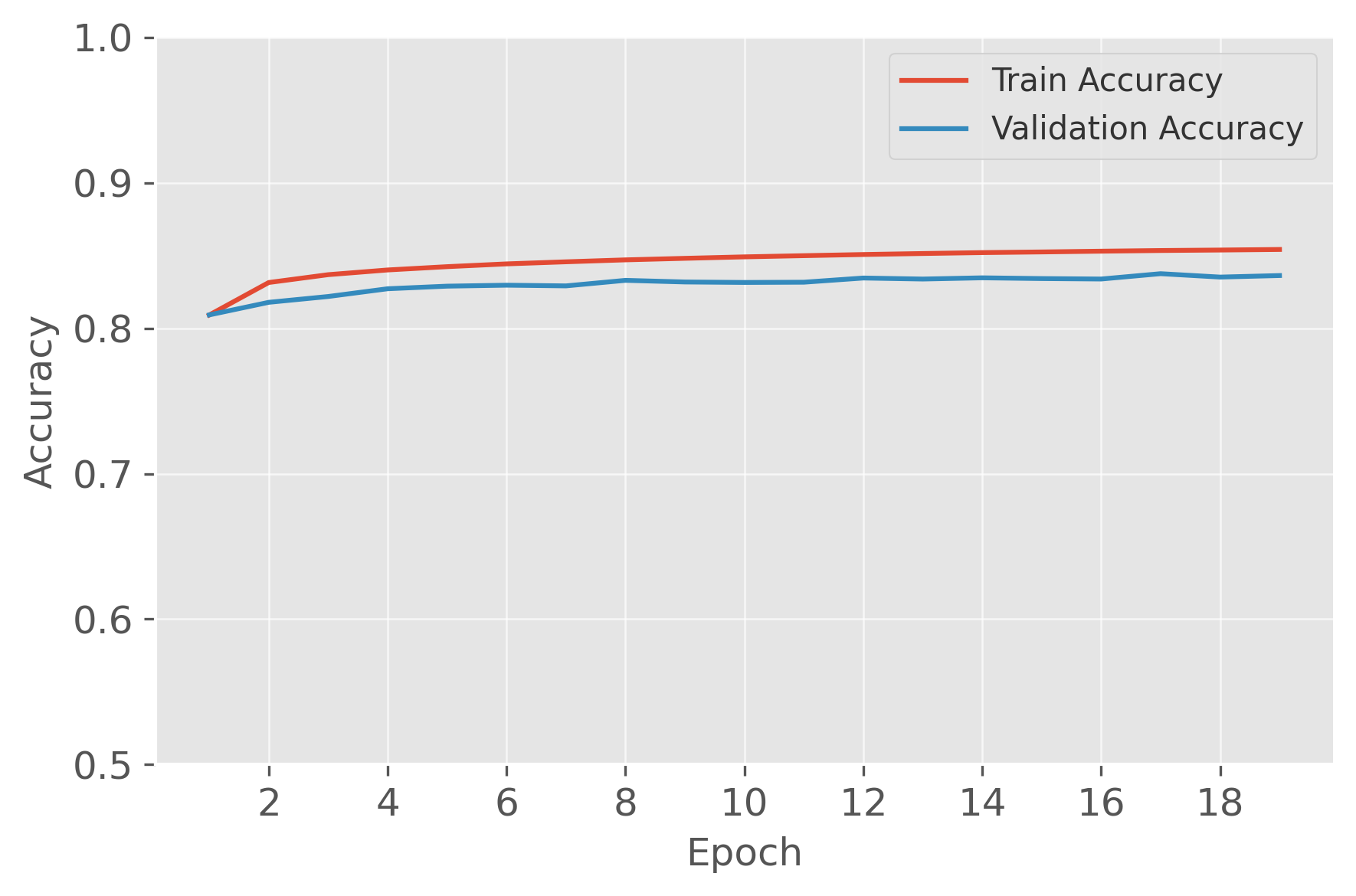}
            \caption{}
            \label{fig:experiment_40_acc}
        \end{subfigure}
    \end{minipage}
    }
    \caption{Average optimality gaps (a) and accuracy (b) obtained in their respective validation data for the best RGATv2 model.}
    \label{fig:experiment_40_val}
\end{figure}

\section{Results}
\begin{figure}[htbp]
    \centering
    \includegraphics[width=1\linewidth]{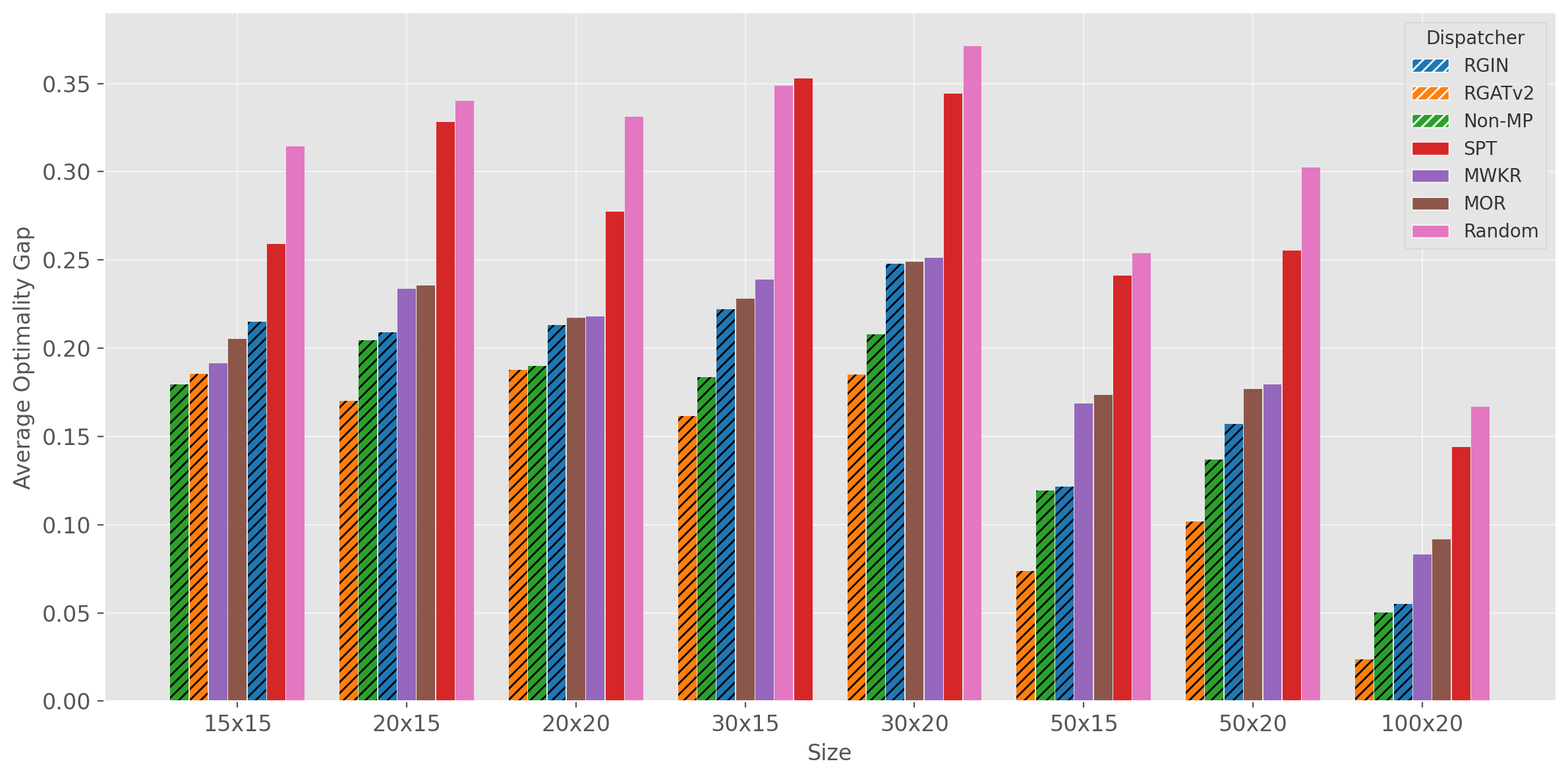}
    \caption[Average optimality gap in Taillard benchmark \citep{taillard1993benchmarks} grouped by size comparison between our GNN-based dispatchers and simple PDRs.]{Average optimality gap in Taillard benchmark \citep{taillard1993benchmarks} grouped by size comparison between our GNN-based dispatchers (painted with a hatch pattern) and simple \ac{PDRs}. The rule \ac{FCFS} is omitted because, in instances with the same number of operations for all jobs, it is equivalent to \ac{MOR}.}
    \label{fig:experiments_ta_with_pdrs}
\end{figure}

\begin{figure}[htbp]
    \centering
    \includegraphics[width=1\linewidth]{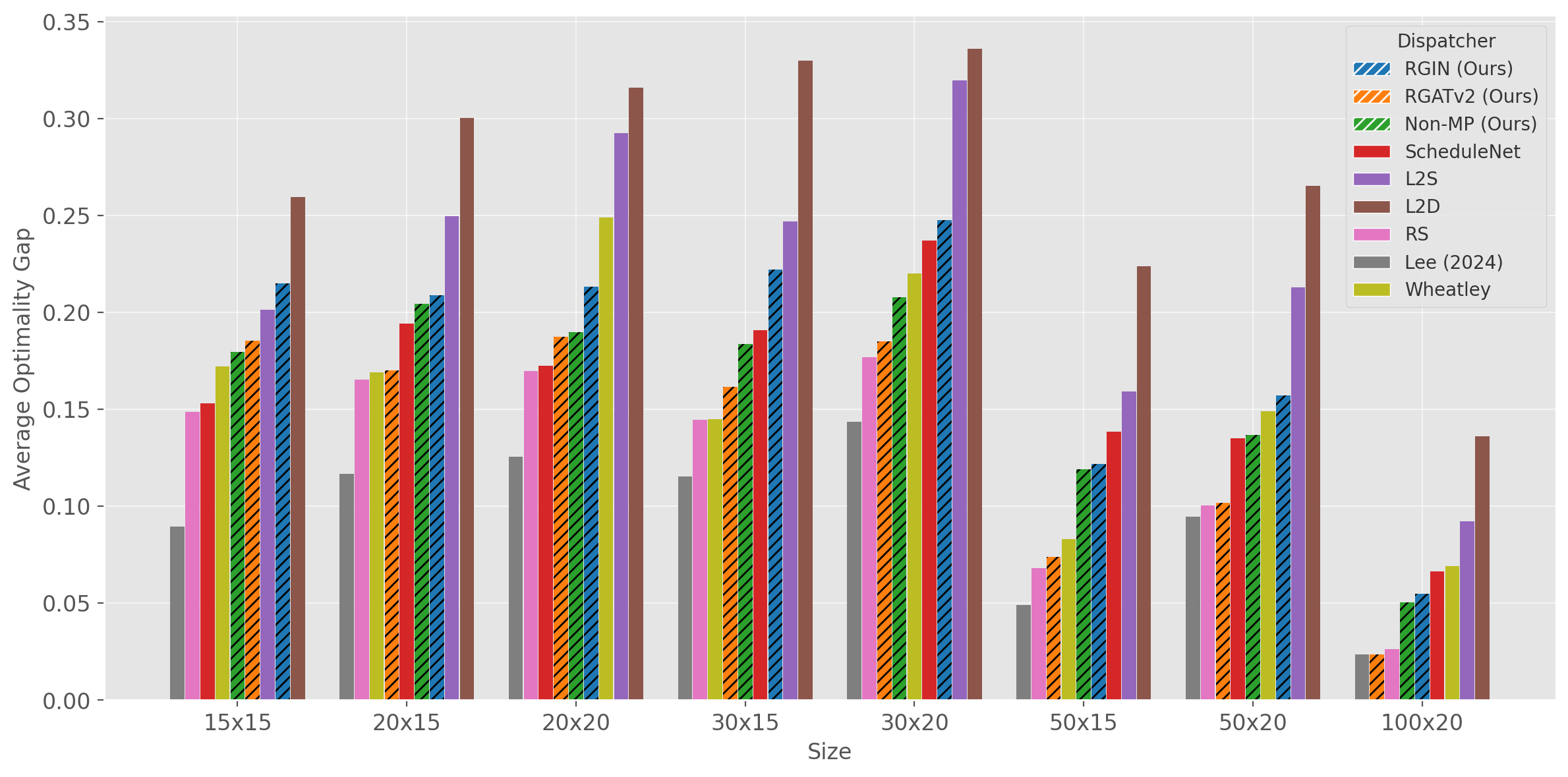}
    \caption[Average optimality gap in Taillard benchmark \citep{taillard1993benchmarks} grouped by size comparison between our GNN-based dispatchers and simple PDRs.]{Average optimality gap obtained in Taillard benchmark \citep{taillard1993benchmarks} by GNN-based dispatchers grouped by size. The figure shows the reported results by \cite{lee2024il_jssp} (“Lee"), \cite{ho2023residual} (“RS," which stands for “residual scheduling"), \cite{park2021schedule_net} (“ScheduleNet"), \cite{Park2021l2s} (“L2S," meaning “learning to schedule"), and \cite{zhang2020l2d} (“L2D," meaning “learning to dispatch"), and Wheatley \citep{infantes2024wheatley}. Our trained models (\ac{RGIN}, \ac{RGATv2}, Non-\ac{MP}) are painted using a hatch pattern.}
    \label{fig:experiments_ta_with_sota}. 
\end{figure}

In this section, we report the results of our trained GNN-based dispatchers. In particular, we compare their results with the simple \ac{PDRs} mentioned in Subsection \ref{ch2:pdr} and with the previous works summarized in Table \ref{tab:sota_comparison}. These results are shown in Figures \ref{fig:experiments_ta_with_pdrs} and \ref{fig:experiments_ta_with_sota}.

We use Taillard's benchmark to plot these results to be able to compare with all the previous works (\cite{Park2021l2s} and \cite{ho2023residual} only reported results in this benchmark). This is also the largest benchmark, comprising a total of 80 instances, and it contains the largest set of instances, up to $100\times20$.

Interestingly, the Non-\ac{MP} baseline performed slightly better than the \ac{RGIN} model. This could be explained because these models can easily distinguish graphs of different sizes for the reasons mentioned in Subsection \ref{subsec:mp}. This expressivity can potentially “distract" the model when seeing graphs out of distribution. In other words, it may not generalize well to larger instances not seen during training.

The best model across all experiments was \ac{RGATv2}. This model updates a node's neighbors by conducting a weighted average of their neighbors. Much of the learning occurs through learning these weights (attention coefficients). This attention mechanism can learn simple patterns that also appear in large instances. For example, when updating available operations, the GNN could give large weights to its own machine nodes or to nodes that connect to machine nodes that connect to other available operations. This phenomenon would explain why this model performs so well in the biggest instances---state-of-the-art performance in many of Taillard's $100\times20$-sized instances. It also justifies the fact that optimality gaps in larger instances were not improving, despite the improvement in validation accuracy. The instances for computing optimality gaps (Taillard's $15\times15$) were larger than those used for accuracy.

\clearpage

\subsection{\textit{JobShopLib}'s Adoption}
\label{subsec:job_shop_lib_adoption}
Since \textit{JobShopLib}'s publication on GitHub in February 2024, it has received more than 30 stars\footnote{In GitHub, a star is a simple way for users to indicate that they like or appreciate a repository and want to keep track of it. It essentially functions as a “like" or bookmark on the platform.}. It has also received a total of six issues with feature requests or inquiries about the library coming from external users. Additionally, we received three pull requests from two users. The first one, made by \href{https://github.com/CarlosMarchMoya}{Carlos March Moya}, corrected an example notebook of the documentation by updating the code to be compatible with a newer \textit{JobShopLib} version. The other two pull requests were made by \href{https://github.com/AlbinLind}{Albin Lind}. He improved the documentation by adding summary tables to the API modules that still lacked one, sorting their rows, and fixing some typos. He also updated type hints to follow the newer standard introduced in Python 3.9 and 3.10 \citep{pep585, pep604}.
This data shows that some users have already benefitted from \textit{JobShopLib}.

\begin{figure}[H]
    \centering
    \includegraphics[width=0.75\linewidth]{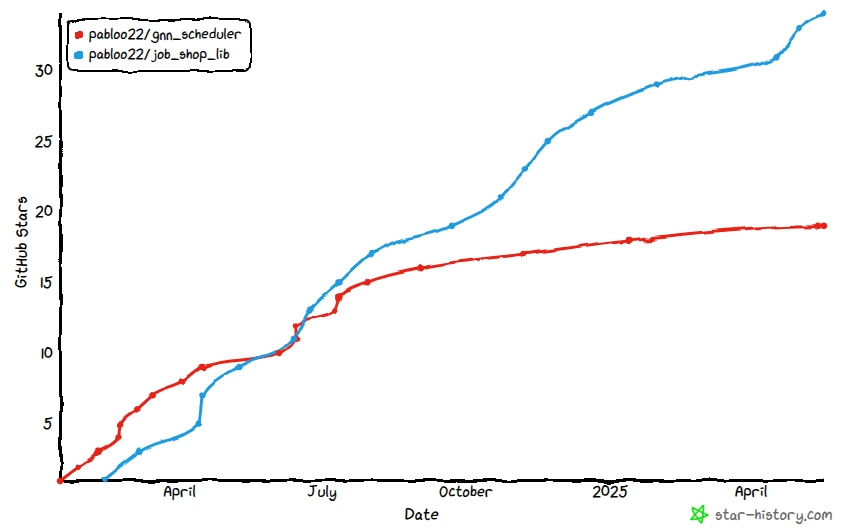}
    \caption{Number of GitHub stars obtained in \href{https://github.com/Pabloo22/job_shop_lib}{\textit{JobShopLib}} and \href{https://github.com/Pabloo22/gnn_scheduler}{\texttt{gnn\_scheduler}} repositories since their creation.}
    \label{fig:gh-stars}
\end{figure}

%% file: chapters/09-Ethics.tex
\doublespacing 

\chapter{Ethical, Environmental, and Social Aspects}
\label{ch9}

\begin{spacing}{1} 
\minitoc 
\end{spacing} 
\thesisspacing 
This chapter reflects on the ethical, environmental, and social aspects affecting our project:

\textbf{Social Considerations}. Our current work focuses on optimizing for makespan, chosen for its simplicity and established role as a canonical objective in scheduling. Achieving a shorter makespan offers clear business benefits, such as increased productivity, which can, in turn, have positive social impacts. For example, clients would benefit from shorter waiting times.

\textbf{Environmental Considerations}. It is important to acknowledge the energy consumption required to run the experiments conducted in this project. Training models and generating data can keep the computer or server used busy for hours. We estimate that the total runtime of all the experiments can be around one week. This CPU and GPU utilization can contribute to CO2 emissions if the energy does not come from green sources \citep{Yu2024Revisit}.

\noindent A way to mitigate this detrimental effect and add additional social benefits is to incorporate different optimization objectives other than the makespan. Especially in flexible problems, there is an opportunity to optimize not only for makespan but also for minimizing energy consumption or CO2 emissions. For example, in problems where operations can be assigned to more than one machine, energy-efficient machines could be prioritized at the expense of a minor sacrifice in makespan. This multi-objective is also aligned with the social aspects previously discussed because reducing energy consumption also reduces costs, which could translate into cheaper products for the consumer. Especially in large \ac{JSSP} instances, the potential for energy reductions given by more efficient schedules can compensate for the energy employed to train the GNN-dispatcher.

\textbf{Ethical Considerations}. As scheduling optimization becomes more efficient, there may be implications for human operators whose roles traditionally included making such decisions. Organizations implementing these systems should consider providing appropriate training or transition opportunities for affected workers \citep{WEF2023FutureJobs}.

%% file: chapters/AppendixA.tex
\doublespacing 

\chapter{Tables with Results for all Benchmarks}

\thesisspacing

\begin{table}
    \centering
    \small
    \begin{tabular}{ccccccccc}
    Size & \ac{RGIN} & \ac{RGATv2} & Non-\ac{MP} & \ac{SPT} & \ac{MWKR} & \ac{MOR} & Random \\
\hline
$15\times15$ & 0.2149 & 0.1854 & \textbf{0.1796} & 0.2589 & 0.1915 & 0.2053 & 0.3144 \\
$20\times15$ & 0.2088 & \textbf{0.1702} & 0.2044 & 0.3283 & 0.2336 & 0.2356 & 0.3401 \\
$20\times20$ & 0.2132 & \textbf{0.1876} & 0.1898 & 0.2775 & 0.2181 & 0.2171 & 0.3313 \\
$30\times15$ & 0.2222 & \textbf{0.1615} & 0.1837 & 0.3527 & 0.2391 & 0.2282 & 0.3486 \\
$30\times20$& 0.2477 & \textbf{0.1850} & 0.2077 & 0.3441 & 0.2514 & 0.2491 & 0.3713 \\
$50\times15$ & 0.1217 & \textbf{0.0739} & 0.1192 & 0.2411 & 0.1686 & 0.1737 & 0.2538 \\
$50\times20 $& 0.1571 & \textbf{0.1019} & 0.1369 & 0.2554 & 0.1795 & 0.1768 & 0.3024 \\
$100\times20 $& 0.0550 & \textbf{0.0237} & 0.0504 & 0.1441 & 0.0831 & 0.0915 & 0.1668 \\
    \end{tabular}
    \caption{Average optimality gap in Taillard benchmark \citep{taillard1993benchmarks} grouped by size comparison between our GNN-based dispatchers and simple PDRs.}
    \label{tab:ta_with_pdrs}
\end{table}

The following tables show the makespan for all the benchmark instances described in Subsection \ref{subsec:benchmark_instances}. Non-MP stands for “Non-Message-Passing." Our experiments: \ac{RGIN}, \ac{RGATv2}, and Non-MP (stands for “Non-Message-Passing") are explained in Chapter \ref{ch8}. \ac{SPT}, \ac{FCFS}, \ac{MWKR}, and Random are all \ac{PDRs} discussed in Subsection \ref{ch2:pdr}. These \ac{PDRs} use the “no reservation" definition of available operations (i.e., only non-idle schedules can be generated). We show the average result of 5 runs for the Random dispatching rule (rounded). The “Best known" column contains the optimum results of these instances or the upper bounds computed (obtained from \cite{Weise2025JSSP_Instances_Results}). The best makespan (excluding the “Best known" column) is bolded.
\newpage
\singlespacing
\begin{longtable}{cccccccccc}
\caption[Makespan obtained for each benchmark instance (excluding Taillard).]{Makespan obtained for each benchmark instance (excluding Taillard).}
\label{tab:all_our_makespans} \\
 \makecell{Instance\\name} & RGIN & RGATv2 & Non-MP & SPT & FCFS & MWKR & MOR & Random & \makecell{Best\\known} \\
\hline
\hline
\endfirsthead
\multicolumn{10}{c}%
{{\tablename\ \thetable{} -- continued from previous page}} \\
\makecell{Instance\\name} & RGIN & RGATv2 & Non-MP & SPT & FCFS & MWKR & MOR & Random & \makecell{Best\\known} \\
\hline
\hline
\endhead

\hline \multicolumn{10}{r}{{Continued on next page}} \\
\endfoot

\hline
\endlastfoot
abz5 & 1450 & 1378 & \textbf{1299} & 1352 & 1336 & 1369 & 1336 & 1498 & 1234 \\
abz6 & \textbf{982} & 1078 & 1003 & 1097 & 1031 & 987 & 1031 & 1089 & 943 \\
abz7 & 767 & \textbf{754} & 820 & 849 & 775 & 769 & 775 & 830 & 656 \\
abz8 & 824 & \textbf{765} & 831 & 929 & 810 & 825 & 810 & 902 & 665 \\
abz9 & 811 & \textbf{779} & 818 & 887 & 899 & 857 & 899 & 890 & 678 \\
\hline
ft06 &  \textbf{59} & \textbf{59} & 61 & 88 & \textbf{59} & 61 & \textbf{59} & 64 & 55 \\
ft10 & 1093 & \textbf{1058} & 1078 & 1074 & 1163 & 1108 & 1163 & 1257 & 930 \\
ft20 & 1308 & \textbf{1247} & 1290 & 1267 & 1601 & 1501 & 1601 & 1476 & 1165 \\
\hline
la01 & 708 & 685 &  \textbf{678} & 751 & 763 & 735 & 763 & 793 & 666 \\
la02 & 784 & \textbf{696} & 784 & 821 & 812 & 817 & 812 & 854 & 655 \\
la03 & 669 & \textbf{656} & 734 & 672 & 726 & 696 & 726 & 791 & 597 \\
la04 & \textbf{637} & 668 & 655 & 711 & 706 & 758 & 706 & 751 & 590 \\
la05 & \textbf{593} & \textbf{593} & \textbf{593} & 610 & \textbf{593} & \textbf{593} & \textbf{593} & 632 & 593 \\
la06 & \textbf{926} & \textbf{926} & 974 & 1200 & \textbf{926} & \textbf{926} & \textbf{926} & 1093 & 926 \\
la07 & \textbf{928} & 935 & 935 & 1034 & 1001 & 970 & 1001 & 1037 & 890 \\
la08 & \textbf{867} & 976 & 888 & 942 & 925 & 957 & 925 & 1059 & 863 \\
la09 & \textbf{951} & \textbf{951} & \textbf{951} & 1045 & \textbf{951} & 1015 & \textbf{951} & 1104 & 951 \\
la10 & \textbf{958} & \textbf{958} & \textbf{958} & 1049 & \textbf{958} & 966 & \textbf{958} & 1019 & 958 \\
la11 & \textbf{1222} & \textbf{1222} & \textbf{1222} & 1473 & \textbf{1222} & 1268 & \textbf{1222} & 1318 & 1222 \\
la12 & \textbf{1039} & 1081 & 1057 & 1203 & \textbf{1039} & 1137 & \textbf{1039} & 1176 & 1039 \\
la13 & \textbf{1150} & 1160 & \textbf{1150} & 1275 & \textbf{1150} & 1166 & \textbf{1150} & 1260 & 1150 \\
la14 & \textbf{1292} & \textbf{1292} & \textbf{1292} & 1427 & \textbf{1292} & \textbf{1292} & \textbf{1292} & 1319 & 1292 \\
la15 & 1309 & \textbf{1207} & 1246 & 1339 & 1436 & 1343 & 1436 & 1513 & 1207 \\
la16 & 1268 & \textbf{1010} & 1125 & 1156 & 1108 & 1054 & 1108 & 1194 & 945 \\
la17 & \textbf{836} & 925 & 874 & 924 & 844 & 846 & 844 & 975 & 784 \\
la18 & 1029 & 1008 & \textbf{907} & 981 & 942 & 970 & 942 & 1083 & 848 \\
la19 & 969 & 969 & 969 & \textbf{940} & 1088 & 1013 & 1088 & 988 & 842 \\
la20 & 1000 & 972 & \textbf{959} & 1000 & 1130 & 964 & 1130 & 1148 & 902 \\
la21 & \textbf{1188} & 1201 & 1205 & 1324 & 1251 & 1264 & 1251 & 1374 & 1046 \\
la22 & 1038 & \textbf{981} & \textbf{981} & 1180 & 1198 & 1079 & 1198 & 1278 & 927 \\
la23 & 1096 & 1154 & \textbf{1085} & 1162 & 1268 & 1185 & 1268 & 1319 & 1032 \\
la24 & \textbf{1054} & 1086 & 1095 & 1203 & 1149 & 1101 & 1149 & 1226 & 935 \\
la25 & 1155 & \textbf{1126} & 1176 & 1449 & 1209 & 1166 & 1209 & 1392 & 977 \\
la26 & 1360 & \textbf{1284} & 1425 & 1498 & 1411 & 1435 & 1411 & 1554 & 1218 \\
la27 & 1358 & \textbf{1327} & 1442 & 1784 & 1566 & 1442 & 1566 & 1664 & 1235 \\
la28 & 1511 & \textbf{1381} & 1462 & 1610 & 1477 & 1487 & 1477 & 1554 & 1216 \\
la29 & \textbf{1276} & 1354 & 1322 & 1556 & 1437 & 1337 & 1437 & 1636 & 1152 \\
la30 & 1451 & 1409 & \textbf{1384} & 1792 & 1565 & 1534 & 1565 & 1685 & 1355 \\
la31 & \textbf{1792} & 1920 & 1806 & 1951 & 1836 & 1931 & 1836 & 1991 & 1784 \\
la32 & \textbf{1850} & \textbf{1850} & 1877 & 2165 & 1984 & 1875 & 1984 & 2230 & 1850 \\
la33 & 1805 & \textbf{1719} & 1805 & 1901 & 1811 & 1875 & 1811 & 2033 & 1719 \\
la34 & \textbf{1721} & \textbf{1721} & 1771 & 2070 & 1853 & 1935 & 1853 & 2025 & 1721 \\
la35 & 1905 & 1905 & \textbf{1888} & 2118 & 2064 & 2118 & 2064 & 2279 & 1888 \\
la36 & 1510 & \textbf{1365} & 1421 & 1799 & 1492 & 1510 & 1492 & 1619 & 1268 \\
la37 & 1594 & 1571 & \textbf{1567} & 1655 & 1606 & 1588 & 1606 & 1736 & 1397 \\
la38 & 1462 & \textbf{1390} & 1403 & 1404 & 1455 & 1421 & 1455 & 1712 & 1196 \\
la39 & \textbf{1442} & 1492 & 1467 & 1534 & 1540 & 1500 & 1540 & 1590 & 1233 \\
la40 & 1358 & \textbf{1344} & 1385 & 1476 & 1358 & 1440 & 1358 & 1608 & 1222 \\
\hline
orb01 & 1226 &  \textbf{1201} & 1249 & 1478 & 1307 & 1359 & 1307 & 1326 & 1059 \\
orb02 & 987 & 987 & \textbf{921} & 1175 & 1047 & 1047 & 1047 & 1152 & 888 \\
orb03 & 1191 & 1206 & \textbf{1119} & 1179 & 1445 & 1247 & 1445 & 1418 & 1005 \\
orb04 & \textbf{1098} & 1139 & 1139 & 1236 & 1287 & 1172 & 1287 & 1307 & 1005 \\
orb05 & \textbf{1002} & 1068 & 1059 & 1152 & 1050 & 1173 & 1050 & 1103 & 887 \\
orb06 & 1198 & \textbf{1156} & 1228 & 1190 & 1345 & 1291 & 1345 & 1336 & 1010 \\
orb07 & 481 & \textbf{431} & 447 & 504 & 500 & 483 & 500 & 515 & 397 \\
orb08 & 1068 & \textbf{1032} & 1110 & 1107 & 1278 & 1180 & 1278 & 1212 & 899 \\
orb09 & 1104 & \textbf{1102} & 1107 & 1262 & 1165 & 1144 & 1165 & 1238 & 934 \\
orb10 & 1058 & \textbf{1051} & 1088 & 1113 & 1256 & 1220 & 1256 & 1220 & 944 \\
\hline
swv01 &  \textbf{1582} & 1632 & 1757 & 1737 & 1971 & 1988 & 1971 & 1975 & 1407 \\
swv02 & 1779 & \textbf{1660} & 1756 & 1706 & 2158 & 1971 & 2158 & 2055 & 1475 \\
swv03 & 1610 & \textbf{1574} & 1747 & 1806 & 1870 & 1874 & 1870 & 1948 & 1398 \\
swv04 & \textbf{1728} & 1816 & 1917 & 1874 & 2026 & 1879 & 2026 & 1923 & 1464 \\
swv05 & 1692 & \textbf{1616} & 1764 & 1922 & 2049 & 1882 & 2049 & 2034 & 1424 \\
swv06 & \textbf{1950} & 2144 & 2216 & 2140 & 2287 & 2135 & 2287 & 2252 & 1671 \\
swv07 & 2014 & \textbf{1860} & 1904 & 2146 & 2101 & 2115 & 2101 & 2183 & 1594 \\
swv08 & 2149 & \textbf{2110} & 2272 & 2231 & 2480 & 2544 & 2480 & 2410 & 1752 \\
swv09 & 1980 & \textbf{1969} & 2116 & 2247 & 2553 & 2118 & 2553 & 2358 & 1655 \\
swv10 & \textbf{1978} & 2050 & 2115 & 2337 & 2431 & 2300 & 2431 & 2349 & 1743 \\
swv11 & 3369 & \textbf{3266} & 3699 & 3714 & 4642 & 4257 & 4642 & 4209 & 2983 \\
swv12 & 3545 & \textbf{3257} & 3523 & 3759 & 4821 & 4327 & 4821 & 4281 & 2977 \\
swv13 & 3526 & \textbf{3353} & 3698 & 3657 & 4755 & 4389 & 4755 & 4355 & 3104 \\
swv14 & 3361 & \textbf{3175} & 3428 & 3506 & 4740 & 4164 & 4740 & 4181 & 2968 \\
swv15 & 3421 & \textbf{3231} & 3647 & 3501 & 4905 & 4374 & 4905 & 4295 & 2885 \\
swv16 & \textbf{2924} & \textbf{2924} & \textbf{2924} & 3453 & \textbf{2924} & \textbf{2924} & \textbf{2924} & 3125 & 2924 \\
swv17 & \textbf{2794} & \textbf{2794} & \textbf{2794} & 3082 & 2848 & 2861 & 2848 & 3088 & 2794 \\
swv18 & \textbf{2852} & \textbf{2852} & 2963 & 3191 & \textbf{2852} & 2935 & \textbf{2852} & 3196 & 2852 \\
swv19 & 2947 & \textbf{2843} & \textbf{2843} & 3161 & 3060 & 3008 & 3060 & 3173 & 2843 \\
swv20 & \textbf{2823} & \textbf{2823} & \textbf{2823} & 3125 & 2851 & \textbf{2823} & 2851 & 2997 & 2823 \\
\hline
yn1 & 1057 &  \textbf{997} & 1059 & 1196 & 1045 & 1005 & 1045 & 1139 & 884 \\
yn2 & 1065 & \textbf{1044} & 1071 & 1256 & 1074 & 1081 & 1074 & 1199 & 904 \\
yn3 & 1057 & 1014 & \textbf{1000} & 1042 & 1100 & 1118 & 1100 & 1145 & 892 \\
yn4 & 1185 & \textbf{1161} & 1209 & 1273 & 1267 & 1164 & 1267 & 1226 & 968 \\
\end{longtable}

\newpage

\begin{longtable}{cccccccccc}
\caption[Makespan obtained for Taillard instances.]{Makespan obtained for Taillard instances.}
\label{tab:taillard_makespans} \\
 \makecell{Instance\\name} & RGIN & RGATv2 & Non-MP & SPT & FCFS & MWKR & MOR & Random & \makecell{Best\\known} \\
\hline
\hline
\endfirsthead
\multicolumn{10}{c}%
{{\tablename\ \thetable{} -- continued from previous page}} \\
\makecell{Instance\\name} & RGIN & RGATv2 & Non-MP & SPT & FCFS & MWKR & MOR & Random & \makecell{Best\\known} \\
\hline
\hline
\endhead

\hline \multicolumn{10}{r}{{Continued on next page}} \\
\endfoot

\hline
\endlastfoot
ta01 & 1418 & 1459 &  \textbf{1394} & 1462 & 1438 & 1491 & 1438 & 1552 & 1231 \\
ta02 & 1542 & \textbf{1396} & 1528 & 1446 & 1452 & 1440 & 1452 & 1671 & 1244 \\
ta03 & 1472 & 1444 & \textbf{1370} & 1495 & 1418 & 1426 & 1418 & 1651 & 1218 \\
ta04 & 1407 & 1458 & 1499 & 1708 & 1457 & \textbf{1387} & 1457 & 1572 & 1175 \\
ta05 & 1523 & 1509 & \textbf{1370} & 1618 & 1448 & 1494 & 1448 & 1640 & 1224 \\
ta06 & 1502 & 1426 & 1463 & 1522 & 1486 & \textbf{1369} & 1486 & 1567 & 1238 \\
ta07 & 1614 & 1444 & 1521 & \textbf{1434} & 1456 & 1470 & 1456 & 1554 & 1227 \\
ta08 & 1399 & \textbf{1367} & 1434 & 1457 & 1482 & 1491 & 1482 & 1507 & 1217 \\
ta09 & 1503 & 1513 & \textbf{1486} & 1622 & 1594 & 1541 & 1594 & 1743 & 1274 \\
ta10 & 1551 & 1549 & \textbf{1426} & 1697 & 1582 & 1534 & 1582 & 1695 & 1241 \\
ta11 & 1701 & 1614 & \textbf{1546} & 1865 & 1665 & 1685 & 1665 & 1822 & 1357 \\
ta12 & 1596 & \textbf{1592} & 1716 & 1667 & 1739 & 1707 & 1739 & 1799 & 1367 \\
ta13 & 1699 & 1649 & 1677 & 1802 & \textbf{1642} & 1690 & \textbf{1642} & 1814 & 1342 \\
ta14 & 1533 & \textbf{1522} & 1588 & 1635 & 1662 & 1563 & 1662 & 1766 & 1345 \\
ta15 & 1669 & \textbf{1583} & 1625 & 1835 & 1682 & 1696 & 1682 & 1841 & 1339 \\
ta16 & 1692 & 1589 & 1700 & 1965 & 1638 & \textbf{1584} & 1638 & 1891 & 1360 \\
ta17 & \textbf{1666} & 1670 & 1723 & 2059 & 1856 & 1804 & 1856 & 1953 & 1462 \\
ta18 & 1768 & 1630 & \textbf{1592} & 1808 & 1710 & 1751 & 1710 & 1785 & 1396 \\
ta19 & 1581 & \textbf{1511} & 1591 & 1789 & 1651 & 1667 & 1651 & 1827 & 1332 \\
ta20 & \textbf{1588} & 1608 & 1674 & 1710 & 1622 & 1689 & 1622 & 1786 & 1348 \\
ta21 & 2166 & 2000 & \textbf{1922} & 2175 & 1964 & 2044 & 1964 & 2099 & 1642 \\
ta22 & 2004 & \textbf{1848} & 1930 & 1965 & 1905 & 1914 & 1905 & 2176 & 1600 \\
ta23 & 2033 & 1829 & \textbf{1778} & 1933 & 1922 & 1983 & 1922 & 2070 & 1557 \\
ta24 & 1970 & 2009 & \textbf{1906} & 2230 & 1943 & 1982 & 1943 & 2210 & 1644 \\
ta25 & 1927 & \textbf{1857} & 1948 & 1950 & 1957 & 1941 & 1957 & 2286 & 1595 \\
ta26 & 1920 & \textbf{1888} & 1919 & 2188 & 1964 & 1951 & 1964 & 2288 & 1643 \\
ta27 & 1941 & \textbf{1897} & 2038 & 2096 & 2160 & 2091 & 2160 & 2158 & 1680 \\
ta28 & \textbf{1839} & 1944 & 1921 & 1968 & 1952 & 1997 & 1952 & 2054 & 1603 \\
ta29 & 2011 & 1995 & 2002 & 2166 & 1899 & \textbf{1860} & 1899 & 2116 & 1625 \\
ta30 & \textbf{1805} & 1938 & 1880 & 1999 & 2017 & 1935 & 2017 & 2066 & 1584 \\
ta31 & 2230 & \textbf{2028} & 2066 & 2335 & 2143 & 2134 & 2143 & 2343 & 1764 \\
ta32 & 2315 & \textbf{2108} & 2128 & 2432 & 2188 & 2223 & 2188 & 2472 & 1784 \\
ta33 & \textbf{2193} & 2261 & 2254 & 2453 & 2308 & 2349 & 2308 & 2507 & 1791 \\
ta34 & 2228 & \textbf{2137} & 2164 & 2434 & 2193 & 2245 & 2193 & 2504 & 1829 \\
ta35 & 2249 & 2136 & \textbf{2119} & 2497 & 2255 & 2226 & 2255 & 2503 & 2007 \\
ta36 & 2162 & \textbf{2073} & 2187 & 2445 & 2307 & 2365 & 2307 & 2469 & 1819 \\
ta37 & 2158 & \textbf{2079} & 2090 & 2664 & 2190 & 2130 & 2190 & 2333 & 1771 \\
ta38 & 1974 & \textbf{1954} & 2029 & 2155 & 2179 & 2050 & 2179 & 2287 & 1673 \\
ta39 & 2082 & \textbf{2009} & 2114 & 2477 & 2167 & 2221 & 2167 & 2410 & 1795 \\
ta40 & 2253 & \textbf{1982} & 2007 & 2301 & 2028 & 2205 & 2028 & 2287 & 1669 \\
ta41 & 2645 & \textbf{2394} & 2488 & 2499 & 2538 & 2620 & 2538 & 2729 & 2005 \\
ta42 & 2421 & \textbf{2217} & 2249 & 2710 & 2440 & 2416 & 2440 & 2684 & 1937 \\
ta43 & 2321 & 2254 & \textbf{2202} & 2434 & 2432 & 2345 & 2432 & 2603 & 1846 \\
ta44 & 2350 & \textbf{2303} & 2432 & 2906 & 2426 & 2544 & 2426 & 2757 & 1979 \\
ta45 & 2407 & \textbf{2353} & 2357 & 2640 & 2487 & 2524 & 2487 & 2722 & 2000 \\
ta46 & 2441 & \textbf{2382} & 2488 & 2667 & 2490 & 2447 & 2490 & 2824 & 2004 \\
ta47 & 2326 & \textbf{2232} & 2254 & 2620 & 2286 & 2263 & 2286 & 2597 & 1889 \\
ta48 & 2571 & \textbf{2237} & 2358 & 2620 & 2371 & 2356 & 2371 & 2628 & 1937 \\
ta49 & 2514 & 2380 & \textbf{2375} & 2666 & 2397 & 2382 & 2397 & 2681 & 1968 \\
ta50 & \textbf{2318} & 2337 & 2339 & 2429 & 2469 & 2493 & 2469 & 2493 & 1923 \\
ta51 & 3237 & \textbf{3016} & 3309 & 3856 & 3567 & 3435 & 3567 & 3498 & 2760 \\
ta52 & 2991 & \textbf{2955} & 3139 & 3266 & 3303 & 3394 & 3303 & 3567 & 2756 \\
ta53 & 3036 & \textbf{2939} & 3003 & 3507 & 3115 & 3098 & 3115 & 3243 & 2717 \\
ta54 & 3169 & \textbf{2878} & 2904 & 3142 & 3265 & 3272 & 3265 & 3626 & 2839 \\
ta55 & 3146 & \textbf{2933} & 3110 & 3225 & 3279 & 3188 & 3279 & 3435 & 2679 \\
ta56 & 2938 & \textbf{2865} & 3032 & 3530 & 3100 & 3134 & 3100 & 3352 & 2781 \\
ta57 & 3177 & \textbf{3048} & 3235 & 3725 & 3335 & 3261 & 3335 & 3603 & 2943 \\
ta58 & 3334 & \textbf{3121} & 3290 & 3365 & 3420 & 3365 & 3420 & 3557 & 2885 \\
ta59 & 3013 & 3064 & \textbf{2998} & 3294 & 3117 & 3131 & 3117 & 3543 & 2655 \\
ta60 & 3062 & \textbf{2947} & 3013 & 3500 & 3044 & 3122 & 3044 & 3336 & 2723 \\
ta61 & 3280 & 3237 & \textbf{3220} & 3606 & 3376 & 3343 & 3376 & 3685 & 2868 \\
ta62 & 3345 & \textbf{3131} & 3237 & 3639 & 3417 & 3462 & 3417 & 3766 & 2869 \\
ta63 & 3268 & \textbf{3019} & 3168 & 3521 & 3276 & 3233 & 3276 & 3578 & 2755 \\
ta64 & 3273 & \textbf{2906} & 3037 & 3447 & 3057 & 3188 & 3057 & 3640 & 2702 \\
ta65 & \textbf{3109} & 3114 & 3228 & 3332 & 3249 & 3429 & 3249 & 3597 & 2725 \\
ta66 & 3213 & \textbf{3131} & 3304 & 3677 & 3335 & 3287 & 3335 & 3550 & 2845 \\
ta67 & 3154 & \textbf{3071} & 3179 & 3487 & 3392 & 3351 & 3392 & 3798 & 2825 \\
ta68 & 3184 & \textbf{3032} & 3085 & 3336 & 3251 & 3203 & 3251 & 3695 & 2784 \\
ta69 & 3421 & \textbf{3344} & 3463 & 3862 & 3526 & 3550 & 3526 & 3792 & 3071 \\
ta70 & 3651 & \textbf{3352} & 3406 & 3801 & 3590 & 3482 & 3590 & 3908 & 2995 \\
ta71 & 5757 & \textbf{5499} & 5648 & 6232 & 5938 & 6036 & 5938 & 6507 & 5464 \\
ta72 & 5508 & \textbf{5252} & 5427 & 5973 & 5639 & 5583 & 5639 & 6028 & 5181 \\
ta73 & 6005 & \textbf{5666} & 5837 & 6482 & 6128 & 6050 & 6128 & 6527 & 5568 \\
ta74 & 5425 & \textbf{5339} & 5396 & 6062 & 5642 & 5678 & 5642 & 6076 & 5339 \\
ta75 & 5906 & \textbf{5668} & 5773 & 6217 & 6212 & 6029 & 6212 & 6459 & 5392 \\
ta76 & 5798 & \textbf{5632} & 5844 & 6370 & 5936 & 5887 & 5936 & 6125 & 5342 \\
ta77 & 5616 & \textbf{5566} & 5706 & 6045 & 5829 & 5905 & 5829 & 6284 & 5436 \\
ta78 & 5549 & \textbf{5404} & 5521 & 6143 & 5886 & 5700 & 5886 & 6206 & 5394 \\
ta79 & 5501 & \textbf{5410} & 5525 & 6018 & 5652 & 5749 & 5652 & 6142 & 5358 \\
ta80 & 5543 & \textbf{5483} & 5675 & 5848 & 5707 & 5505 & 5707 & 6248 & 5183 \\
\end{longtable}

\thesisspacing 